\documentclass{article}

\usepackage{microtype}
\usepackage{graphicx}
\usepackage{tabularx}
\usepackage{booktabs} % for professional tables
\usepackage{textcomp}
\usepackage{listings}
\usepackage{amsmath,mathpazo}
\usepackage{amsfonts}
\usepackage{subcaption}
\usepackage{multirow}
\usepackage{caption}
\usepackage{threeparttable}
\usepackage[super]{nth}
\usepackage{soul}

\usepackage{amssymb}
\let\emptyset\varnothing

\usepackage{tcolorbox}
\tcbuselibrary{skins}
\usepackage{xspace}
\newcommand{\irect}[1]{%
  \tikz[baseline=(char.base)]\node[anchor=south west, draw,rectangle, rounded corners=0.5mm, inner sep=1.5pt, minimum size=3mm,outer xsep=0pt,column sep=0pt,row sep=0pt,
    text height=2.2mm](char){#1} ;}
\newcounter{FndCounter}
\newcommand{\Fnd}[1]{%
  \refstepcounter{FndCounter}
  \label{#1}
  \irect{\textit{F}.\theFndCounter}\xspace
}
\newcommand{\RefFnd}[1]{%
  \irect{\hyperref[#1]{\textit{F}.\ref{#1}}}\xspace
}

\newenvironment{rlscope-finding}[1]
{
    \begin{tcolorbox}[size=fbox,after skip=0.4\baselineskip]
    \Fnd{#1} 
    \bfseries
}
{
    \end{tcolorbox}
}
\usepackage{hyperref}
\usepackage{xurl}

\usepackage[accepted]{mlsys2021}

\mlsystitlerunning{Horizontally Fused Training Array: An Effective Hardware Utilization Squeezer for Training Novel Deep Learning Models}

\begin{document}

\twocolumn[
\mlsystitle{Horizontally Fused Training Array: An Effective Hardware Utilization Squeezer for Training Novel Deep Learning Models}

% unless you've provided the [accepted] option to the mlsys2021


\mlsyssetsymbol{equal}{*}

\begin{mlsysauthorlist}
\mlsysauthor{Shang Wang}{nv,vt}
\mlsysauthor{Peiming Yang}{equal,sj,vt}
\mlsysauthor{Yuxuan Zheng}{equal,it}
\mlsysauthor{Xin Li}{equal,vt}
\mlsysauthor{Gennady Pekhimenko}{ut,vt}
\end{mlsysauthorlist}

\mlsysaffiliation{ut}{Department of Computer Science, University of Toronto}
\mlsysaffiliation{vt}{Vector Institute}
\mlsysaffiliation{sj}{Department of Computer Science and Engineering, Shanghai Jiao Tong University}
\mlsysaffiliation{nv}{NVIDIA}
\mlsysaffiliation{it}{Intel}

\mlsyscorrespondingauthor{Shang Wang}{wangsh46@cs.toronto.edu}

\mlsyskeywords{Machine Learning, MLSys}

\vskip 0.3in

% Variables shared in multiple files

%%%%%%%%%%%%%%%%%%%%%%%%%%%%%%%%%%%%%%%%%%%%%%%%%%%%%%
% Peak throughput speedup of HFTA over baselines
%
% gpuv means GPU V100; gpua means GPU A100; gpur means GPU RTX6000; tpu means TPU v3
% Mainly for Table 4
%%%%%%%%%%%%%%%%%%%%%%%%%%%%%%%%%%%%%%%%%%%%%%%%%%%%%%

% V100 Cls
\newcommand{\gpuvFpPeakHFTAvsMPSCls}{2.36}
\newcommand{\gpuvFpPeakHFTAvsConcurrentCls}{2.54}
\newcommand{\gpuvFpPeakHFTAvsSerialCls}{2.62}
\newcommand{\gpuvAmpPeakHFTAvsMPSCls}{4.50}
\newcommand{\gpuvAmpPeakHFTAvsConcurrentCls}{5.02}
\newcommand{\gpuvAmpPeakHFTAvsSerialCls}{5.02}
\newcommand{\gpuvMergePeakHFTAvsSerialCls}{5.02}
\newcommand{\gpuvMergePeakHFTAvsMPSCls}{4.50}
\newcommand{\gpuvMergePeakHFTAvsConcurrentCls}{4.87}

% V100 Seg
\newcommand{\gpuvFpPeakHFTAvsMPSSeg}{1.17}
\newcommand{\gpuvFpPeakHFTAvsConcurrentSeg}{1.62}
\newcommand{\gpuvFpPeakHFTAvsSerialSeg}{1.62}
\newcommand{\gpuvAmpPeakHFTAvsMPSSeg}{3.03}
\newcommand{\gpuvAmpPeakHFTAvsConcurrentSeg}{4.24}
\newcommand{\gpuvAmpPeakHFTAvsSerialSeg}{4.29}
\newcommand{\gpuvMergePeakHFTAvsSerialSeg}{4.29}
\newcommand{\gpuvMergePeakHFTAvsMPSSeg}{3.03}
\newcommand{\gpuvMergePeakHFTAvsConcurrentSeg}{4.24}

% V100 DCGAN
\newcommand{\gpuvFpPeakHFTAvsMPSDCGAN}{1.95}
\newcommand{\gpuvFpPeakHFTAvsConcurrentDCGAN}{1.95}
\newcommand{\gpuvFpPeakHFTAvsSerialDCGAN}{4.18}
\newcommand{\gpuvAmpPeakHFTAvsMPSDCGAN}{2.03}
\newcommand{\gpuvAmpPeakHFTAvsConcurrentDCGAN}{2.01}
\newcommand{\gpuvAmpPeakHFTAvsSerialDCGAN}{4.59}
\newcommand{\gpuvMergePeakHFTAvsSerialDCGAN}{4.59}
\newcommand{\gpuvMergePeakHFTAvsMPSDCGAN}{2.03}
\newcommand{\gpuvMergePeakHFTAvsConcurrentDCGAN}{2.01}

% RTX6000 Cls
\newcommand{\gpurFpPeakHFTAvsMPSCls}{2.07}
\newcommand{\gpurFpPeakHFTAvsConcurrentCls}{2.46}
\newcommand{\gpurFpPeakHFTAvsSerialCls}{2.46}
\newcommand{\gpurAmpPeakHFTAvsMPSCls}{3.79}
\newcommand{\gpurAmpPeakHFTAvsConcurrentCls}{4.26}
\newcommand{\gpurAmpPeakHFTAvsSerialCls}{4.36}
\newcommand{\gpurMergePeakHFTAvsSerialCls}{4.36}
\newcommand{\gpurMergePeakHFTAvsMPSCls}{3.79}
\newcommand{\gpurMergePeakHFTAvsConcurrentCls}{4.26}

% RTX6000 Seg
\newcommand{\gpurFpPeakHFTAvsMPSSeg}{1.22}
\newcommand{\gpurFpPeakHFTAvsConcurrentSeg}{1.95}
\newcommand{\gpurFpPeakHFTAvsSerialSeg}{1.97}
\newcommand{\gpurAmpPeakHFTAvsMPSSeg}{2.54}
\newcommand{\gpurAmpPeakHFTAvsConcurrentSeg}{3.54}
\newcommand{\gpurAmpPeakHFTAvsSerialSeg}{3.63}
\newcommand{\gpurMergePeakHFTAvsSerialSeg}{3.63}
\newcommand{\gpurMergePeakHFTAvsMPSSeg}{2.54}
\newcommand{\gpurMergePeakHFTAvsConcurrentSeg}{3.54}

% RTX6000 DCGAN
\newcommand{\gpurFpPeakHFTAvsMPSDCGAN}{1.69}
\newcommand{\gpurFpPeakHFTAvsConcurrentDCGAN}{1.64}
\newcommand{\gpurFpPeakHFTAvsSerialDCGAN}{6.69}
\newcommand{\gpurAmpPeakHFTAvsMPSDCGAN}{1.82}
\newcommand{\gpurAmpPeakHFTAvsConcurrentDCGAN}{1.72}
\newcommand{\gpurAmpPeakHFTAvsSerialDCGAN}{6.29}
\newcommand{\gpurMergePeakHFTAvsSerialDCGAN}{6.29}
\newcommand{\gpurMergePeakHFTAvsMPSDCGAN}{1.82}
\newcommand{\gpurMergePeakHFTAvsConcurrentDCGAN}{1.72}

% A100 Cls
\newcommand{\gpuaFpPeakHFTAvsMIGCls}{2.10}
\newcommand{\gpuaFpPeakHFTAvsMPSCls}{2.05}
\newcommand{\gpuaFpPeakHFTAvsConcurrentCls}{5.47}
\newcommand{\gpuaFpPeakHFTAvsSerialCls}{5.47}
\newcommand{\gpuaAmpPeakHFTAvsMPSCls}{4.72}
\newcommand{\gpuaAmpPeakHFTAvsMIGCls}{4.88}
\newcommand{\gpuaAmpPeakHFTAvsConcurrentCls}{12.98}
\newcommand{\gpuaAmpPeakHFTAvsSerialCls}{11.50}
\newcommand{\gpuaMergePeakHFTAvsSerialCls}{11.50}
\newcommand{\gpuaMergePeakHFTAvsMPSCls}{4.72}
\newcommand{\gpuaMergePeakHFTAvsConcurrentCls}{12.98}
\newcommand{\gpuaMergePeakHFTAvsMIGCls}{4.88}

% A100 Seg
\newcommand{\gpuaFpPeakHFTAvsMPSSeg}{1.31}
\newcommand{\gpuaFpPeakHFTAvsMIGSeg}{1.35}
\newcommand{\gpuaFpPeakHFTAvsConcurrentSeg}{4.56}
\newcommand{\gpuaFpPeakHFTAvsSerialSeg}{4.56}
\newcommand{\gpuaAmpPeakHFTAvsMPSSeg}{2.93}
\newcommand{\gpuaAmpPeakHFTAvsMIGSeg}{3.02}
\newcommand{\gpuaAmpPeakHFTAvsConcurrentSeg}{10.26}
\newcommand{\gpuaAmpPeakHFTAvsSerialSeg}{9.48}
\newcommand{\gpuaMergePeakHFTAvsSerialSeg}{9.48}
\newcommand{\gpuaMergePeakHFTAvsMPSSeg}{2.93}
\newcommand{\gpuaMergePeakHFTAvsConcurrentSeg}{10.26}
\newcommand{\gpuaMergePeakHFTAvsMIGSeg}{3.02}

% A100 DCGAN
\newcommand{\gpuaFpPeakHFTAvsMPSDCGAN}{1.37}
\newcommand{\gpuaFpPeakHFTAvsMIGDCGAN}{1.59}
\newcommand{\gpuaFpPeakHFTAvsConcurrentDCGAN}{1.39}
\newcommand{\gpuaFpPeakHFTAvsSerialDCGAN}{4.46}
\newcommand{\gpuaAmpPeakHFTAvsMPSDCGAN}{1.09}
\newcommand{\gpuaAmpPeakHFTAvsMIGDCGAN}{1.09}
\newcommand{\gpuaAmpPeakHFTAvsConcurrentDCGAN}{1.06}
\newcommand{\gpuaAmpPeakHFTAvsSerialDCGAN}{3.61}
\newcommand{\gpuaMergePeakHFTAvsSerialDCGAN}{4.41}
\newcommand{\gpuaMergePeakHFTAvsMPSDCGAN}{1.33}
\newcommand{\gpuaMergePeakHFTAvsConcurrentDCGAN}{1.29}
\newcommand{\gpuaMergePeakHFTAvsMIGDCGAN}{1.33}

% TPU for 3 benchmarks
\newcommand{\tpuPeakHFTAvsSerialCls}{4.93}
\newcommand{\tpuPeakHFTAvsSerialSeg}{1.20}
\newcommand{\tpuPeakHFTAvsSerialDCGAN}{15.13}

% V100 ResNet
\newcommand{\gpuvPeakHFTAvsMPSRes}{4.18}
\newcommand{\gpuvPeakHFTAvsConcurrentRes}{4.21}
\newcommand{\gpuvPeakHFTAvsSerialRes}{8.16}

%%%%%%%%%%%%%%%%%%%%%%%%%%%%%%%%%%%%%%%%%%%%%%%%%%%%%%
% Speedup of HFTA over baselines with the same number of workloads
%
% Mainly for Table 5
%%%%%%%%%%%%%%%%%%%%%%%%%%%%%%%%%%%%%%%%%%%%%%%%%%%%%%

% V100 Cls
\newcommand{\gpuvFpSameBHFTAvsMPSCls}{1.65}
\newcommand{\gpuvFpSameBHFTAvsConcurrentCls}{1.77}
\newcommand{\gpuvAmpSameBHFTAvsMPSCls}{3.05}
\newcommand{\gpuvAmpSameBHFTAvsConcurrentCls}{3.41}

% V100 Seg
\newcommand{\gpuvFpSameBHFTAvsMPSSeg}{1.17}
\newcommand{\gpuvFpSameBHFTAvsConcurrentSeg}{1.62}
\newcommand{\gpuvAmpSameBHFTAvsMPSSeg}{2.23}
\newcommand{\gpuvAmpSameBHFTAvsConcurrentSeg}{3.12}

% V100 DCGAN
\newcommand{\gpuvFpSameBHFTAvsMPSDCGAN}{1.95}
\newcommand{\gpuvFpSameBHFTAvsConcurrentDCGAN}{1.91}
\newcommand{\gpuvAmpSameBHFTAvsMPSDCGAN}{2.23}
\newcommand{\gpuvAmpSameBHFTAvsConcurrentDCGAN}{2.27}

% RTX6000 Cls
\newcommand{\gpurFpSameBHFTAvsMPSCls}{1.92}
\newcommand{\gpurFpSameBHFTAvsConcurrentCls}{2.32}
\newcommand{\gpurAmpSameBHFTAvsMPSCls}{3.75}
\newcommand{\gpurAmpSameBHFTAvsConcurrentCls}{4.14}

% RTX6000 Seg
\newcommand{\gpurFpSameBHFTAvsMPSSeg}{1.22}
\newcommand{\gpurFpSameBHFTAvsConcurrentSeg}{1.95}
\newcommand{\gpurAmpSameBHFTAvsMPSSeg}{2.35}
\newcommand{\gpurAmpSameBHFTAvsConcurrentSeg}{3.21}

% RTX6000 DCGAN
\newcommand{\gpurFpSameBHFTAvsMPSDCGAN}{1.78}
\newcommand{\gpurFpSameBHFTAvsConcurrentDCGAN}{1.96}
\newcommand{\gpurAmpSameBHFTAvsMPSDCGAN}{1.90}
\newcommand{\gpurAmpSameBHFTAvsConcurrentDCGAN}{1.73}

% A100 Cls
\newcommand{\gpuaFpSameBHFTAvsMIGCls}{1.51}
\newcommand{\gpuaFpSameBHFTAvsMPSCls}{1.64}
\newcommand{\gpuaFpSameBHFTAvsConcurrentCls}{4.91}
\newcommand{\gpuaAmpSameBHFTAvsMPSCls}{3.18}
\newcommand{\gpuaAmpSameBHFTAvsMIGCls}{2.07}
\newcommand{\gpuaAmpSameBHFTAvsConcurrentCls}{9.16}

% A100 Seg
\newcommand{\gpuaFpSameBHFTAvsMPSSeg}{1.04}
\newcommand{\gpuaFpSameBHFTAvsMIGSeg}{1.07}
\newcommand{\gpuaFpSameBHFTAvsConcurrentSeg}{3.97}
\newcommand{\gpuaAmpSameBHFTAvsMPSSeg}{2.13}
\newcommand{\gpuaAmpSameBHFTAvsMIGSeg}{1.58}
\newcommand{\gpuaAmpSameBHFTAvsConcurrentSeg}{7.86}

% A100 DCGAN
\newcommand{\gpuaFpSameBHFTAvsMPSDCGAN}{9.41}
\newcommand{\gpuaFpSameBHFTAvsMIGDCGAN}{1.51}
\newcommand{\gpuaFpSameBHFTAvsConcurrentDCGAN}{8.94}
\newcommand{\gpuaAmpSameBHFTAvsMPSDCGAN}{7.48}
\newcommand{\gpuaAmpSameBHFTAvsMIGDCGAN}{1.20}
\newcommand{\gpuaAmpSameBHFTAvsConcurrentDCGAN}{9.07}

%%%%%%%%%%%%%%%%%%%%%%%%%%%%%%%%%%%%%%%%%%%%%%%%%%%%%%
% Speedup of AMP over FP32
%
% Mainly for Table 6
%%%%%%%%%%%%%%%%%%%%%%%%%%%%%%%%%%%%%%%%%%%%%%%%%%%%%%

% V100 Cls
\newcommand{\gpuvAmpvsFpSerialCls}{1.00}
\newcommand{\gpuvAmpvsFpConcurrentCls}{0.97}
\newcommand{\gpuvAmpvsFpMPSCls}{1.01}
\newcommand{\gpuvAmpvsFpHFTACls}{1.92}

% V100 Seg
\newcommand{\gpuvAmpvsFpSerialSeg}{1.00}
\newcommand{\gpuvAmpvsFpConcurrentSeg}{1.01}
\newcommand{\gpuvAmpvsFpMPSSeg}{1.03}
\newcommand{\gpuvAmpvsFpHFTASeg}{2.65}

% V100 DCGAN
\newcommand{\gpuvAmpvsFpSerialDCGAN}{1.00}
\newcommand{\gpuvAmpvsFpConcurrentDCGAN}{1.07}
\newcommand{\gpuvAmpvsFpMPSDCGAN}{1.06}
\newcommand{\gpuvAmpvsFpHFTADCGAN}{1.10}

% RTX6000 Cls
\newcommand{\gpurAmpvsFpSerialCls}{1.06}
\newcommand{\gpurAmpvsFpConcurrentCls}{1.09}
\newcommand{\gpurAmpvsFpMPSCls}{1.03}
\newcommand{\gpurAmpvsFpHFTACls}{1.88}

% RTX6000 Seg
\newcommand{\gpurAmpvsFpSerialSeg}{1.19}
\newcommand{\gpurAmpvsFpConcurrentSeg}{1.22}
\newcommand{\gpurAmpvsFpMPSSeg}{1.05}
\newcommand{\gpurAmpvsFpHFTASeg}{2.20}

% RTX6000 DCGAN
\newcommand{\gpurAmpvsFpSerialDCGAN}{1.16}
\newcommand{\gpurAmpvsFpConcurrentDCGAN}{1.05}
\newcommand{\gpurAmpvsFpMPSDCGAN}{1.02}
\newcommand{\gpurAmpvsFpHFTADCGAN}{1.09}

% A100 Cls
\newcommand{\gpuaAmpvsFpSerialCls}{1.13}
\newcommand{\gpuaAmpvsFpConcurrentCls}{1.00}
\newcommand{\gpuaAmpvsFpMPSCls}{1.03}
\newcommand{\gpuaAmpvsFpMIGCls}{1.02}
\newcommand{\gpuaAmpvsFpHFTACls}{2.37}

% A100 Seg
\newcommand{\gpuaAmpvsFpSerialSeg}{1.13}
\newcommand{\gpuaAmpvsFpConcurrentSeg}{1.05}
\newcommand{\gpuaAmpvsFpMPSSeg}{1.06}
\newcommand{\gpuaAmpvsFpMIGSeg}{1.05}
\newcommand{\gpuaAmpvsFpHFTASeg}{2.36}

% A100 DCGAN
\newcommand{\gpuaAmpvsFpSerialDCGAN}{1.01}
\newcommand{\gpuaAmpvsFpConcurrentDCGAN}{1.08}
\newcommand{\gpuaAmpvsFpMPSDCGAN}{1.03}
\newcommand{\gpuaAmpvsFpMIGDCGAN}{1.20}
\newcommand{\gpuaAmpvsFpHFTADCGAN}{0.82}

% Max values
% We are using only V100 for this data now
\newcommand{\maxGpuAmpvsFpSerial}{1.00} % \gpuvAmpvsFpSerialSeg
\newcommand{\maxGpuAmpvsFpConcurrent}{1.07} % \gpuvAmpvsFpConcurrentDCGAN
\newcommand{\maxGpuAmpvsFpMPS}{1.06} % \gpuvAmpvsFpMPSDCGAN
\newcommand{\maxGpuAmpvsFpHFTA}{2.65} % \gpuvAmpvsFpHFTASeg

%%%%%%%%%%%%%%%%%%%%%%%%%%%%%%%%%%%%%%%%%%%%%%%%%%%%%%
% Summary of performance data for introduction
%%%%%%%%%%%%%%%%%%%%%%%%%%%%%%%%%%%%%%%%%%%%%%%%%%%%%%

% This is the ratio of "max(hfta_fp32, hfta_amp) / max(mps_fp32, mps_amp)"
\newcommand{\gpuPeakHFTAvsMPSUnderRange}{1.33} % DCGAN A100
\newcommand{\gpuPeakHFTAvsMPSUpperRange}{4.72} % Cls A100
\newcommand{\gpuPeakHFTAvsMIGUnderRange}{1.33} % DCGAN A100
\newcommand{\gpuPeakHFTAvsMIGUpperRange}{4.88} % Cls A100
\newcommand{\gpuPeakHFTAvsSerialUnderRange}{3.63} % Seg RTX6000
\newcommand{\gpuPeakHFTAvsSerialUpperRange}{11.50} % Cls A100
\newcommand{\tpuPeakHFTAvsSerialUnderRange}{4.93} % Cls TPU
\newcommand{\tpuPeakHFTAvsSerialUpperRange}{15.13} % DCGAN TPU
\newcommand{\tpuPeakHFTAvsSerialUpperRangeOneDecimal}{15.1} % DCGAN TPU

% For secondary benchmarks in appendix
\newcommand{\gpuPeakHFTAvsSerialUnderRangeNewModels}{2.42} % Transformer V100
\newcommand{\gpuPeakHFTAvsSerialUpperRangeNewModels}{3.94} % MobileNet V100
\newcommand{\gpuPeakHFTAvsConcurrentUnderRangeNewModels}{1.67} % Transformer V100
\newcommand{\gpuPeakHFTAvsConcurrentUpperRangeNewModels}{3.02} % Bert V100
\newcommand{\gpuPeakHFTAvsMPSUnderRangeNewModels}{1.25} % Transformer V100
\newcommand{\gpuPeakHFTAvsMPSUpperRangeNewModels}{2.24} % MobileNet V100
\newcommand{\tpuPeakHFTAvsSerialUnderRangeNewModels}{2.98} % Bert TPU
\newcommand{\tpuPeakHFTAvsSerialUpperRangeNewModels}{6.43} % MobileNet TPU

%%%%%%%%%%%%%%%%%%%%%%%%%%%%%%%%%%%%%%%%%%%%%%%%%%%%%%
% Scalability of HFTA over MPS & Concurrent
%%%%%%%%%%%%%%%%%%%%%%%%%%%%%%%%%%%%%%%%%%%%%%%%%%%%%%

% Ratio of max number of workloads (B) of HFTA over MPS and Concurrent
% Under limit % Not needed right now
\newcommand{\numWorkloadsPNClsUnderRange}{1.67} % RTX6000
\newcommand{\numWorkloadsPNSegUnderRange}{1.50} % RTX6000
\newcommand{\numWorkloadsDCGANUnderRange}{6.54} % RTX6000

% Upper limit
\newcommand{\numWorkloadsPNClsUpperRange}{1.80} % V100
\newcommand{\numWorkloadsPNSegUpperRange}{1.60} % V100
\newcommand{\numWorkloadsDCGANUpperRange}{7.57} % V100
\newcommand{\numWorkloadsResNetUpperRange}{9.43} % V100

%%%%%%%%%%%%%%%%%%%%%%%%%%%%%%%%%%%%%%%%%%%%%%%%%%%%%%
% Memory Overhead from linear regression of GPU memory points
%%%%%%%%%%%%%%%%%%%%%%%%%%%%%%%%%%%%%%%%%%%%%%%%%%%%%%

\newcommand{\hftaFpMemOverheadCls}{1.52}
\newcommand{\hftaAmpMemOverheadCls}{2.12}

%%%%%%%%%%%%%%%%%%%%%%%%%%%%%%%%%%%%%%%%%%%%%%%%%%%%%%
% HFHT evaluation results (GPU hours)
%%%%%%%%%%%%%%%%%%%%%%%%%%%%%%%%%%%%%%%%%%%%%%%%%%%%%%

%%% PointNet Classification
%% Hyperband
% Serial
\newcommand{\hfhtGpuvPnClsHbSerialFp}{2.21}
\newcommand{\hfhtGpuvPnClsHbSerialAmp}{2.31}
% Concurrent
\newcommand{\hfhtGpuvPnClsHbConcurrentFp}{2.09}
\newcommand{\hfhtGpuvPnClsHbConcurrentAmp}{2.12}
% MPS
\newcommand{\hfhtGpuvPnClsHbMpsFp}{1.97}
\newcommand{\hfhtGpuvPnClsHbMpsAmp}{1.99}
% HFTA
\newcommand{\hfhtGpuvPnClsHbHftaFp}{1.10}
\newcommand{\hfhtGpuvPnClsHbHftaAmp}{1.00}

%% Random Search
% Serial
\newcommand{\hfhtGpuvPnClsRsSerialFp}{5.10}
\newcommand{\hfhtGpuvPnClsRsSerialAmp}{5.33}
% Concurrent
\newcommand{\hfhtGpuvPnClsRsConcurrentFp}{4.81}
\newcommand{\hfhtGpuvPnClsRsConcurrentAmp}{4.94}
% MPS
\newcommand{\hfhtGpuvPnClsRsMpsFp}{4.51}
\newcommand{\hfhtGpuvPnClsRsMpsAmp}{4.57}
% HFTA
\newcommand{\hfhtGpuvPnClsRsHftaFp}{1.49}
\newcommand{\hfhtGpuvPnClsRsHftaAmp}{1.00}

%%% MobileNet
%% Hyperband
% Serial
\newcommand{\hfhtGpuvMnHbSerialFp}{}
\newcommand{\hfhtGpuvMnHbSerialAmp}{}
% Concurrent
\newcommand{\hfhtGpuvMnHbConcurrentFp}{}
\newcommand{\hfhtGpuvMnHbConcurrentAmp}{}
% MPS
\newcommand{\hfhtGpuvMnHbMpsFp}{}
\newcommand{\hfhtGpuvMnHbMpsAmp}{}
% HFTA
\newcommand{\hfhtGpuvMnHbHftaFp}{}
\newcommand{\hfhtGpuvMnHbHftaAmp}{}

%% Random Search
% Serial
\newcommand{\hfhtGpuvMnRsSerialFp}{}
\newcommand{\hfhtGpuvMnRsSerialAmp}{}
% Concurrent
\newcommand{\hfhtGpuvMnRsConcurrentFp}{}
\newcommand{\hfhtGpuvMnRsConcurrentAmp}{}
% MPS
\newcommand{\hfhtGpuvMnRsMpsFp}{}
\newcommand{\hfhtGpuvMnRsMpsAmp}{}
% HFTA
\newcommand{\hfhtGpuvMnRsHftaFp}{}
\newcommand{\hfhtGpuvMnRsHftaAmp}{}
\vspace{-0.4cm}
\begin{abstract}
Driven by the tremendous effort in researching novel deep learning (DL) algorithms, the training cost of developing new models increases staggeringly in recent years. %Despite the much effort on optimizing DL training workloads from system and architecture communities, the ``real-world'' GPU cluster usage statistics we collect 
%To  
%reduce this training cost and optimize the cluster-wide hardware resource usage, we 
%collect and analyze 
%analyze
%``real-world'' GPU cluster usage statistics.
%GPU cluster usage statistics from a well-known research institute.
We analyze GPU cluster usage statistics from a 
%well-known 
top research institute 
%to get 
for more insights into the hardware efficiency 
achieved by 
%of 
typical DL training jobs.
Our study 
%suggest that DL researchers and practitioners often lack the required expertise to independently optimize their own workloads, resulting in single-accelerator training jobs with severe hardware under-utilization that, when launched repetitively (e.g., for hyper-parameter tuning), dominates the compute cluster resource consumption. 
reveals that single-accelerator training jobs can dominate the cluster-wide resource consumption when launched repetitively (e.g., for hyper-parameter tuning) while severely under-utilizing the hardware. 
%This is because DL researchers and practitioners often lack the required expertise to independently optimize their own workloads.
Fortunately, we observe that such workloads have the following unique characteristics: (i) the models among jobs often have the same types of operators with the same shapes, and (ii) the inter-model horizontal fusion of such operators is mathematically equivalent to other already well-optimized operators. 
%\hl{Fortunately, we observe that such workloads have unique characteristics that exposes the opportunity for \emph{inter-model horizontal operator fusion}.}
Thus, to help DL researchers and practitioners effectively improve the hardware utilization of their novel DL training workloads, we propose \textbf{H}orizontally \textbf{F}used \textbf{T}raining \textbf{A}rray (HFTA). HFTA is a new DL framework extension library that \emph{horizontally fuses} the models from different repetitive jobs deeply down to operators and then trains them simultaneously on a shared accelerator. 
To show the generality of our solution, we apply HFTA to six 
%different
%emerging 
DL 
%training workloads running 
models training
on state-of-the-art accelerators (GPUs and TPUs). 
Our 
%experimental 
results indicate that HFTA is highly effective in improving hardware utilization and achieves up to $\tpuPeakHFTAvsSerialUpperRangeOneDecimal \times$ higher training throughput vs. the standard practice of running each job on a separate accelerator.
%On the three instances of emerging DL training workloads (PointNet classification, PointNet segmentation and DCGAN) and state-of-the-art (SOTA) DL accelerators (e.g., NVIDIA A100 GPUs and Google TPU v3), HFTA demonstrates strong effectiveness on squeezing out hardware utilization and achieves up to $\tpuPeakHFTAvsSerialUpperRange \times$ higher training throughput than the common practice of running each job on a separate accelerator, and on GPUs, up to $\gpuPeakHFTAvsMPSUpperRange \times$ and $\gpuPeakHFTAvsMIGUpperRange \times$ respectively than the stronger baselines of MPS and MIG which are the SOTA commercial hardware-sharing solutions.
\end{abstract}\vspace{-0.2cm}
]



\printAffiliationsAndNotice{\mlsysEqualContribution} % otherwise use the standard text.

\section{Introduction} \label{sec:intro}
\vspace{-0.2cm}

Deep Learning (DL) algorithms have facilitated tremendous progress in a range of domains, 
%Deep learning (DL) algorithms are applied in various domains nowadays
including natural language translation \cite{gnmt}, recommendation systems \cite{dlrm}, magnetic resonance imaging segmentation \cite{mri}, video game bots \cite{openai_dota}, real-time high-resolution rendering \cite{dlss}, and 
very-large-scale integrated 
circuit placement \cite{dreamplace}. 
This is driven by the abundant and continuous efforts in researching and developing novel DL models by 
%both 
academia and industry in recent years. 
Developing these models is computationally intensive, requiring an army of expensive, specialized accelerators such as GPUs and TPUs~\cite{tpu}, leading to staggeringly
high training costs~\cite{dawnbench, openai_ai_and_compute, tbd, mlperf_train, daydream}.
%  [Why? I think the focus of this paper is not on training large distributed SOTA models, 
%   which are heavily tuned, but on the more widely observed training tasks]
%However, these state-of-the-art (SOTA) DL models become more and more sophisticated and compute-hungry with exponential growing training cost~\cite{openai_ai_and_compute, dawnbench, tbd, mlperf_train, daydream}.

%To study the composition of the training cost with the hope of finding solutions to optimize the cluster-wide hardware resource usage, we collected GPU usage statistics on a large GPU cluster from a well-known top research institution over two consecutive months. 
To reduce this training cost and optimize the cluster-wide hardware resource usage, we analyze 
GPU usage statistics over two consecutive months on a large GPU cluster from the Vector Institute~\cite{vector}.
We observe that, despite significant attention on optimizing DL training workloads from the computer system and architecture communities, especially on distributed training optimizations~\cite{nvidia_rnn_fusion, mxnet_checkpointing, deep_gradient_compression, zero, mlperf_train}, single-accelerator (e.g., single-GPU) training jobs, often launched repetitively by DL researchers (to perform \emph{hyper-parameter tuning}, \emph{model architecture search} or \emph{convergence stability tests}), can (i) dominate the cluster-wide hardware resource consumption (e.g., 46.2\% in our study) while (ii) having extremely low hardware utilization (Section~\ref{sec:inefficiency_repetitive_training_workloads} and \ref{sec:gpu_perf_analysis}).
%as we show in Section~\ref{sec:inefficiency_repetitive_training_workloads}, we found that 
%single-accelerator (e.g., single-GPU) training jobs, often launched repetitively by DL researchers (using techniques such as \emph{hyper-parameter tuning}, \emph{model architecture search} and \emph{convergence stability test}), can sometimes dominate (46.2\%) the cluster-wide hardware resource consumption. Unfortunately, many of these jobs severely under-utilize the accelerators, leading to humongous resource wastage.

The root cause of this phenomenon is manifold. DL researchers and practitioners often lack the expertise to 
%independently 
optimize their 
%own 
training workloads 
independently. 
As a result, basic %na\"ive 
techniques, such as increasing the batch size, often become the only approach at their disposal to improve hardware utilization. 
However, this technique can be impractical due to many reasons, including generalization gap~\cite{generalization_gap}, batch size scaling limit~\cite{batch_size_scaling_limit}, and GAN training instability~\cite{gan_open_questions}. 
On the other hand, accelerators (e.g., GPUs and TPUs) evolve towards more computing power and larger memory capacities (Table~\ref{tab:tpu_spec} and \ref{tab:gpu_spec}).
%, and this 
This trend amplifies the severity of the hardware under-utilization caused by the inability of such training workloads to scale their performance well. 

Thus, this phenomenon motivates \emph{hardware sharing} approaches. To the best of our knowledge, the only widely used hardware-based sharing solutions applicable to DL training are the MPS~\cite{mps} and MIG~\cite{mig} features on NVIDIA GPUs. However, as we later show in Section~\ref{sec:hardware_sharing_hardware_feature}, these generic GPU sharing features that aim at arbitrary workloads are far from the ``silver bullets'' to effectively improve the hardware utilization in the case of repetitive single-GPU training workloads. 
The situation is even worse for emerging DL accelerators (e.g., TPUs) that currently do not have any hardware-based sharing features.

To address such hardware under-utilization on 
%a variety of 
various accelerators, we make two key observations based on the unique characteristics of these workloads. 
First, the models across jobs belonging to the same workload (e.g., hyper-parameter tuning) often have the \emph{same types} of operators with the \emph{same shapes}. 
Second, if these operators are horizontally fused across the models, the outcome is mathematically equivalent to other well-optimized operators found in existing DL framework stacks and accelerators (e.g., fusing multiple convolution operators can be realized using grouped convolutions). 
Inspired by these key observations, we propose to horizontally merge multiple training jobs with the same or similar DL models by deeply \emph{fusing} most, if not all, operators in those models. 
These models' training is then performed collectively on the same shared accelerator (instead of training each model separately on its accelerator). 
Our proposed idea of \emph{inter-model horizontal fusion} is drastically different from and also more effective than major-related prior works as it better exercises the full potential of modern accelerators while
(i) not relying on the generic sharing primitives (e.g., CUDA streams) that are ineffective for repetitive single-GPU workloads, and (ii) avoiding limited fusion techniques that, for example, support only stateless operators or require the weights across models to be the same \cite{hivemind}.

We leverage this novel idea to build a new DL framework extension library for DL researchers and practitioners, called \textbf{H}orizontally \textbf{F}used \textbf{T}raining \textbf{A}rray (\textbf{HFTA})\footnote{\label{footnote:hfta_github}\url{https://github.com/UofT-EcoSystem/hfta}}, that
%greatly 
significantly simplifies the adoption of our proposed inter-model horizontal fusion technique. In summary, this work makes the following major contributions.

$\bullet$ To understand the nature of the jobs running on modern DL accelerator clusters, we collect and study GPU cluster usage statistics, including 51K jobs running for 472K GPU hours in total, from real research workloads. The results of this study demonstrate that repetitive single-accelerator training jobs (i) dominate the hardware resource usage (i.e., 46.2\%) and (ii) have extremely low hardware utilization.

$\bullet$ Motivated by this study, we make two key observations about these jobs that our proposal is built upon: (1) The models often have the same types of operators with the same shapes. (2) The inter-model horizontal fusion of such operators is mathematically equivalent to other existing and well-optimized operators.

$\bullet$ We develop HFTA, a new library that helps DL researchers and practitioners (even with 
a
limited computer system and architecture expertise) to easily extract better performance from their hardware when training novel DL models. 
While doing so, we avoid (i) 
%the introduction of 
introducing any additional device-specific operator implementations that would limit the generality of our idea across different accelerators and (ii) any effect on individual models' convergence as the speedup is achieved only through mathematically equivalent transformations. HFTA 
%is applicable 
applies to a wide variety of models and can run on any hardware backends supported by existing DL frameworks.
Furthermore, we propose a simple yet effective method to integrate HFTA into existing 
%hyper-parameter 
tuning algorithms and develop a lightweight 
%hyper-parameter 
tuning framework named \textbf{H}orizontally \textbf{F}used \textbf{H}yper-parameter \textbf{T}uning (\textbf{HFHT}).

$\bullet$ 
%We evaluate HFTA on the PointNet \cite{pointnet_pytorch} classification and segmentation tasks (ShapeNet part \cite{shapenet_part} dataset), and on DCGAN \cite{dcgan} (LSUN \cite{lsun} dataset), which are examples of highly impactful DL models in the machine learning (ML) community, but not yet fully investigated/optimized by the experienced system engineers and computer architects.\footnote{As opposed to the models from the MLPerf Training Benchmark suite \cite{mlperf_train} that are intensively optimized.} 
We evaluate HFTA on six highly impactful DL models, covering a wide range of tasks, in the machine learning (ML) community.
On modern GPUs (V100, RTX6000
%, 
and A100), HFTA achieves $\gpuPeakHFTAvsSerialUnderRangeNewModels \times$ to $\gpuPeakHFTAvsSerialUpperRange \times$ higher training throughput than running the training jobs without sharing, which is commonly employed by hyper-parameter tuning 
%frameworks~\cite{wandb_sweep, hyperparameter_tuning_common_practice},
frameworks~\cite{hyperparameter_tuning_common_practice},
$\gpuPeakHFTAvsMPSUnderRangeNewModels \times$ to $\gpuPeakHFTAvsMPSUpperRange \times$ than MPS and $\gpuPeakHFTAvsMIGUnderRange \times$ to $\gpuPeakHFTAvsMIGUpperRange \times$ than MIG. HFTA can also fit $\numWorkloadsPNSegUnderRange \times$ to $\numWorkloadsResNetUpperRange \times$ more training jobs on the same GPU than MPS. On TPUs, 
%which currently do not have hardware sharing support, 
HFTA achieves $\tpuPeakHFTAvsSerialUnderRangeNewModels \times$ to $\tpuPeakHFTAvsSerialUpperRange \times$ higher training throughput, which demonstrates HFTA's ability to 
%significantly 
improve performance across different hardware backends significantly. 
Finally, when HFTA is integrated into two tuning algorithms via HFHT, we reduce the cost of total GPU hours by up to $5.10 \times$ among four end-to-end hyper-parameter tuning workloads.

\vspace{-0.3cm}
\section{Background and Motivation} \label{sec:background}
\vspace{-0.2cm}

\subsection{Inefficiency in Repetitive Training Jobs} \label{sec:inefficiency_repetitive_training_workloads}
\vspace{-0.2cm}

As DL research continues to evolve in recent years, the accompanied training cost 
%has been increasing 
increased dramatically. 
For example, \cite{openai_ai_and_compute} show that the amount of compute for training SOTA DL models doubles every 3.4 month, outpacing even Moore’s Law~\cite{moores_law}. Motivated by the practical goal of reducing cluster-wide training cost, using the methodology detailed in Appendix~\ref{append:hardware_under_utilization}, we collect and study the GPU usage statistics of real research workloads for two consecutive months on a large GPU cluster from the Vector Institute~\cite{vector}. 
To our surprise, we find that single-accelerator 
(e.g., single-GPU) 
training jobs dominate the cluster-wide hardware resource consumption when these jobs are launched repetitively in groups, and the aggregated cost of these jobs can even outweigh that of distributed training (the primary focus of many research efforts from the computer system and architecture communities~\cite{deep_gradient_compression, p3, zero, mlperf_train, pytorch_distributed}).
%Despite the many research efforts on optimizing large-scale distributed training workloads from the computer system and architecture communities~\cite{deep_gradient_compression, p3, zero, mlperf_train, pytorch_distributed}, to our surprise, we found that single-accelerator training jobs can dominate the cluster-wide hardware resource consumption when these jobs are launched repetitively, and their aggregated cost can even outweigh that of distributed training. 
Potential reasons of these repetitive jobs include (but are not limited to) hyper-parameter tuning \cite{hyper_parameter_tuning_energy_cost} and convergence stability testing.

\textbf{Background}\hspace{0.1cm} \emph{Hyper-parameter tuning} finds the optimal set of hyper-parameters unknown a priori, which are usually necessary for building accurate models targeting a previously unexplored problem \cite{algorithms_for_hyper_parameter_optimization, random_search_for_hyper_parameter_optimization}. Typical hyper-parameters include learning rates, the choices of weight initializers, and optimizer settings. \emph{Model architecture search}~\cite{neural_architecture_search} is a subset of hyper-parameter tuning where the hyper-parameters directly impact the model architecture (e.g., the number of layers). \emph{Convergence stability testing} trains the same model many times with different random seeds to verify the final accuracy results. 

\begin{table}[]
\centering
\vspace{-0.6cm}
\caption{GPU hour usage breakdown for two consecutive months of a large GPU cluster from the Vector Institute.} \label{tab:vector_gpu_hour}
\scriptsize
\begin{tabular}{l|llll}
\toprule
\begin{tabular}[c]{@{}l@{}}Training \\ Jobs\end{tabular} & \begin{tabular}[c]{@{}l@{}}Repetitive\\ Single-\\ GPU\end{tabular} & \begin{tabular}[c]{@{}l@{}}Isolated\\ Single-\\ GPU\end{tabular} & Distributed & Other \\ \midrule
GPU Hours   & \textbf{218K}(\textbf{46.2}\%)   & 19K(3.5\%)   & 113K(24.0\%) & 124K(26.3\%)     \\ 
\bottomrule
\end{tabular}
\vspace{-0.7cm}
\end{table}

In our study, we classify the jobs into four main categories: (1) multi-node or single-node distributed training, (2) repetitive single-GPU training, (3) isolated single-GPU training, and (4) others (meaning the jobs that do not belong to the first three categories or can not be identified).
Table~\ref{tab:vector_gpu_hour} shows the GPU hour usage distribution among these categories, from which we can observe that the repetitive single-GPU training jobs consume as much as 46.2\% of the cluster-wide total GPU hours.
Furthermore, those repetitive single-accelerator training jobs often have low hardware utilization, as we show in Appendix~\ref{append:hardware_under_utilization}. The cause of such phenomenon is manifold:

$\bullet$ Improving the hardware utilization for DL training jobs can be very challenging. DL researchers and practitioners often lack the system and architecture expertise to optimize their training workloads
%on their own
independently. 
Increasing the batch size, which is the na\"ive and often the only approach at their disposal to increase hardware utilization, is not universally applicable. 
For instance, large batch sizes can lead to training instability for the generative adversarial networks (GAN) \cite{gan_open_questions, biggan}, generalization gap \cite{generalization_gap}, and diminishing returns due to batch size scaling limit \cite{batch_size_scaling_limit}.
Even with the help from the computer system and architecture experts, applying various advanced optimization techniques (e.g., kernel fusion~\cite{nvidia_rnn_fusion} or checkpointing~\cite{mxnet_checkpointing, echo}) on each new model requires an enormous amount of engineering efforts~\cite{mlperf_train}. 
Meanwhile, novel DL models are being proposed at the exponential pace in recent years \cite{nips_stats}.

$\bullet$ As DL research progresses, accelerators (e.g., GPUs and TPUs~\cite{tpu}) evolve towards more compute power (e.g., more streaming multiprocessors (SMs) and the introduction of specialized compute units for fast matrix multiplications in GPUs called tensor cores (TCs)~\cite{tensor_core}) and larger memory capacity/bandwidth. 
We can observe this trend from Tables~\ref{tab:tpu_spec} and \ref{tab:gpu_spec} that list the specifications of the most recent NVIDIA data center GPUs and Google Cloud TPUs, where the largest accelerators suffer from under-utilization the most.

\begin{table}[]
\centering
\vspace{-0.6cm}
\caption{Cloud TPU Core Specifications \cite{tpu_architecture}} \label{tab:tpu_spec}
\scriptsize
\begin{threeparttable}
\begin{tabular}{l|lll}
\toprule
TPU      & v2 (2017) & v3 (2018) & v4 (2020?)\tnote{\textdagger} \\\midrule
MXUs     & 1         & 2         & $\geq 4$ ?                            \\
Memory (HBM) & 8 GB        & 16 GB       & ? GB                            \\
\bottomrule
\end{tabular}
\begin{tablenotes}
\item[\textdagger] TPU v4 is expected to double the FLOPs of TPU v3 along with other enhancements \cite{tpu_v4}.
\end{tablenotes}
\end{threeparttable}
\vspace{-0.3cm}
\end{table}

\begin{table}[]
\centering
\vspace{-0.3cm}
\caption{NVIDIA Data Center GPU Specifications} \label{tab:gpu_spec}
\scriptsize
\begin{tabular}{l|llll}
\toprule
GPU         & SMs & HBM (GB) & HBM Bandwidth & TC Types      \\\midrule
P100 (2016) & 56  & 12/16    & 549/732 GB/s  & -             \\
V100 (2018) & 80  & 16/32    & 900 GB/s      & FP16          \\
A100 (2020) & 108 & 40       & 1.6 TB/s      & TF32 \& FP16  \\
\bottomrule
\end{tabular}
\vspace{-0.8cm}
\end{table}

The fast development of both new DL models and accelerators together exacerbates the hardware under-utilization from repetitive single-accelerator training jobs, which motivates hardware sharing methods discussed below.

\vspace{-0.3cm}
\subsection{Hardware-based Sharing} \label{sec:hardware_sharing_hardware_feature}
\vspace{-0.2cm}

The most well-known and (to the best of our knowledge) the only widely-used hardware-based sharing solutions applicable to DL training workloads\footnote{AMD GPUs also have a hardware-based sharing feature called CU-mask~\cite{cu_masking}; however, we skip its discussion due to their irrelevance in mainstream training workloads.} are the Multi-Process Service (MPS)~\cite{mps} and Multi-Instance GPU (MIG)~\cite{mig} on NVIDIA GPUs. MPS allows CUDA kernels from different processes to potentially run concurrently on the same GPU via a hardware feature called Hyper-Q~\cite{hyperq}. MIG, which is currently only available on the most recent A100 GPUs~\cite{a100}, partitions a single GPU into multiple (up to 7) isolated GPU instances (GIs) where each job now run on a single GI.
%However, two logically parallel kernels can only execute concurrently if the former-issued kernel leaves enough SM resources for the other kernel, which often only happens when executing the the former kernel's last thread-block wave (causing \emph{head-of-line blocking}) \cite{head_of_line_blocking}. 
%Combining with the complexity in determining the kernel tiling strategies, logically parallel kernels executing under MPS can lead to SM under-utilization even if each kernel itself has the minimal latency.
%This is because, from the kernel developers' perspective, the optimization goal is to minimize the latency of each kernel, which naturally leads to distributing the work among as many SMs as needed, even if each SM itself is under-utilized.

%MIG, which is currently only available on the most recent A100 GPUs\cite{a100}, partitions a single GPU into multiple (up to 7) isolated GPU instances (GIs) where each training job now execute on a single GI. However, 
%when the training workload incurs high runtime overhead on the host (e.g., with low GPU utilization),
%even with the finest granularity of MIG (7 GIs), each workload can still under-utilize a single GI as we show in Section~\ref{sec:eval_gpus}.
However, as we quantitatively demonstrate in Section~\ref{sec:eval_gpus}, both MPS and MIG still leave the significant potential of training performance unharnessed due to the following reasons.
First, both MPS and MIG duplicate the runtime overhead among kernels from different training jobs, including kernel launches~\cite{kernel_launch_overhead}, GEMM setups and teardowns~\cite{setup_teardown_overhead}, and/or memory format conversions (related explicitly to TCs)~\cite{memory_format_conversion_overhead}. 
Thus, they can not effectively improve the SM and TC utilization.
Second, both MPS and MIG require running training jobs as separate processes, which duplicates the GPU memory overhead reserved by the DL framework stack \cite{pytorch_reserved_gpu_mem} and leads to a higher overall GPU memory footprint.
Therefore, we can fit fewer training jobs into the same GPU.
Finally, MIG's partitioning granularity can be too coarse for many training workloads.
Even with the finest granularity of MIG (7 GIs), each job can still under-utilize a single GI, as we show in Section~\ref{sec:eval_gpus}.

%Moreover, running training jobs as separate processes via both MPS and MIG causes the following two problems (as we empirically show in Section~\ref{sec:gpu_perf_analysis}). First, such approaches duplicate the GPU memory overhead reserved by the DL framework stack \cite{pytorch_reserved_gpu_mem} and lead to a higher overall GPU memory footprint, thus, we can fit fewer training jobs into the same GPU. Second, such approaches duplicate the runtime overhead from kernel launches \cite{kernel_launch_overhead}, GEMM setups and teardowns \cite{setup_teardown_overhead}, and/or memory format conversions \cite{memory_format_conversion_overhead}, thus, can not effectively improve the utilization of TCs.

%\input{prior_works}

\vspace{-0.3cm}
\section{Our Proposal: HFTA} \label{sec:proposed_method}
\vspace{-0.2cm}

To address the challenge of improving hardware utilization for novel repetitive training workloads on a variety of accelerators, we make the following two key observations on the unique characteristics of these workloads:

$\bullet$ When launched repetitively (such as during hyper-parameter tuning or convergence stability testing), the models used across these jobs often have the \emph{same types} of operators with the \emph{same shapes}.

$\bullet$ \emph{Horizontally fusing} the same types of operators with the same shapes often results in other mathematically equivalent operators that already exist in many SOTA DL models and thus have been optimized in most DL framework stacks on different accelerators.   

\begin{figure}[t]
    \centering
    \vspace{-0.3cm}
    \includegraphics[width=0.9\linewidth]{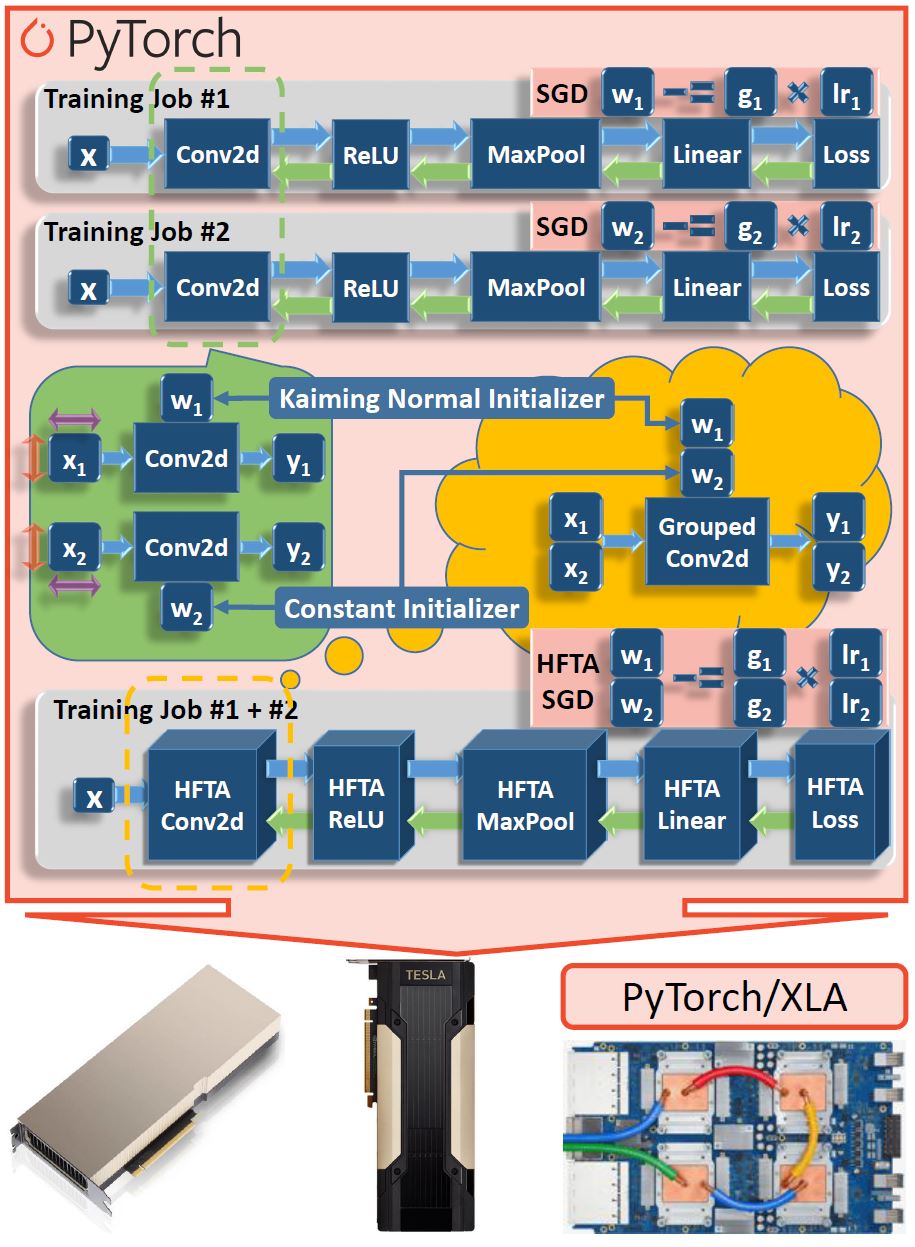} 
    \vspace{-0.5cm}
    \caption{An example showing the key idea of HFTA where two training jobs for hyper-parameter tuning are fused into one via inter-model horizontal operator fusion.}
    \label{fig:horizontal_fusion}
    \vspace{-0.5cm}
\end{figure}

Figure~\ref{fig:horizontal_fusion} explains the above observations with a concrete example of hyper-parameter tuning where the goal is to determine which weight initializer and learning rate work the best. Regardless of which weight initializer or learning rate is used, the first operators in both models are Conv2d of the same shape; the horizontal fusion of many Conv2d operators is mathematically equivalent to a grouped Conv2d which is already used in the ResNeXt \cite{resnext} and MobileNets \cite{mobilenets} models and supported by cuDNN \cite{cudnn_group_conv} on NVDIA GPUs and XLA \cite{xla} on TPUs.

Inspired by the above observations, instead of the common practice~\cite{hyperparameter_tuning_common_practice} of running each job with a single model on a separate accelerator, we propose to better utilize existing hardware by deeply fusing the the same (class of) models across multiple jobs together.
Most, if not all, operators of these models can be \emph{horizontally fused}, and we train these models simultaneously on the same accelerator. 
Thus, as depicted in Figure~\ref{fig:horizontal_fusion}, we can fuse many training jobs into a single one, without the need to implement any new device-specific operator from scratch, which is both time-consuming and error-prone. 
Moreover, this approach easily generalizes to any hardware backends that the DL frameworks already support 
%(e.g., with PyTorch, we can already support all NVIDIA GPUs and Google TPUs). 
(e.g., all NVIDIA GPUs and Google TPUs in the case of PyTorch)
Since horizontal operator fusion can be performed for both single-accelerator and distributed training, our approach 
%is applicable 
applies to both use cases.
%\footnote{Although we assume that no one would scale out a training workload if it even under-utilizes a single accelerator.}

\begin{figure}[t]
    \centering
    \vspace{-0.0cm}
    \includegraphics[width=\linewidth]{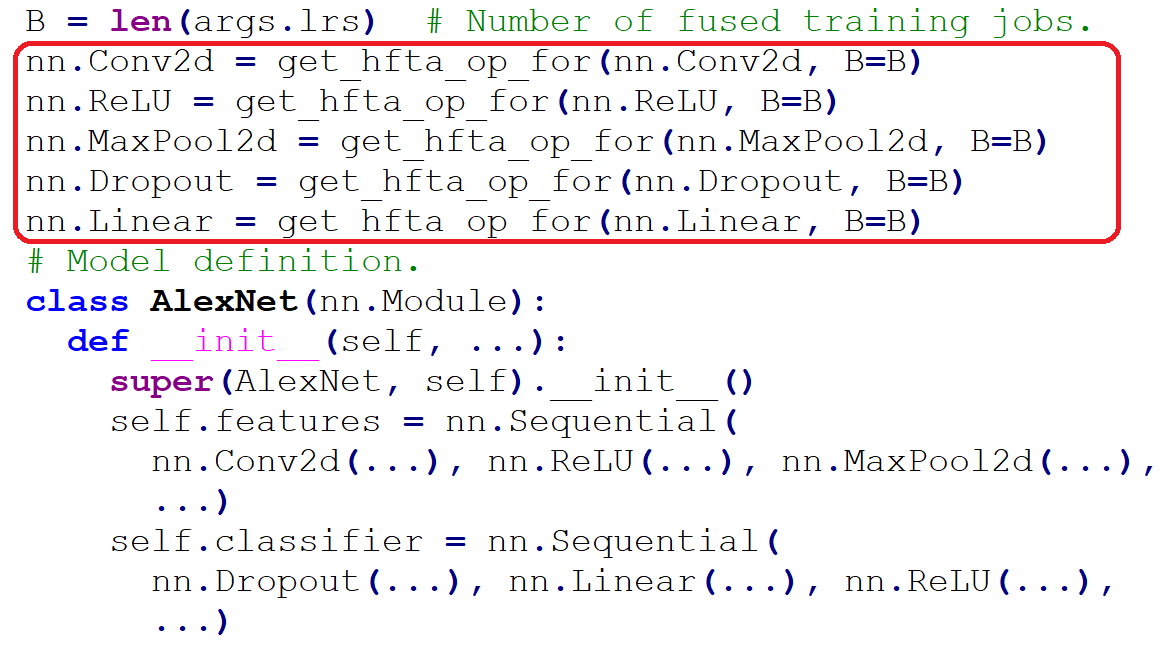} 
    \vspace{-0.9cm}
    \caption{How to enable HFTA for AlexNet.} \label{fig:alexnet_hfta}
    \vspace{-0.7cm}
\end{figure}

However, manually implementing or porting existing training workloads to the fused ones from scratch can be challenging for DL researchers and practitioners. 
To much simplify the associated engineering efforts, we develop a new DL framework library called \textbf{H}orizontally \textbf{F}used \textbf{T}raining \textbf{A}rray (\textbf{HFTA}). 
Even though we choose PyTorch~\cite{pytorch} as our prototyping DL framework due to its user-friendliness and increased popularity within the ML community~\cite{pytorch_popularity}, the same idea can be implemented on top of other DL frameworks (e.g., TensorFlow \cite{tensorflow} and MXNet \cite{mxnet}). Also, HFTA is carefully designed to accommodate computer system and architecture ``novices''. It can be used seamlessly with PyTorch-native training scripts and only requires changing very few lines of code. As an illustrative example, Figure~\ref{fig:alexnet_hfta} shows how to enable HFTA for AlexNet~\cite{alexnet}. We can observe that the model definition is kept 
%exactly 
the same with only a few extra lines of code (highlighted in the red box) to update the PyTorch's operator classes.

%\begin{listing}[]
%\inputminted[bgcolor=LightGray, baselinestretch=1.1, %fontsize=\scriptsize]{python}{python/alexnet_hfta.py}
%\vspace{-0.8cm}
%\caption{How to enable HFTA for AlexNet.}
%\label{listing:alexnet_hfta}
%\end{listing}

We now discuss the HFTA's individual components:
%(Section~\ref{sec:hfta_components}) and 
%then demonstrate both theoretically (Section~\ref{sec:loss_scale}) and empirically (Section~\ref{sec:converge})
%that HFTA has no impact on individual models' convergence.
%\hl{how HFTA can be integrated with existing hyper-parameter tuning frameworks} (Section~\ref{sec:hfht}).

%Zooming in on our design, Section~\ref{sec:hfta_op} and Section~\ref{sec:loss_scale} discuss the individual components of HFTA, and Section~\ref{sec:converge} demonstrates that HFTA reconstructs the original training processes and is free of numerical instability, therefore, does not impact convergence.

%\vspace{-0.2cm}
%\subsection{\hl{HFTA's Components}} \label{sec:hfta_components}
%\vspace{-0.2cm}

\begin{figure}[t]
    \centering
    \vspace{-0.3cm}
    \includegraphics[width=\linewidth]{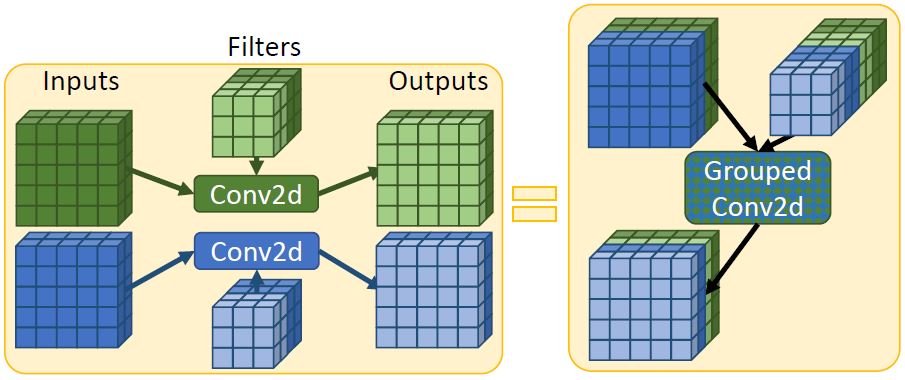} 
    \vspace{-0.9cm}
    \caption{An example showing how two Conv2d operators can be fused into a single grouped Conv2d with two groups.}
    \label{fig:conv2d_to_grouped_conv2d}
    \vspace{-0.8cm}
\end{figure}

\textbf{HFTA Operators}\hspace{0.1cm} To relieve the DL researchers and practitioners from the need to implement any horizontally fused operators themselves,
HFTA covers most common operators used in DL research and development (with detailed fusion rules provided in Appendix~\ref{append:hfta_op_fusion_rule}). 
%For example, the fusion of operators from the (de)convolution family (e.g., Conv1d or ConvTranspose2d) can be replaced by their grouped (de)convolution counterparts, and the fusion of linear layers can be replaced by the baddbmm operator.
As an example, Figure~\ref{fig:conv2d_to_grouped_conv2d} illustrates that the fusion of operators from the (de)convolution family (e.g., Conv2d or ConvTranspose1d) is equivalent to their grouped (de)convolution counterparts by concatenating (1) the inputs along the channel dimension, and (2) weights (filters) and biases along the dimension of the output channel.

%Since (1) hyper-parameter tuning is a common use case in repetitive training workloads, and (2) learning rates, learning rate schedules, and optimizer settings (e.g., momentum~\cite{sgd_with_momentum, nesterov_momentum}) are common hyper-parameters that require tuning for many DL models, HFTA supports inter-model horizontally fused optimizers (e.g., Adam~\cite{adam} and Adadelta~\cite{adadelta}) and learning rate schedulers (e.g., StepLR~\cite{steplr}) where the 

\textbf{HFTA Optimizers and Learning Rate Schedulers}\hspace{0.1cm} In addition, HFTA supports inter-model horizontally fused optimizers (e.g., Adam~\cite{adam} and Adadelta~\cite{adadelta}) and learning rate schedulers (e.g., StepLR~\cite{steplr}). 
This is because (1) hyper-parameter tuning is a common use case in repetitive training workloads, and (2) learning rates, learning rate schedules, and optimizer settings (e.g., momentum~\cite{sgd_with_momentum, nesterov_momentum}) are common hyper-parameters that require tuning for many DL models.
The scalar-vector operations (e.g., multiplying a learning rate under tuning with the gradients) in the original implementations are now replaced by broadcasted vector-vector operations (e.g., multiplying a vector of learning rates with the concatenated gradients of all models) in HFTA's implementations (as we show in Figure~\ref{fig:horizontal_fusion}). 
We also plan to continue improving the HFTA's coverage to support more operators, optimizers, and learning rate schedulers beyond 
this work's
%the 
publication\textsuperscript{\ref{footnote:hfta_github}}. 
%of this work. %, and syntax sugar for other hyper-parameters.

\textbf{Loss Scaling}\hspace{0.1cm} The loss functions across multiple models can be fused as well. 
If the loss is reduced by 
%taking the mean of 
averaging over the mini-batch, we scale the fused loss value by the number of models under fusion to reconstruct mathematically equivalent gradients.
%to reconstruct mathematically equivalent gradients when training with HFTA, the final fused loss needs to be scaled by the number of models contributing to this loss. 
If the loss is not reduced or reduced by sum, such scaling rule is no longer needed. 
We provide detailed derivations in Appendix~\ref{append:loss_scale}.

\textbf{Convergence}\hspace{0.1cm} Since the fusion of all components are achieved through mathematically equivalent transformations, HFTA theoretically does not have any effect on the models' original convergence. 
We also provide the related empirical validation in Appendix~{\ref{append:converge}}.

%\vspace{-0.2cm}
%\subsection{\hl{HFTA's Integration with Hyper-parameter Tuning Frameworks}} %\label{sec:hfht}
%\vspace{-0.2cm}
\textbf{Integration with Tuning Algorithms}\hspace{0.1cm} During hyper-parameter tuning, as a tuning algorithm proposes different sets of hyper-parameters to try, we can leverage HFTA to improve the overall hardware utilization by partitioning these sets and fusing each partition as a single training job. To prototype this idea, we develop a lightweight hyper-parameter tuning framework called \textbf{H}orizontally \textbf{F}used \textbf{H}yper-parameter \textbf{T}uning (\textbf{HFHT}). HFHT currently supports (1) two hyper-parameter tuning algorithms (Hyperband~\cite{hyperband} and random search~\cite{random_search_for_hyper_parameter_optimization}), and (2) not only sharing accelerators via HFTA but also sharing GPUs via MPS\footnote{HFHT also supports MIG. HFHT's evaluation on MIG is not included due to space constraints.} 
%and MIG 
(to compare with HFTA). We discuss HFHT's detailed design in Appendix~\ref{append:hfht}.

\vspace{-0.3cm}
\section{Methodology} \label{sec:eval_method}
\vspace{-0.2cm}

\begin{table*}[]
\centering
\vspace{-0.5cm}
\caption{Specifications of our experiment platforms. \emph{Dev. Mem.} and \emph{VM/Host Mem.} stands for device memory and VM/host memory respectively in GB. \emph{CSP} stands for cloud service provider.} \label{tab:platform_spec}
\scriptsize
\begin{tabularx}{\linewidth}{l|p{1.15cm}p{0.3cm}p{1.5cm}p{0.5cm}p{0.5cm}p{1.2cm}p{1.8cm}p{1.5cm}p{1cm}p{2.0cm}}
\toprule

Accelerator & Dev. Mem. & CSP & VM Instance   & CUDA & cuDNN & GPU Driver & PyTorch         & PyTorch/XLA & (v)CPUs & VM/Host Mem. \\ \midrule
V100        & 16                 & AWS                    & p3.2xlarge    & 10.2 & 7.6.5 & 450.51.05  & 1.6.0           & -           & 8     & 61             \\
RTX6000     & 24                 & -                      & -             & 10.2   & 7.6.5 & 450.66     & 1.6.0           & -           & 8     & 16             \\
A100        & 40                 & GCP                    & a2-highgpu-1g & 11.0.3   & 8.0.2 & 450.51.06  & 1.7.0a0+8deb4fe       & -           & 12    & 85             \\
TPU v3      & 16                 & GCP                    & n1-highmem-8  & -    & -     & -          & 1.7.0a0+626e410 & 1.6+8af57fb & 8     & 52   \\

\bottomrule
\end{tabularx}\vspace{-0.6cm}
\end{table*}

\textbf{Workloads}\hspace{0.1cm}
Our major benchmarks are carefully selected based on the following three criteria. 
First, our workloads should represent important models in their corresponding DL sub-fields, making sure that HFTA is effective in improving the hardware utilization for important DL models. 
Second, we select models that have not yet received much attention from the computer system and architecture communities and 
%hence 
are not over-optimized.
%(as opposed to the conventional DL models used in MLPerf Training Benchmark suite~\cite{mlperf_train} that typically have high hardware utilization). 
This is a much more realistic scenario for DL researchers and practitioners who typically lack the expertise 
%in computer system and architecture 
to apply advanced optimization techniques. Third, we would like to cover both compute-bound and memory-bound DL models. Based on the aforementioned criteria, two classes of models (three different workloads) are selected as our major benchmarks.

\emph{PointNet}~\cite{pointnet} is a memory-bound neural network that performs (i) object classification and (ii) segmentation tasks on 3D point clouds. 
The models for both tasks are trained on the ShapeNet part dataset~\cite{shapenet_part}. 
We leverage a third-party PyTorch implementation of PointNet~\cite{pointnet_pytorch}
%that is 
endorsed by Qi et al.~\cite{pointnet_tf}.

%\vspace{-0.2cm}
\emph{DCGAN}~\cite{dcgan} is a compute-bound generative adversarial network (GAN) that synthesizes natural-apparent images. The model is trained on the LSUN dataset \cite{lsun}. We leverage an implementation of DCGAN from PyTorch official examples~\cite{pytorch_examples}.

To emulate the hardware usage habits of DL researchers and practitioners without the influence from the computer system and architecture experts, the batch sizes
used in both benchmarks are kept the same as reported in their corresponding publications. 

To empirically validate that HFTA does not affect convergence and to demonstrate that HFTA can also improve the hardware utilization for conventional models, we further include ResNet-18~\cite{resnet}, MobileNetV3-Large~\cite{mobilenetv3l}, Transformer~\cite{transformer}, and BERT-Medium~\cite{bert_medium} as our secondary benchmarks. 
To evaluate HFHT, for each of the the PointNet and MobileNet classification tasks, we prepare two end-to-end hyper-parameter tuning workloads using different tuning algorithms (random search and Hyperband) as our benchmarks. 
All four workloads aim at maximizing the validation accuracy on their corresponding datasets. Each workload tunes eight independent hyper-parameters. 
%that are either discrete or continuous. 
We discuss the detailed setup of these benchmarks in Appendix~\ref{append:secondary_eval_method}.

%train ResNet-18~\cite{resnet} on V100 with the CIFAR-10~\cite{cifar10} dataset using Adadelta~\cite{adadelta} with a batch size of 1000.

\vspace{-0.05cm}
\textbf{Experimental Setup}\hspace{0.1cm} 
Our experiments are performed on two types of ML accelerators (NVIDIA GPUs and Google TPUs) including the most recent three generations of GPUs and the latest available generation of TPUs: (i) Volta-based V100~\cite{v100}, (ii) Turing-based RTX6000~\cite{rtx6000}, and (iii) very recent Ampere-based A100~\cite{a100},\footnote{Using GCP A2 Alpha version instances.} (iv) TPU v3~\cite{cloud_tpu}. We provide the detailed specifications in Table~\ref{tab:platform_spec}.

\vspace{-0.05cm}
\textbf{Baselines}\hspace{0.1cm} We use \emph{hyper-parameter tuning} (including learning rate, learning rate schedule, and optimizer settings) as the use case for our repetitive single-accelerator training jobs under experimentation. 
We compare HFTA with the following four SOTA baselines. (1) \emph{Serial}: each training job is executed %end-to-end 
on a single accelerator. 
This scheme is employed by most hyper-parameter tuning frameworks~\cite{wandb_sweep, hyperparameter_tuning_common_practice}. 
(2) \emph{Concurrent}: multiple training jobs are executed as independent processes on the same GPU.
In this case, the kernels from different processes are \emph{time-multiplexed}, but can \emph{not} execute concurrently on the same GPU (without the help of MPS or other hardware features). 
This scheme is used when MPS is not preferable due to reasons related to infrastructure and/or security 
%related reasons 
(e.g., custom-built infrastructure or CUPTI tools that are not compatible with MPS). 
(3) \emph{MPS}: similar to \emph{concurrent}, except the independent processes are executed via MPS. 
(4) \emph{MIG}: similar to \emph{concurrent}, except the independent processes are executed via MIG. 
This scheme is currently only available on the A100 GPUs.
We use \emph{concurrent}, \emph{MPS}, and \emph{MIG} only on GPUs
%(where applicable), 
since TPUs 
%have no hardware support for concurrently running 
do not support running concurrent processes as of now.
We do not evaluate
the major related prior work, HiveMind~\cite{hivemind},  
since it is both close-sourced and implemented on a different ML framework (TensorFlow). 
We provide the detailed qualitative comparison against HiveMind in Section~\ref{sec:prior_works}.

\vspace{-0.05cm}
\textbf{Metrics}\hspace{0.1cm} We use the \emph{per-device training throughput} as our key performance metric to compare HFTA against our baselines since HFTA has no impact on the model convergence. 
We calculate this throughput by measuring the end-to-end training latency of
%: 
(i) 10 epochs for both classification and segmentation tasks on PointNet; and (2) 5 epochs, 1000 iterations per epoch on DCGAN (enough for these workloads to enter the execution steady state). 
We skip the first epoch on GPUs and the first two epochs on TPUs to 
%properly 
warm up the hardware properly before making any measurements. 
We repeat each experiment 
%at least 
three times and report the average, minimum, and maximum per experiment.

To measure the effect of each technique on the hardware utilization, we use the \texttt{sm\_active} and \texttt{sm\_occupancy} performance counters that represent the SM temporal and spatial utilization, respectively. 
We also use the \texttt{tensor\_active} performance counter to measure the TC temporal utilization \cite{dcgm_field_identifiers}. 
Details on these performance counters can be found in Appendix~\ref{append:dcgm}. 

%\textbf{Floating-point (FP) Format}\hspace{0.1cm} In order to evaluate how effective HFTA utilizes TCs, we run the benchmarks in both FP32 and native PyTorch Automatic Mixed Precision (AMP) \cite{pytorch_amp} on GPUs. We run the TPU v3 experiments with the default bfloat16 \cite{bfloat16} FP format.

\vspace{-0.3cm}
\section{Evaluation} \label{sec:eval}
\vspace{-0.2cm}
Our evaluation results on the major benchmarks (i.e., PointNet classification task, PointNet segmentation task, and DCGAN) are thoroughly analyzed here, including end-to-end training performance on GPUs (Section~\ref{sec:eval_gpus}) and TPUs (Section~\ref{sec:eval_tpus}). 
We also analyze key GPU hardware performance counters to explain why HFTA achieves significantly better training performance (Section~\ref{sec:gpu_perf_analysis}). 
We discuss the results on our secondary benchmarks (i.e., ResNet-18, MobileNetV3-Large, Transformer, and BERT-Medium) in Appendix~\ref{append:secondary_eval}. 
Finally, we showcase HFTA's potential to significantly improve the hardware utilization for existing 
hyper-parameter 
tuning algorithms (Section~\ref{sec:hfht_cost_saving}).

\vspace{-0.3cm}
\subsection{End-to-end Training Performance on GPUs} \label{sec:eval_gpus}
\vspace{-0.2cm}

\begin{figure*}[t]
    \centering
    \vspace{-0.2cm}
    \begin{subfigure}[]{\linewidth}
        \centering
        \includegraphics[width=0.9\linewidth]{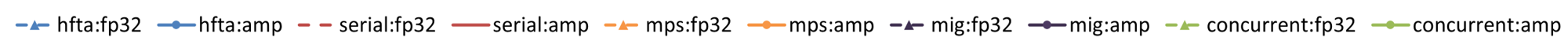}
        \vspace{-0.2cm}
    \end{subfigure}
    \hfill
    \begin{subfigure}[]{0.32\linewidth}
        \centering
        \includegraphics[width=0.9\linewidth]{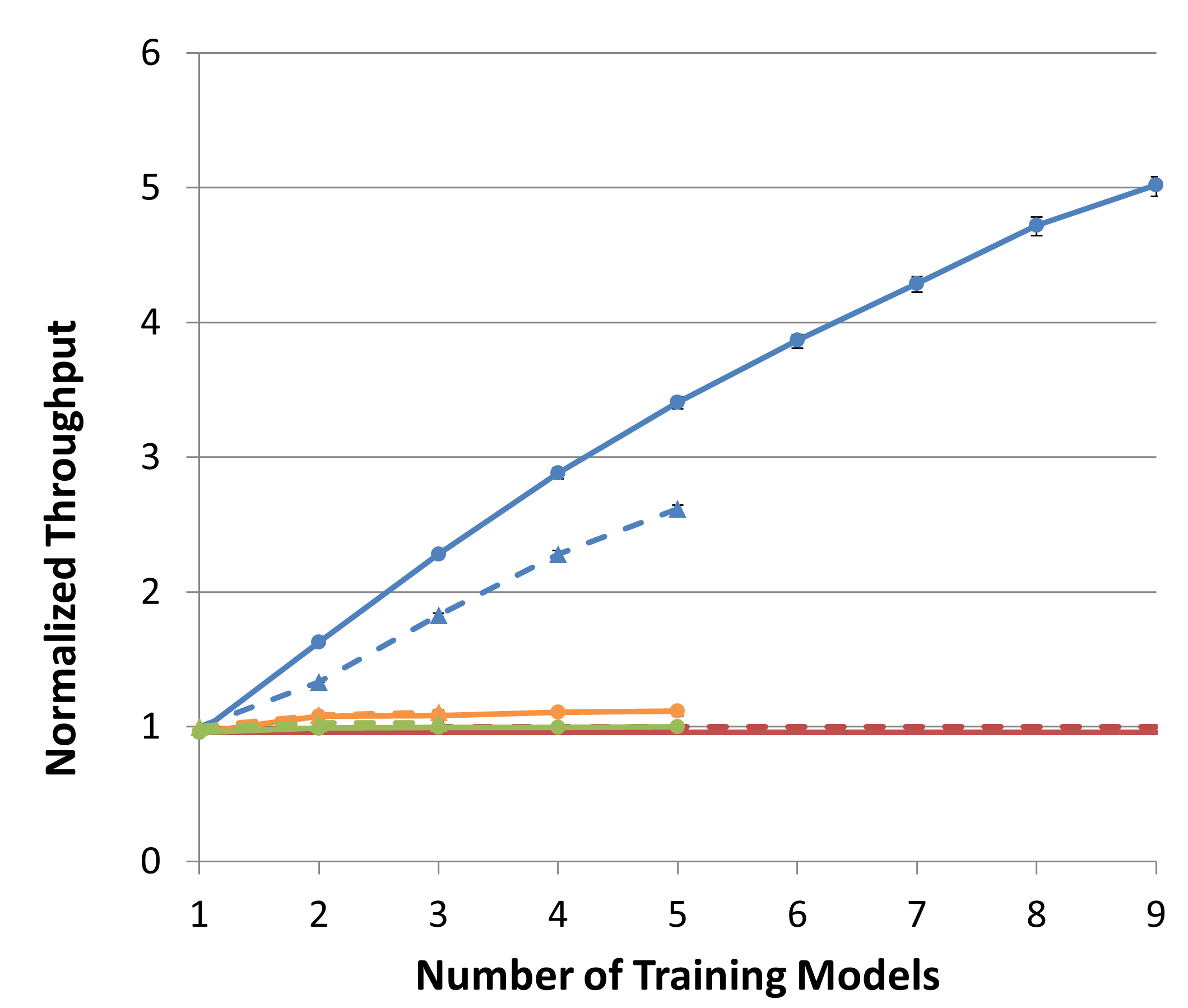}
        \vspace{-0.2cm}
        \caption{PointNet Classification on V100}
        \label{fig:speedups_pointnet_cls_v100}
    \end{subfigure}
    \hfill
    \begin{subfigure}[]{0.32\linewidth}
        \centering
        \includegraphics[width=0.9\linewidth]{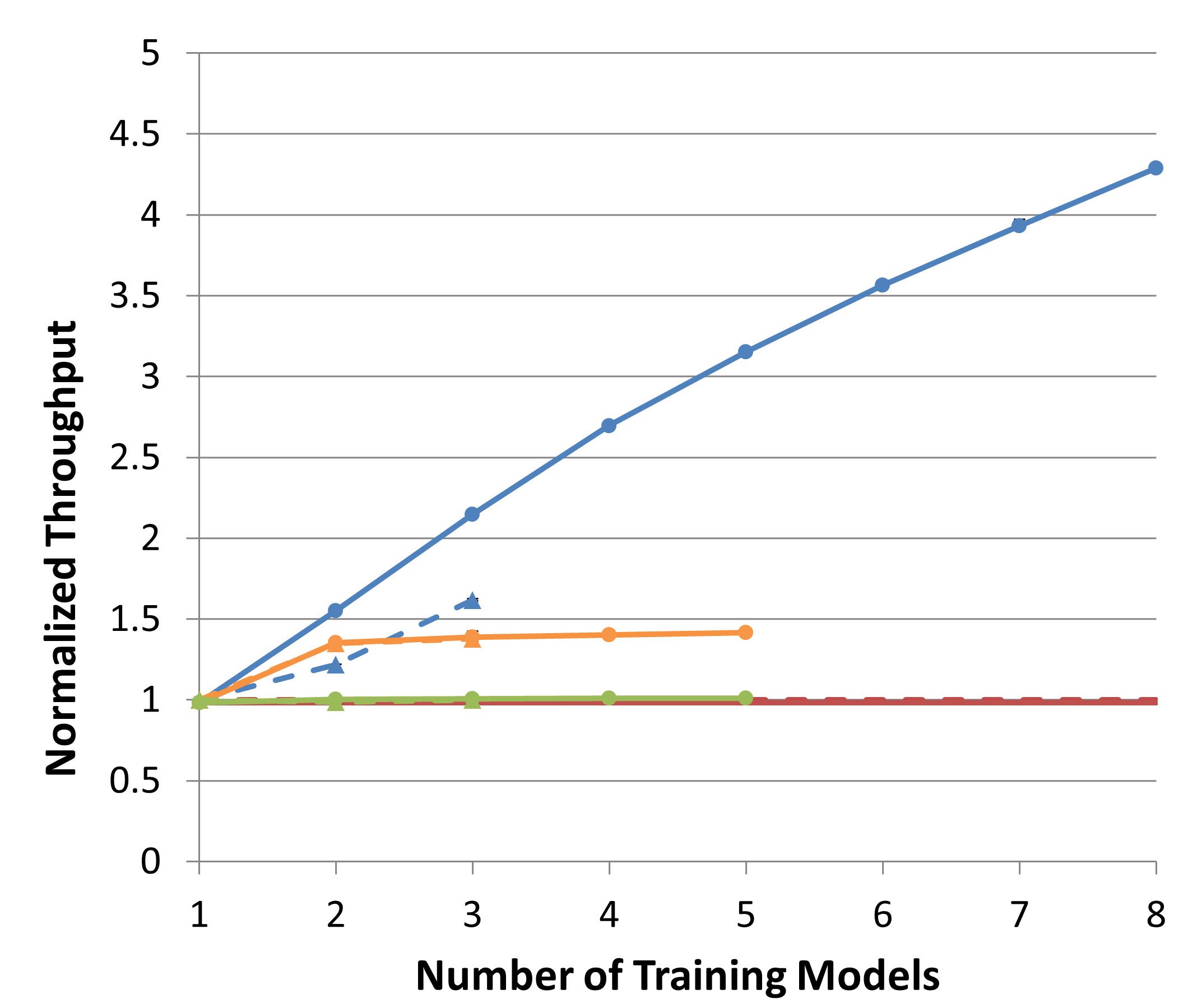}
        \vspace{-0.2cm}
        \caption{PointNet Segmentation on V100}
        \label{fig:speedups_pointnet_seg_v100}
    \end{subfigure}
    \hfill
    \begin{subfigure}[]{0.32\linewidth}
        \centering
        \includegraphics[width=0.9\linewidth]{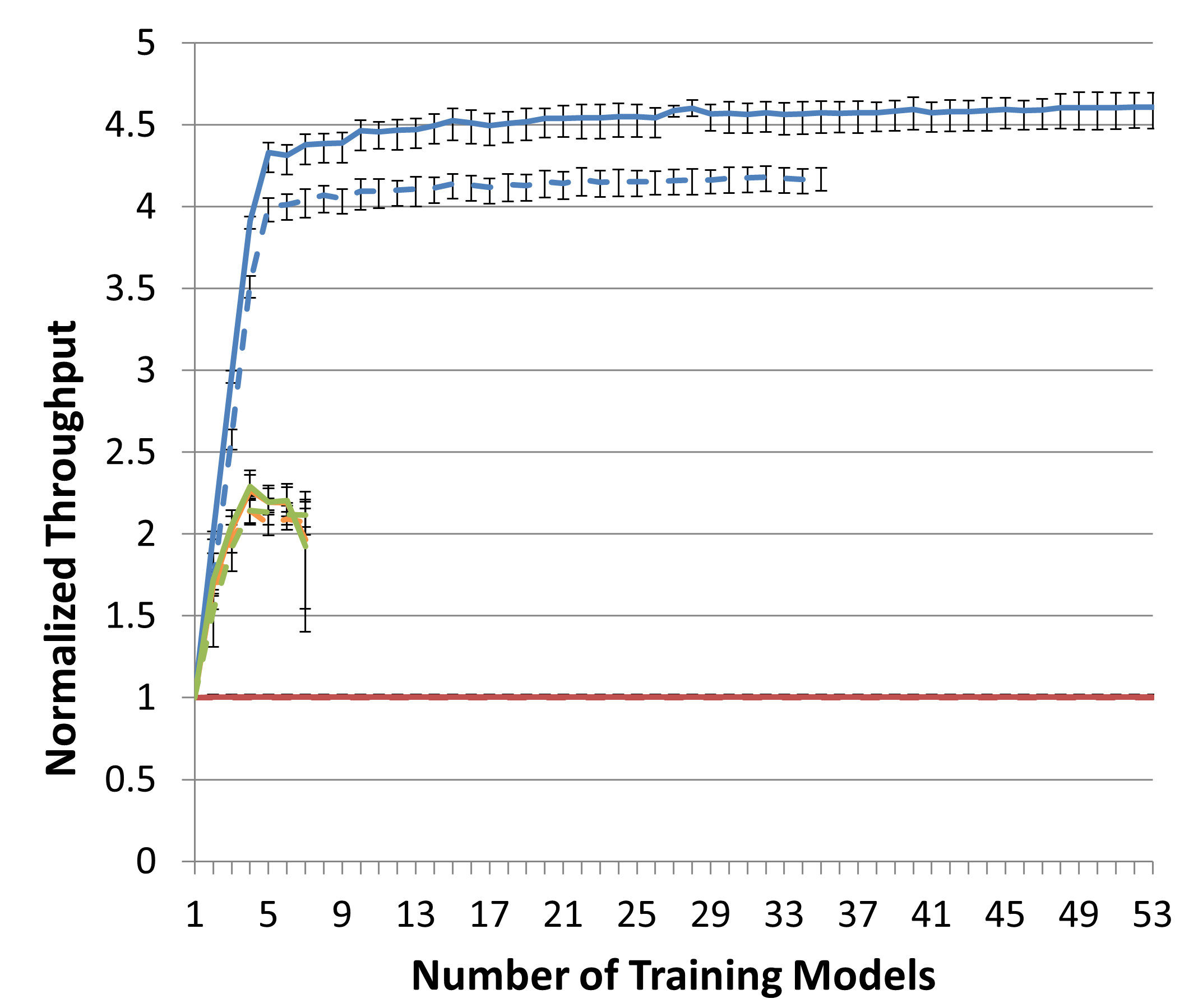}
        \vspace{-0.2cm}
        \caption{DCGAN on V100}
        \label{fig:speedups_dcgan_v100}
    \end{subfigure}
    \hfill
    \begin{subfigure}[]{0.32\linewidth}
        \centering
        \includegraphics[width=0.9\linewidth]{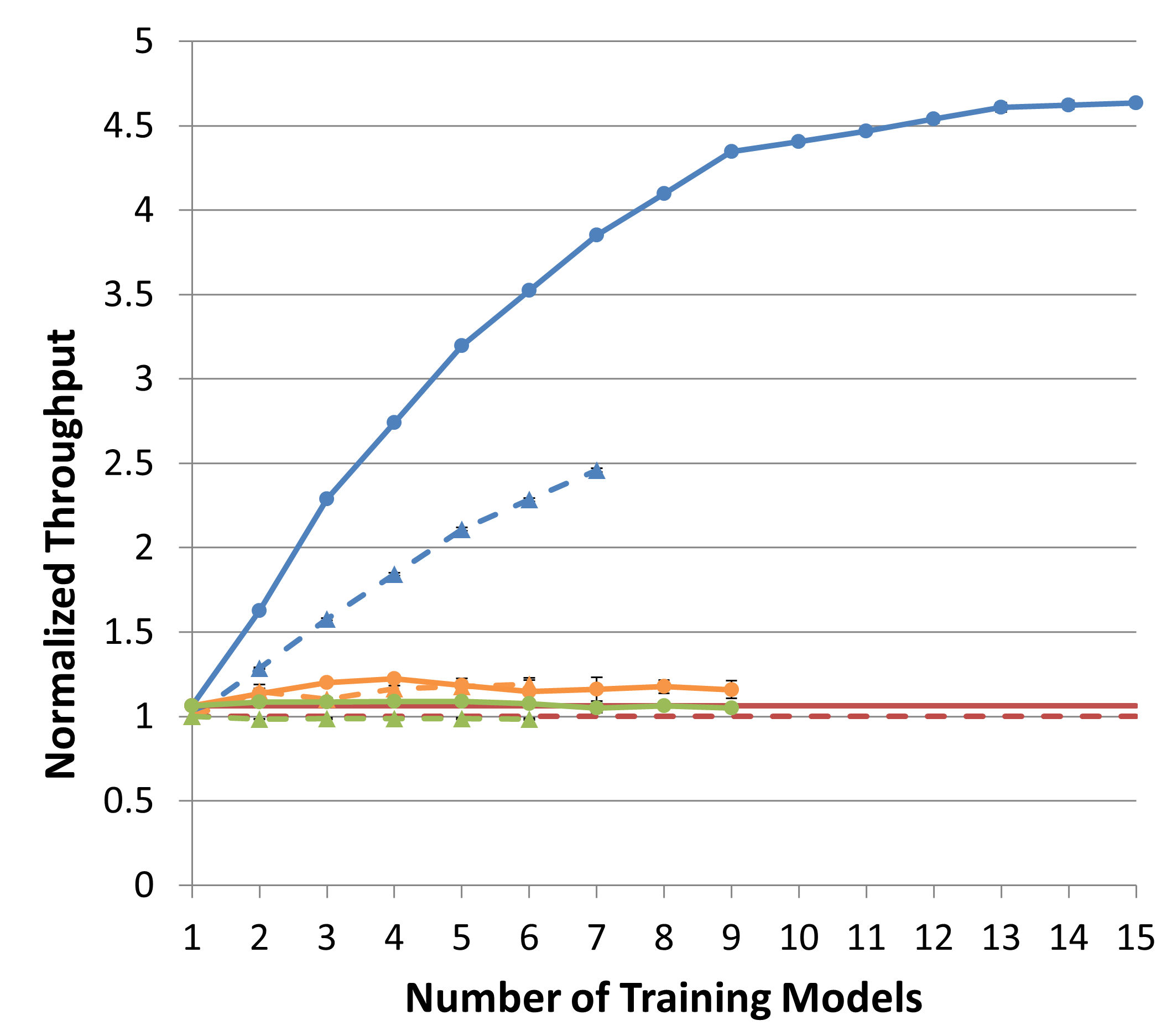}
        \vspace{-0.2cm}
        \caption{PointNet Classification on RTX6000}
        \label{fig:speedups_pointnet_cls_rtx6000}
    \end{subfigure}
    \hfill
    \begin{subfigure}[]{0.32\linewidth}
        \centering
        \includegraphics[width=0.9\linewidth]{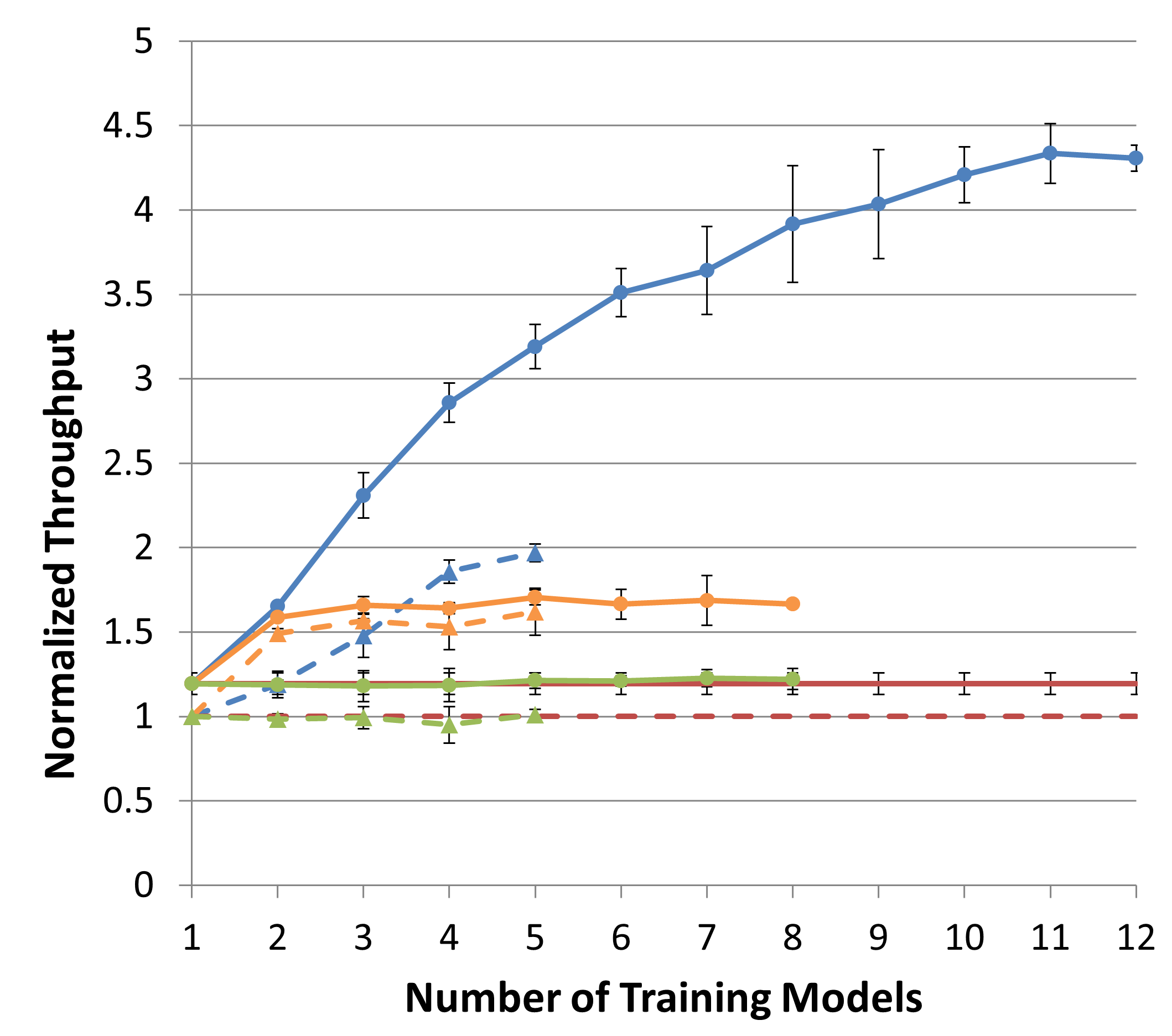}
        \vspace{-0.2cm}
        \caption{PointNet Segmentation on RTX6000}
        \label{fig:speedups_pointnet_seg_rtx6000}
    \end{subfigure}
    \hfill
    \begin{subfigure}[]{0.32\linewidth}
        \centering
        \includegraphics[width=0.9\linewidth]{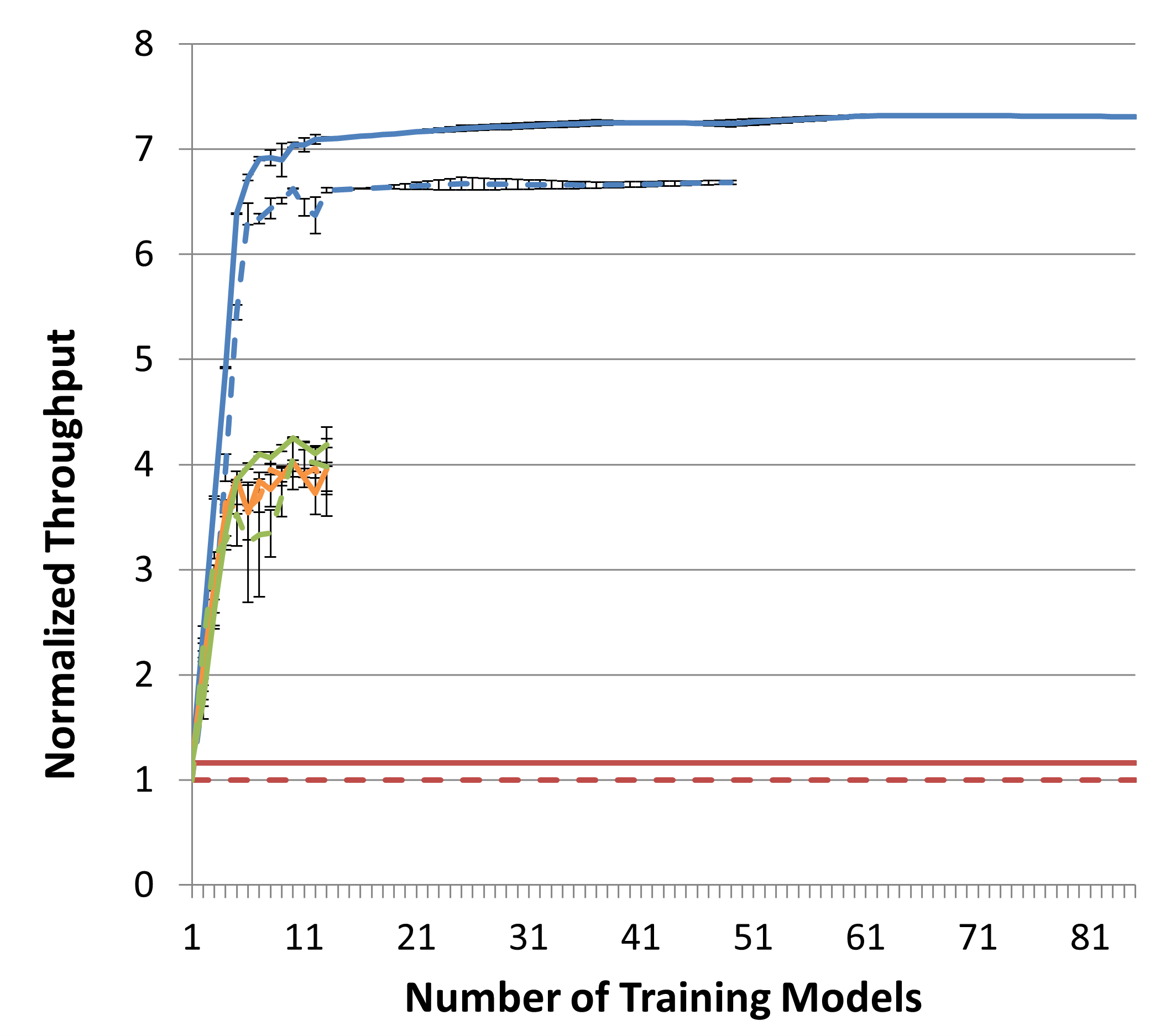}
        \vspace{-0.2cm}
        \caption{DCGAN on RTX6000}
        \label{fig:speedups_dcgan_rtx6000}
    \end{subfigure}
    \hfill
    \begin{subfigure}[]{0.32\linewidth}
         \centering
        \includegraphics[width=0.9\linewidth]{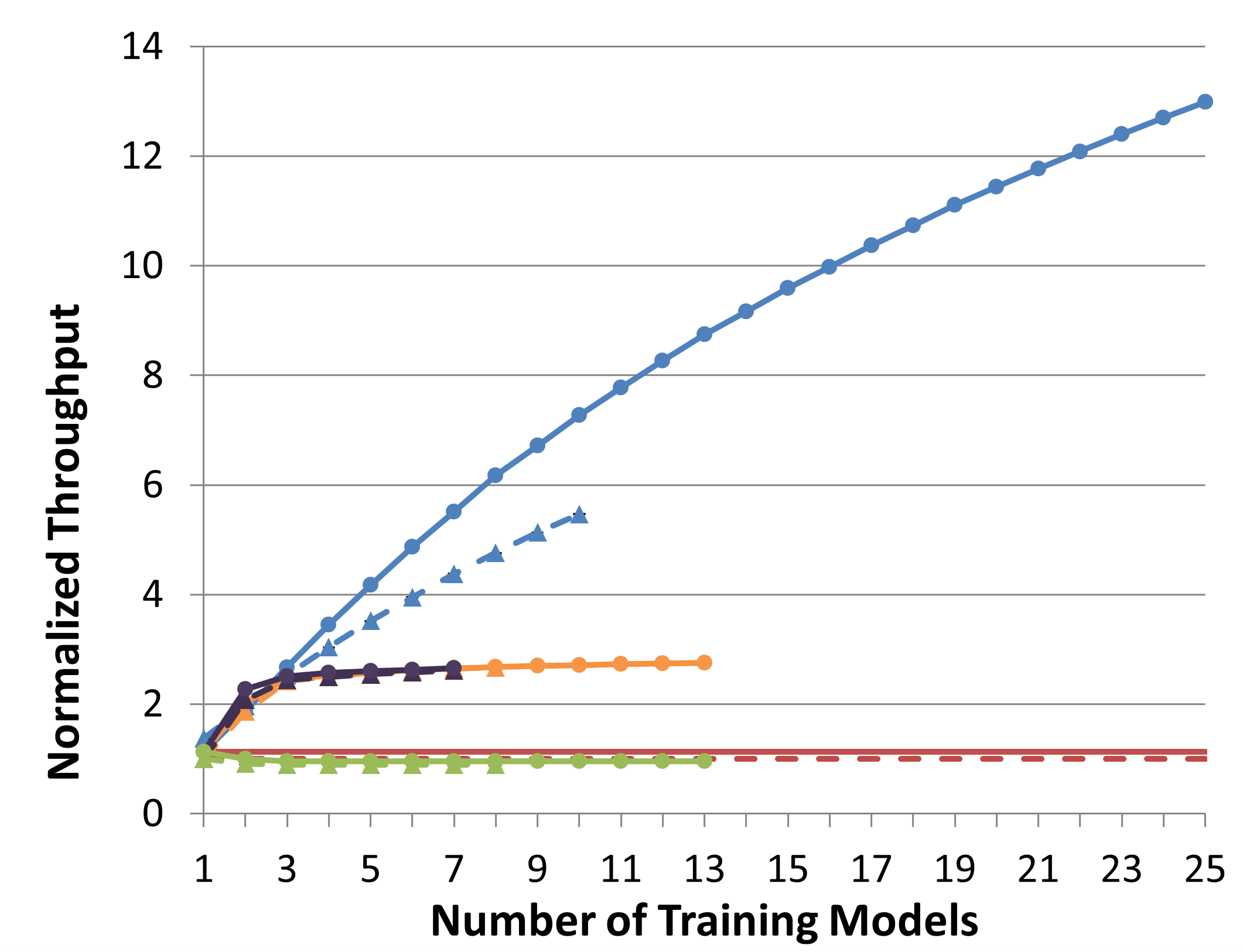}
        \vspace{-0.2cm}
        \caption{PointNet Classification on A100}
        \label{fig:speedups_pointnet_cls_a100}
    \end{subfigure}
    \hfill
    \begin{subfigure}[]{0.32\linewidth}
        \centering
        \includegraphics[width=0.9\linewidth]{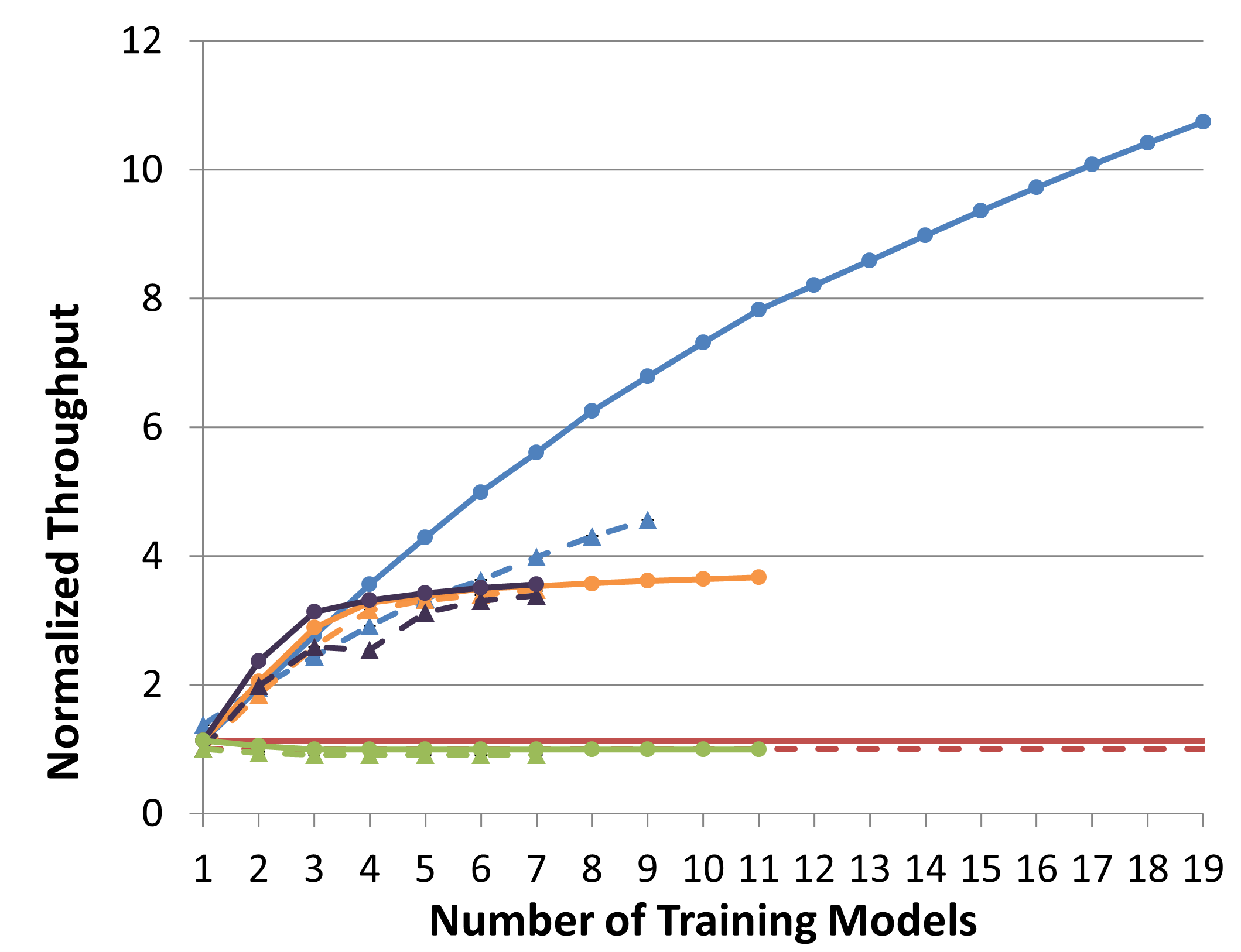}
        \vspace{-0.2cm}
        \caption{PointNet Segmentation on A100}
        \label{fig:speedups_pointnet_seg_a100}
    \end{subfigure}
    \hfill
    \begin{subfigure}[]{0.32\linewidth}
        \centering
        \includegraphics[width=0.9\linewidth]{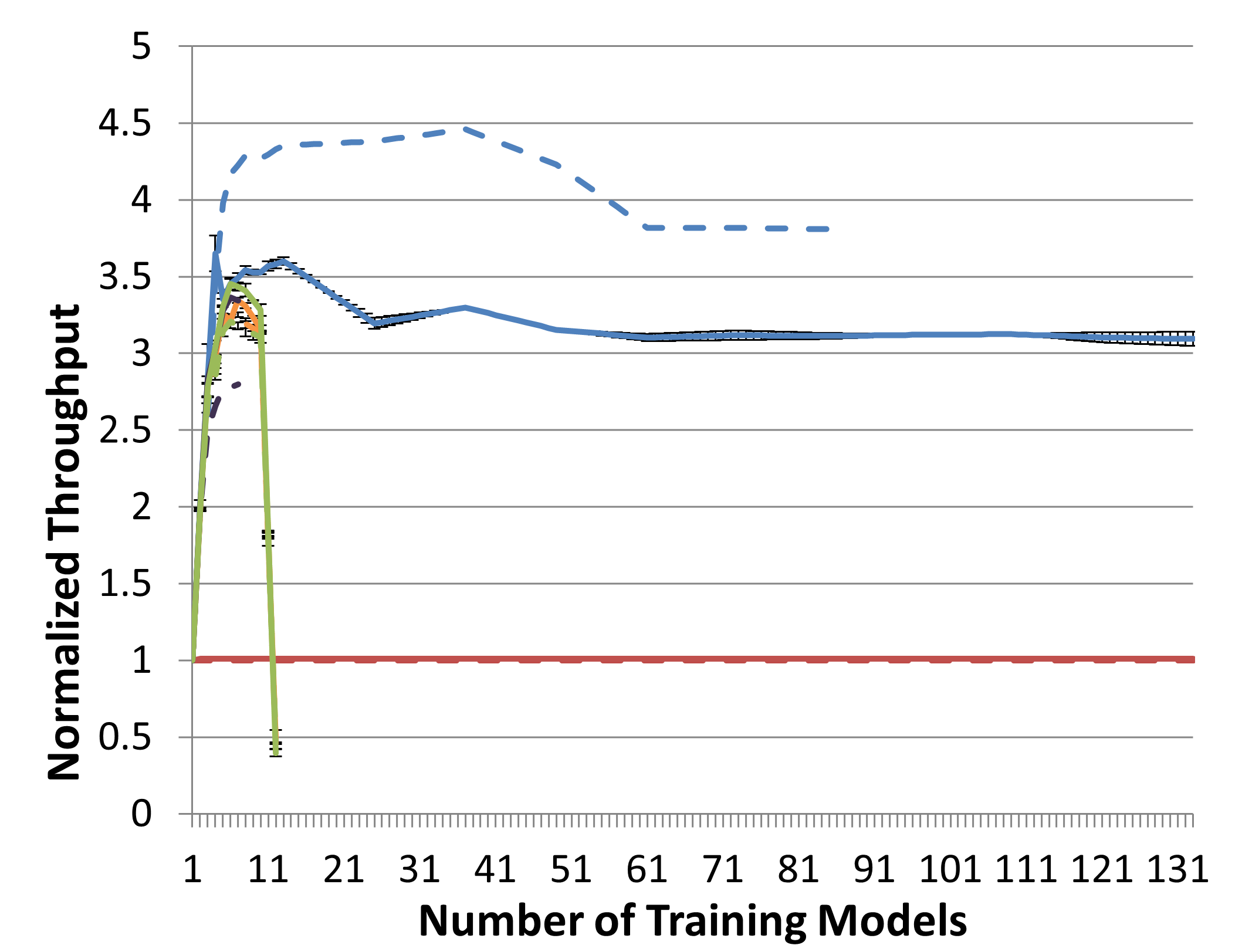}
        \vspace{-0.2cm}
        \caption{DCGAN on A100}
        \label{fig:speedups_dcgan_a100}
    \end{subfigure}
    \vspace{-0.2cm}
    \caption{The normalized training throughput as we increase the number of models sharing the same GPU.}
    \label{fig:speedups_gpus}
    \vspace{-0.5cm}
\end{figure*}

\textbf{V100 Results}\hspace{0.1cm}
To compare the HFTA's end-to-end training performance with other alternatives (i.e., \emph{serial}, \emph{concurrent}, \emph{MPS}), Figure~\ref{fig:speedups_pointnet_cls_v100}, \ref{fig:speedups_pointnet_seg_v100} and \ref{fig:speedups_dcgan_v100} plot the per-GPU normalized training throughput on the V100 GPUs (Volta architecture~\cite{volta}) with the PointNet classification task, PointNet segmentation task, and DCGAN respectively. 
We normalize the throughput for each experiment by the respective FP32 \emph{serial} baseline. 
%For each experiment, w
We show both FP32 and AMP~\cite{pytorch_amp} training results for each experiment. 
Each curve 
%shows the normalized throughput
grows as we increase the number of models that either co-run together (for the \emph{concurrent} and \emph{MPS} baselines) or run in the fused form with HFTA. Each curve ``stops'' when it reaches the maximum number of models before the GPU runs out of memory. Based on these figures, we make several major observations:

%experiment
%\begin{rlscope-finding}{find:v100_peak_throughput}
%HFTA achieves significantly higher peak throughput than all baselines (up to %$\gpuvAmpPeakHFTAvsSerialCls \times$).
%\end{rlscope-finding}

First, \emph{HFTA achieves significantly higher peak throughput than all baselines}; specifically, $\gpuvMergePeakHFTAvsSerialSeg \times$ to $\gpuvMergePeakHFTAvsSerialCls \times$ over \emph{serial}, $\gpuvMergePeakHFTAvsConcurrentDCGAN \times$ to $\gpuvMergePeakHFTAvsConcurrentCls \times$ over \emph{concurrent} and $\gpuvMergePeakHFTAvsMPSDCGAN \times$ to $\gpuvMergePeakHFTAvsMPSCls \times$ over \emph{MPS}. The significant throughput improvement is due to a much higher achieved utilization in both compute cores (details in Section~\ref{sec:gpu_perf_analysis}) and GPU memory (discussed in the next observation).

Second, \emph{HFTA enables more models to share the same GPU than MPS and concurrent}; specifically, up to $\numWorkloadsPNClsUpperRange \times$ on the PointNet classification task, up to $\numWorkloadsPNSegUpperRange \times$ on the segmentation task and up to $\numWorkloadsDCGANUpperRange \times$ on DCGAN. This is because HFTA does not duplicate the GPU memory overhead as we explain in Section~\ref{sec:gpu_perf_analysis}.

Third, \emph{as we increase the number of models sharing the same GPU, the throughput of HFTA scales up and, in some cases, plateaus eventually}. 
This is because using HFTA, the SM and TC utilization increases with the number of co-executing models (as we explain in Section~\ref{sec:gpu_perf_analysis}). 
In contrast, \emph{MPS} and \emph{concurrent} either (i) plateau at a smaller number of models with a lower throughput, as we observe in Figure~\ref{fig:speedups_pointnet_cls_v100} and \ref{fig:speedups_pointnet_seg_v100}, or (ii) even experience performance degradation as we observe in Figure~\ref{fig:speedups_dcgan_v100} due to host resource (e.g., CPUs, disk I/O bandwidth, and/or memory) contention among many training processes.

Fourth, \emph{even with the same number of models sharing the same GPU, HFTA often achieves higher throughput than all baselines}. The maximum speedups range from $\gpuvFpSameBHFTAvsConcurrentSeg \times$ to $\gpuvAmpSameBHFTAvsConcurrentCls \times$ over \emph{concurrent} and $\gpuvFpSameBHFTAvsMPSSeg \times$ to $\gpuvAmpSameBHFTAvsMPSCls \times$ over \emph{MPS}. 

Fifth, \emph{HFTA can better exploit computing power from advanced hardware features such as TCs used during AMP training compared to the baselines}. Specifically, the maximum speedup of AMP training over FP32 is $\maxGpuAmpvsFpHFTA \times$ with HFTA, but only $\maxGpuAmpvsFpSerial \times$ for \emph{serial}, $\maxGpuAmpvsFpConcurrent \times$ for \emph{concurrent}, and $\maxGpuAmpvsFpMPS \times$ for \emph{MPS}.  

Therefore, we conclude that HFTA can significantly outperform major hardware-based sharing alternatives in improving hardware utilization and, as a result, improve the throughput of emerging ML models during repetitive single-accelerator training.

\textbf{RTX6000 and A100 Results}\hspace{0.1cm} 
To check whether HFTA's significant performance gains are general across different GPU architectures (e.g., Turing~\cite{turing} and Ampere~\cite{ampere}), we conduct the same set of experiments on the RTX6000 (Figure~\ref{fig:speedups_pointnet_cls_rtx6000}, \ref{fig:speedups_pointnet_seg_rtx6000} and \ref{fig:speedups_dcgan_rtx6000} respectively) and the A100 (Figure~\ref{fig:speedups_pointnet_cls_a100}, \ref{fig:speedups_pointnet_seg_a100} and \ref{fig:speedups_dcgan_a100} respectively) while adding the extra \emph{MIG} baseline for the A100. The general trends in these figures are similar to those we observe for V100. To simplify the comparison, for each workload on each GPU, Table~\ref{tab:gpus_peak_speedups} presents the peak throughput speedups of HFTA over the baselines, while Appendix~\ref{append:additional_eval_data} presents (i) the maximum throughput speedups of HFTA over the baselines given a fixed number of models, and (ii) the maximum AMP training throughput speedups over FP32 for both HFTA and the baselines. In addition, we make the following new observations:

First, both RTX6000 and A100 have higher GPU memory (HBM) capacities than V100 (24 GB and 40 GB vs. 16 GB); therefore, \emph{both HFTA and the baselines can co-run more models on the same RTX6000/A100 compared with V100}. 
For example, AMP training of the PointNet classification task via HFTA can run up to 15/25 models on RTX6000/A100 vs. 9 on V100. 

Second, since A100 has more compute capability and a larger GPU memory capacity than V100, the comparison of Figure~\ref{fig:speedups_pointnet_cls_a100} vs. \ref{fig:speedups_pointnet_cls_v100} and \ref{fig:speedups_pointnet_seg_a100} vs. \ref{fig:speedups_pointnet_seg_v100} reveals that \emph{HFTA not only fits more models on the same hardware, but also achieves a higher peak throughput speedup over the baselines on A100 than on V100} (e.g., for PointNet segmentation task, the peak throughput speedup over serial is as high as $\gpuaAmpPeakHFTAvsSerialSeg \times$ on A100 vs. $\gpuvAmpPeakHFTAvsSerialSeg \times$ on V100).

Third, we observe one anomaly in DCGAN training on A100 (Figure~\ref{fig:speedups_dcgan_a100}) where HFTA's FP32 throughput is higher than that of AMP. After profiling the AMP run of this experiment via the PyProf~\cite{pyprof} tool, we pinpoint a few suspicious cuDNN-related FP32 kernels (which are supposed to be replaced by the equivalent TC kernels) in the backward pass. Since the Ampere architecture and the corresponding versions of cuDNN/PyTorch are very recently released, and we do not observe similar problems on older cuDNN/PyTorch versions for V100 and RTX6000, we believe that this issue is temporary due to the insufficient optimization in some of the new cuDNN kernels for A100. We hope it will be addressed in future cuDNN releases/fixes and we will be able to update the results accordingly. 
%will ``disappear" with future cuDNN releases whose follow-up is left for future work.

Fourth, we notice that \emph{on A100, the MIG partitioning (only up to 7 GIs) can be too coarse-grained}. As we observe in Figure~\ref{fig:speedups_pointnet_cls_a100}, \ref{fig:speedups_pointnet_seg_a100} and \ref{fig:speedups_dcgan_a100}, both \emph{MPS} and \emph{concurrent} could often share the A100 with more than seven models.

Therefore, we conclude that HFTA's performance generally scales well with the compute and memory capabilities of modern GPUs. 
We observe higher performance benefits in the newer GPU architectures that would otherwise suffer more significantly from the hardware under-utilization when training without HFTA (as we qualitatively discuss in Section~\ref{sec:inefficiency_repetitive_training_workloads} and empirically show in Appendix~\ref{append:additional_eval_data}).

% Peak throughput table
\begin{table}[]
    \centering
    \vspace{-0.5cm}
    \caption{The peak training throughput speedups of HFTA over the baselines. For each experiment, the higher throughput between FP32 and AMP is used in the calculation. The detailed breakdown between FP32 and AMP is included in Appendix~\ref{append:additional_eval_data}.} \label{tab:gpus_peak_speedups}
    \scriptsize
    \begin{tabular}{cl|ccc}
\toprule
\multicolumn{2}{l|}{Benchmark}        & \multicolumn{1}{l}{\begin{tabular}[c]{@{}l@{}}PointNet \\ Classification\end{tabular}} & \multicolumn{1}{l}{\begin{tabular}[c]{@{}l@{}}PointNet \\ Segmentation\end{tabular}} & \multicolumn{1}{l}{DCGAN}           \\ \midrule
\multirow{3}{*}{V100}    & \emph{serial}     & \gpuvMergePeakHFTAvsSerialCls                                                & \gpuvMergePeakHFTAvsSerialSeg                                                & \gpuvMergePeakHFTAvsSerialDCGAN     \\
                         & \emph{concurrent} & \gpuvMergePeakHFTAvsConcurrentCls                                            & \gpuvMergePeakHFTAvsConcurrentSeg                                            & \gpuvMergePeakHFTAvsConcurrentDCGAN \\
                         & \emph{MPS}        & \gpuvMergePeakHFTAvsMPSCls                                                   & \gpuvMergePeakHFTAvsMPSSeg                                                   & \gpuvMergePeakHFTAvsMPSDCGAN        \\ \hline
\multirow{3}{*}{RTX6000} & \emph{serial}     & \gpurMergePeakHFTAvsSerialCls                                                & \gpurMergePeakHFTAvsSerialSeg                                                & \gpurMergePeakHFTAvsSerialDCGAN     \\
                         & \emph{concurrent} & \gpurMergePeakHFTAvsConcurrentCls                                            & \gpurMergePeakHFTAvsConcurrentSeg                                            & \gpurMergePeakHFTAvsConcurrentDCGAN \\
                         & \emph{MPS}        & \gpurMergePeakHFTAvsMPSCls                                                   & \gpurMergePeakHFTAvsMPSSeg                                                   & \gpurMergePeakHFTAvsMPSDCGAN        \\ \hline
\multirow{4}{*}{A100}    & \emph{serial}     & \textbf{\gpuaMergePeakHFTAvsSerialCls}                                                & \gpuaMergePeakHFTAvsSerialSeg                                                & \gpuaMergePeakHFTAvsSerialDCGAN     \\
                         & \emph{concurrent} & \textbf{\gpuaMergePeakHFTAvsConcurrentCls}                                            & \gpuaMergePeakHFTAvsConcurrentSeg                                            & \gpuaMergePeakHFTAvsConcurrentDCGAN \\
                         & \emph{MPS}        & \textbf{\gpuaMergePeakHFTAvsMPSCls}                                                   & \gpuaMergePeakHFTAvsMPSSeg                                                   & \gpuaMergePeakHFTAvsMPSDCGAN        \\
                         & \emph{MIG}        & \textbf{\gpuaMergePeakHFTAvsMIGCls}                                                   & \gpuaMergePeakHFTAvsMIGSeg                                                   & \gpuaMergePeakHFTAvsMIGDCGAN        \\ 
\bottomrule
\end{tabular}
\vspace{-0.8cm}
\end{table}

\begin{figure*}[t]
    \centering
    \vspace{-0.3cm}
    \begin{minipage}{0.55\linewidth}
        \begin{subfigure}[]{0.5\linewidth}
            \centering
            \includegraphics[width=\linewidth]{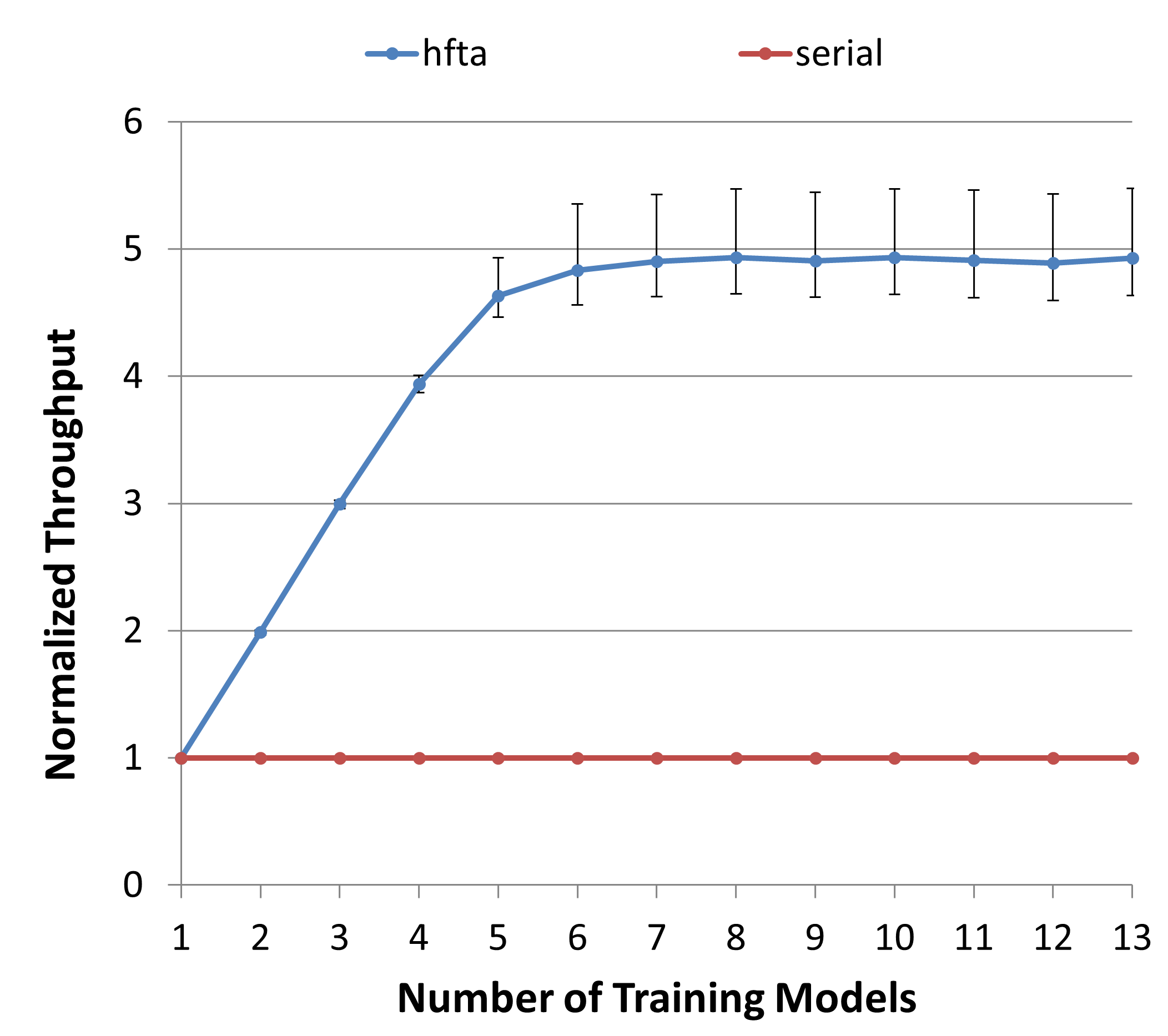}
            \vspace{-0.6cm}
            \caption{PointNet Classification}
            \label{fig:speedups_pointnet_cls_tpu_v3}
        \end{subfigure}
        \hfill
        \begin{subfigure}[]{0.5\linewidth}
            \centering
            \includegraphics[width=\linewidth]{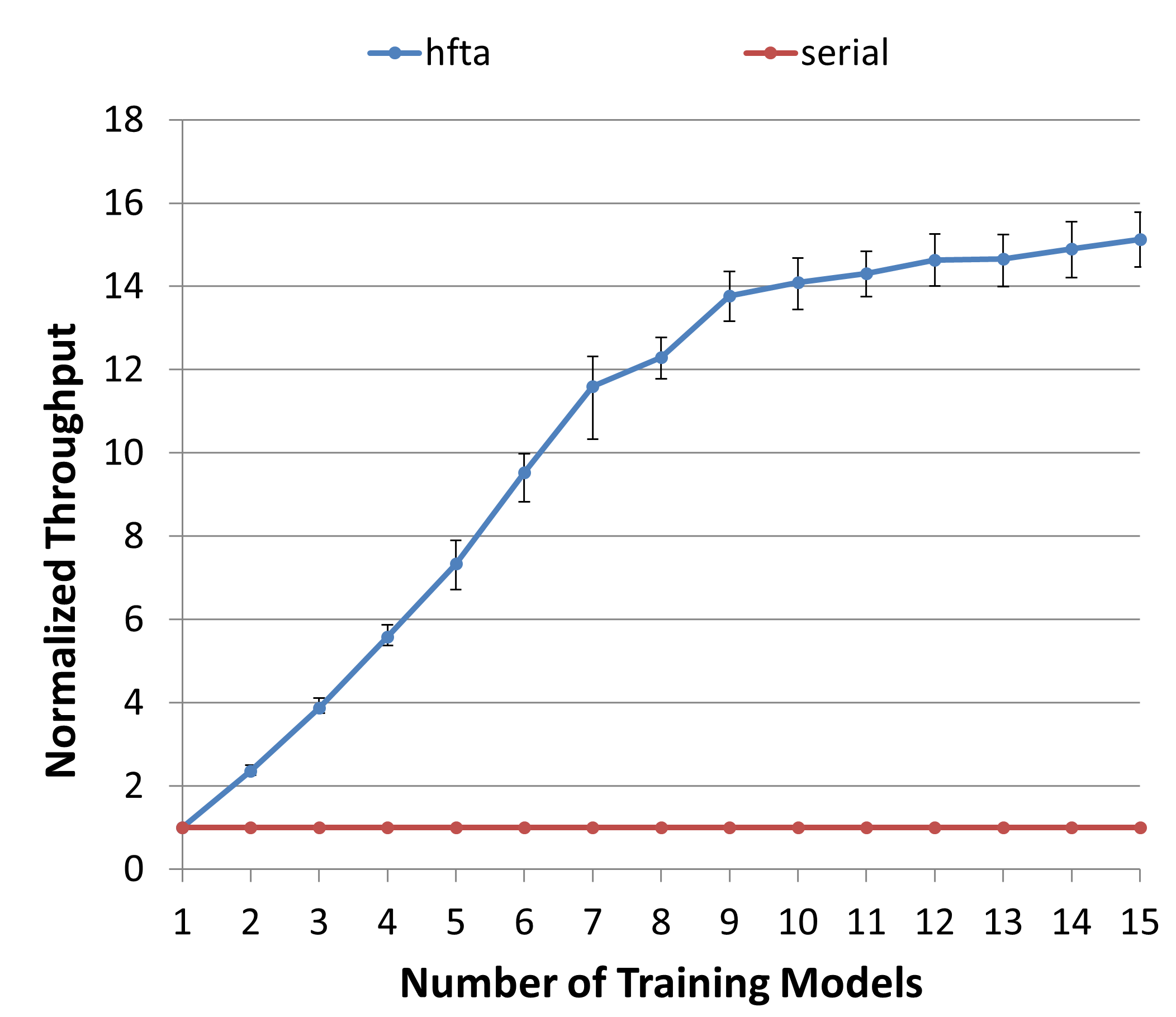}
            \vspace{-0.6cm}
            \caption{DCGAN}
            \label{fig:speedups_dcgan_tpu_v3}
        \end{subfigure}
        \vspace{-0.5cm}
        \caption{The normalized training throughput as we increase the number of models sharing (via HFTA) the same TPU v3 core.}
        \label{fig:speedups_tpus}
    \end{minipage}
    \hfill
    \begin{minipage}{0.44\linewidth}
        \centering
        \includegraphics[width=\linewidth]{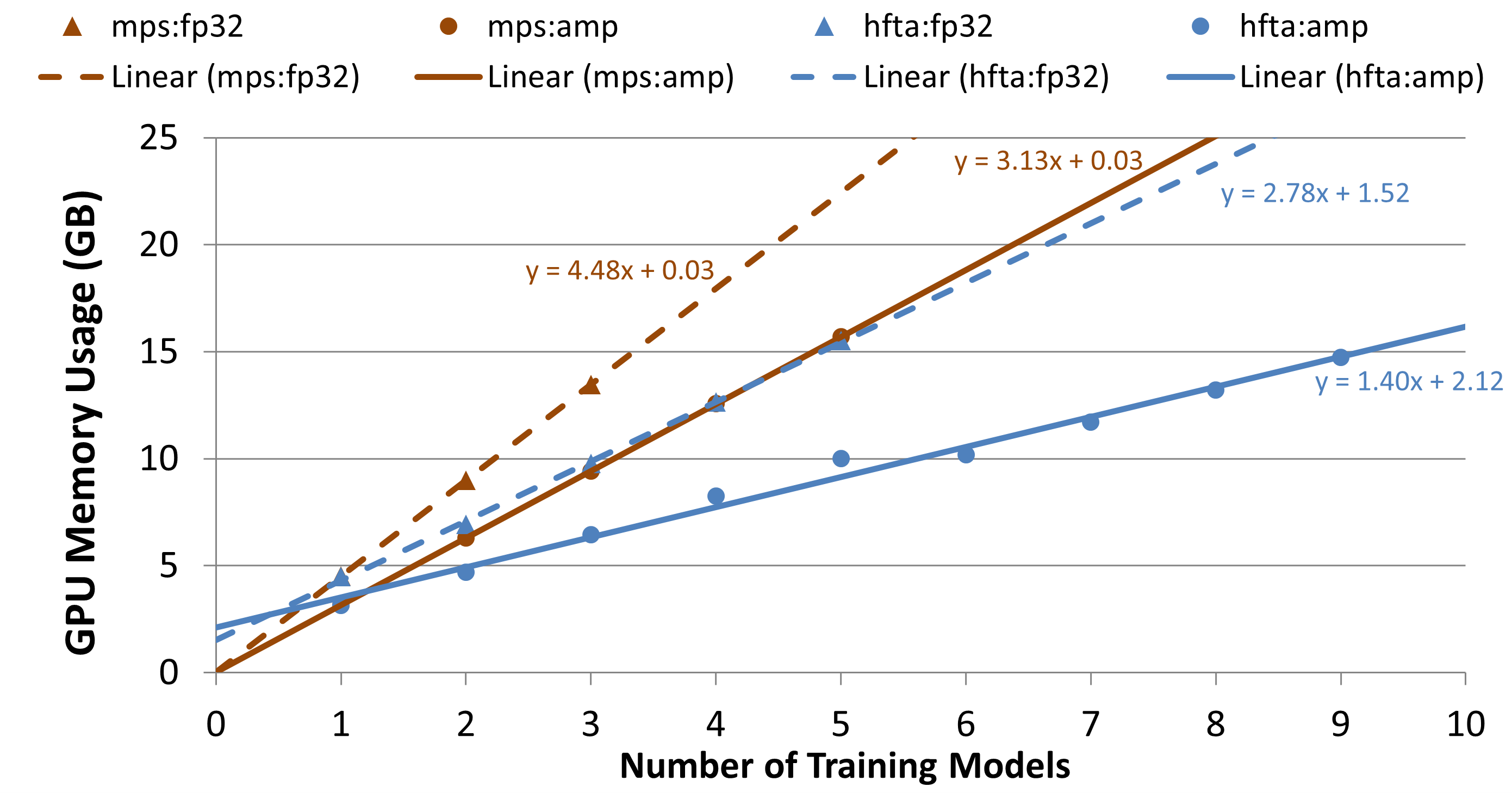}
        \vspace{-0.7cm}
        \caption{GPU Memory Footprints of \emph{MPS} and HFTA for PointNet classification task as we increase the number of models sharing the same V100.}
        \label{fig:gpu_mem_used_fit_pointnet_cls_v100}
    \end{minipage}
\end{figure*}

\vspace{-0.3cm}
\subsection{End-to-end Training Performance on TPUs} \label{sec:eval_tpus}
\vspace{-0.2cm}

\begin{figure*}[]
    \centering
    \vspace{-0.3cm}
    \begin{subfigure}[]{\linewidth}
        \centering
        \includegraphics[width=0.9\linewidth]{imgs/speedups_legend.png}
    \end{subfigure}
    \hfill
    \begin{subfigure}[]{0.32\linewidth}
        \centering
        \includegraphics[width=\linewidth]{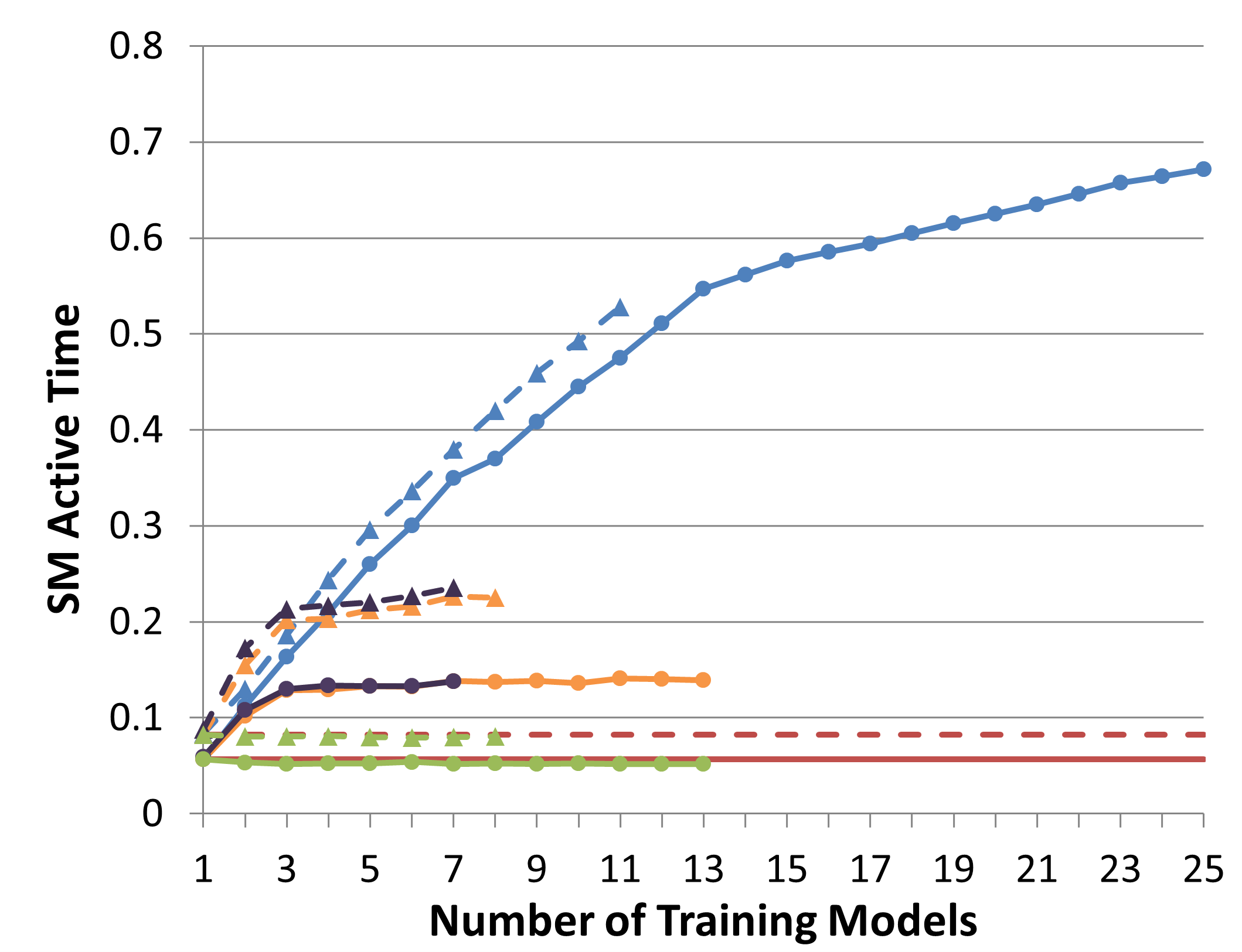}
        \caption{\texttt{sm\_active}}
        \label{fig:sm_active_pointnet_cls_a100}
    \end{subfigure}
    \hfill
    \begin{subfigure}[]{0.32\linewidth}
        \centering
        \includegraphics[width=\linewidth]{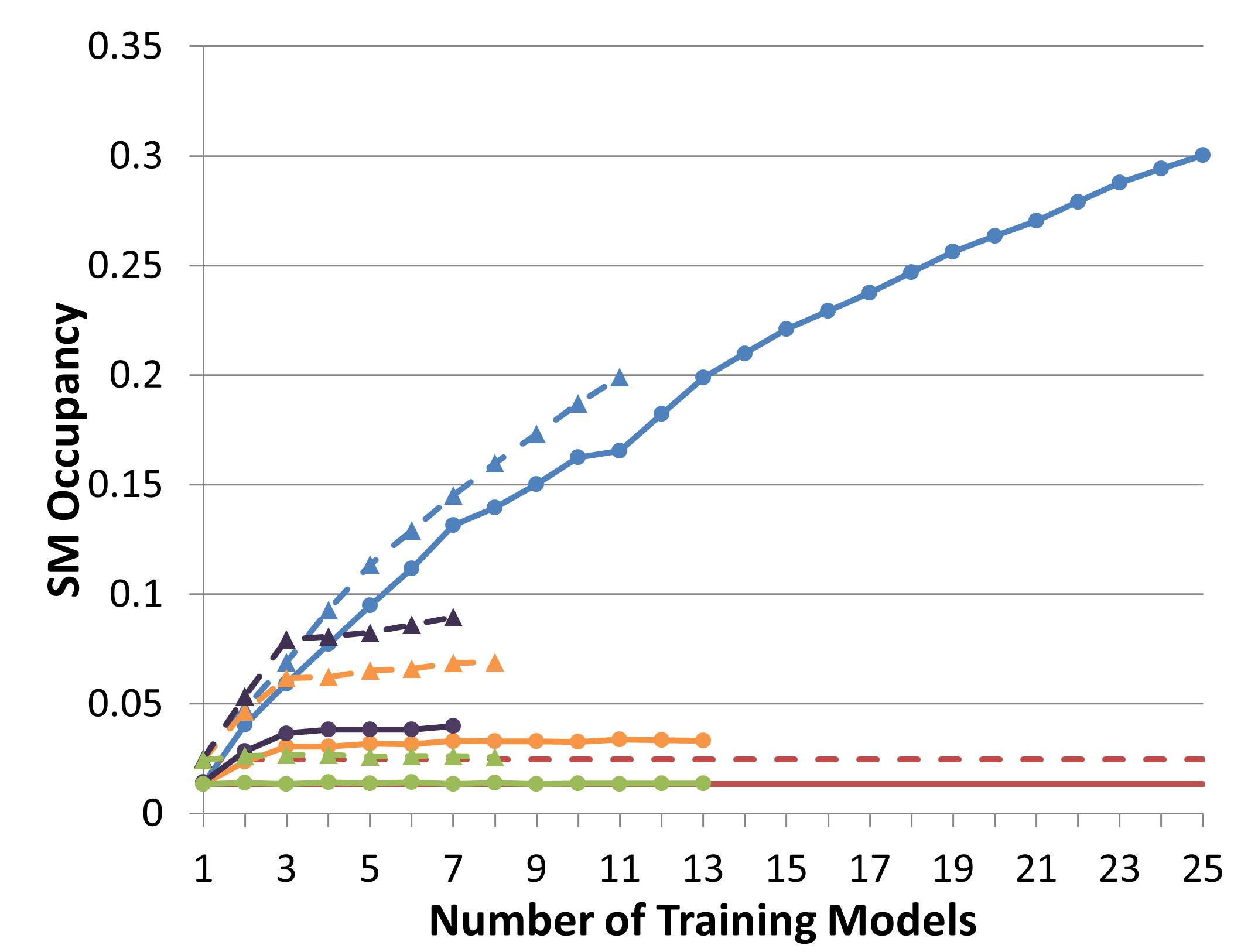}
        \caption{\texttt{sm\_occupancy}}
        \label{fig:sm_occupancy_pointnet_cls_a100}
    \end{subfigure}
    \hfill
    \begin{subfigure}[]{0.32\linewidth}
        \centering
        \includegraphics[width=\linewidth]{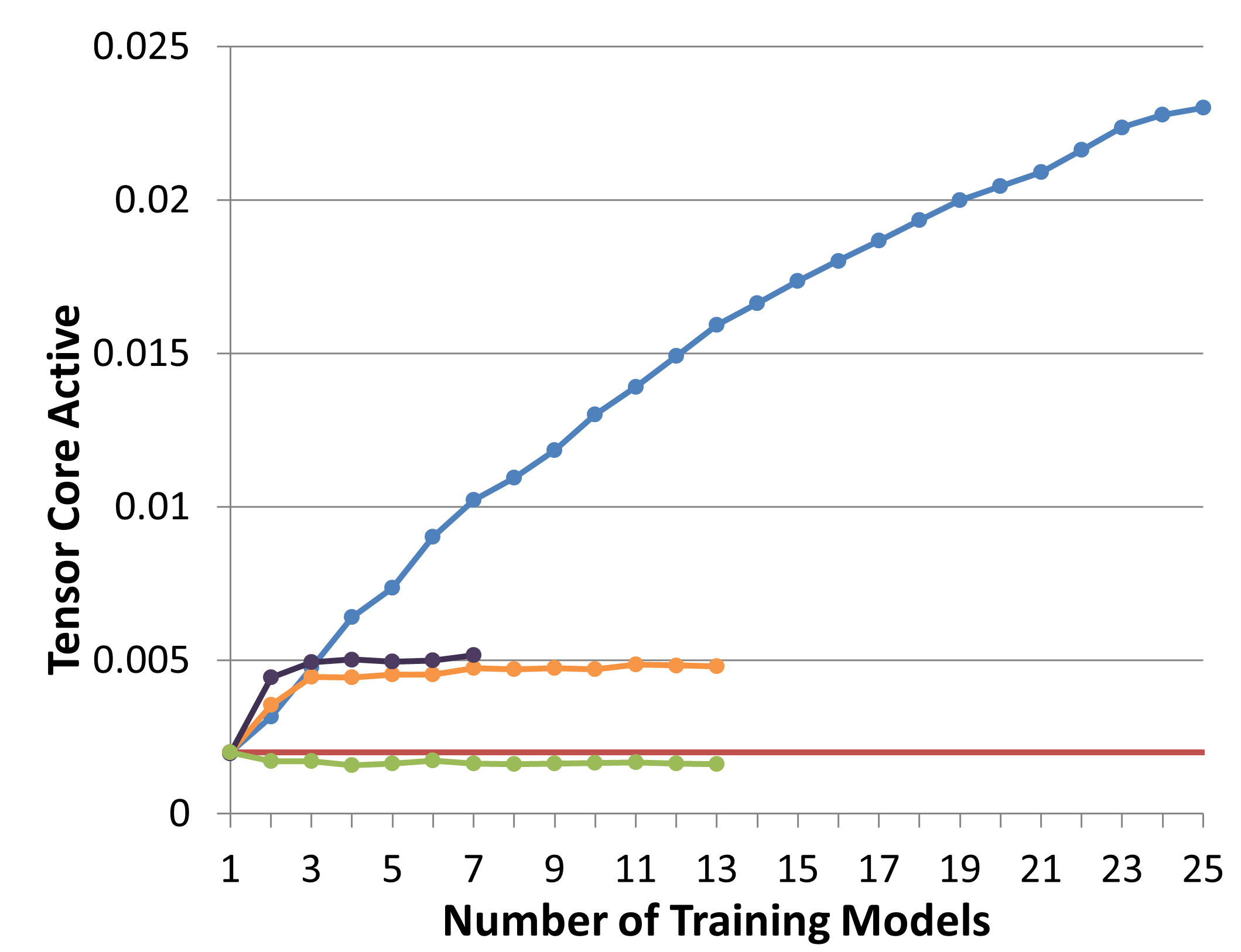}
        \caption{\texttt{tensor\_active}}
        \label{fig:tensor_active_pointnet_cls_a100}
    \end{subfigure}
%    \hfill
%    \begin{subfigure}[]{0.24\linewidth}
%        \centering
%        \includegraphics[width=\linewidth]{imgs/gpu_util_pointnet_cls_a100.png}
%        \caption{GPU utilization}
%        \label{fig:gpu_util_pointnet_cls_a100}
%    \end{subfigure}
    \vspace{-0.3cm}
    \caption{The hardware performance counters for PointNet classification task as we increase the number of models sharing the same A100.}
    \label{fig:perf_counters_pointnet_cls_a100}
    \vspace{-0.3cm}
\end{figure*}

As we aim to build a general solution that works for different ML accelerators,
we also evaluate HFTA on an entirely different type of accelerator: Google TPU v3. 
Figure~\ref{fig:speedups_tpus} plots the per-core training throughput for 
%the 
\emph{serial} 
%baseline 
vs. HFTA on the PointNet classification and DCGAN experiments on TPU v3, normalized by the throughput of the respective \emph{serial} baseline. 
Similar to previous results on GPUs, each HFTA curve shows how the normalized throughput increases with the number of models sharing the same TPU (until the fused models can not fit into the TPU HBM memory). 
We make three major observations from these figures.

First, \emph{HFTA achieves $\tpuPeakHFTAvsSerialCls \times$ / $\tpuPeakHFTAvsSerialDCGAN \times$ higher peak throughput than serial on the PointNet classification / DCGAN}. 

Second, we observe that \emph{for DCGAN, HFTA can sometimes achieve ``super-linear'' speedups}. Our current investigation concludes that the most likely cause of such a behaviour is the tensor padding added in the \emph{serial} baseline by the XLA~\cite{pytorch_xla} compiler \cite{tpu_troubleshooting}, making this baseline weaker than it should be otherwise. 

Additionally, we also investigate the HFTA's potential on the PointNet segmentation task. Unfortunately, HFTA currently achieves a less impressive $\tpuPeakHFTAvsSerialSeg \times$ speedup over the \emph{serial} baseline, which we attribute to the PointNet segmentation variant having many non-GEMM-based operators that intrinsically do \emph{not} map well to systolic arrays by the XLA compiler. Deeper analysis, however, is limited due to the xprof \cite{xprof} tool, just recently released, do not directly support PyTorch/XLA. We will perform deeper analysis of this problem and research potential solutions as soon as a proper version of the profiler is released.

\vspace{-0.3cm}
\subsection{In-depth Performance Analysis} \label{sec:gpu_perf_analysis}
\vspace{-0.2cm}

Using PointNet classification task as a case study, we perform deeper analysis 
%through 
by profiling GPU hardware performance counters to explain why HFTA %is able to 
can share the same GPU with more training workloads and achieves higher training throughput than the baselines.

Figure~\ref{fig:gpu_mem_used_fit_pointnet_cls_v100} plots the GPU memory footprint of \emph{MPS} and HFTA as we increase the number of models sharing the same V100 GPU~\footnote{The trends on RTX6000 and A100 are consistent with V100.} and the linear regression lines fitted on those measurements. 
Training models in independent processes duplicates the associated GPU memory overheads (reserved by the DL framework stack \cite{pytorch_reserved_gpu_mem}), which is a challenge that HFTA addresses. Thus, we can observe 
%from Figure~\ref{fig:gpu_mem_used_fit_pointnet_cls_v100}
that: (1) \emph{MPS}'s linear regression lines pass through the (0, 0) coordinate and have higher slopes than HFTA's; and (2) the intercepts of HFTA's linear regression lines essentially represent the exact amounts of memory overhead which are \hftaFpMemOverheadCls GB for FP32 training and \hftaAmpMemOverheadCls GB for AMP.

Figure~\ref{fig:perf_counters_pointnet_cls_a100} plots the \texttt{sm\_active}, \texttt{sm\_occupancy}, and \texttt{tensor\_active} %, and \texttt{nvidia-smi}-defined ``GPU utilization'' 
of HFTA and the baselines as we increase the number of models sharing the same A100 GPU.\footnote{V100 results are similar and shown in Appendix~\ref{append:additional_eval_data}.} We can observe: (1) HFTA's SM and TC utilization keeps scaling up as we fuse more models horizontally. (2) \emph{MIG}'s and \emph{MPS}'s SM and TC utilization plateaus at a smaller number of models and lower utilization, which supports our qualitative reasoning in Section~\ref{sec:hardware_sharing_hardware_feature} that both leave significant potential of training performance unharnessed; (3) \emph{concurrent}'s SM and TC utilization stays the same as \emph{serial} because the kernels from parallel processes can not execute concurrently without MPS or MIG.
%scheduled and executed serially without MPS or MIG.
% and (4) the \texttt{nvidia-smi}-defined ``GPU utilization'' is a weak utilization indicator, since the curves in Figure~\ref{fig:gpu_util_pointnet_cls_a100} appear rather noisy and do not follow the trends in any of the Figures~\ref{fig:speedups_pointnet_cls_a100}, ~\ref{fig:sm_active_pointnet_cls_a100}, ~\ref{fig:sm_occupancy_pointnet_cls_a100} or ~\ref{fig:tensor_active_pointnet_cls_a100}.

%\vspace{-0.3cm}
\subsection{Overall Cost Saving for Tuning Algorithms} \label{sec:hfht_cost_saving}
\vspace{-0.2cm}

\begin{figure}[]
    \centering
    \vspace{-0.0cm}
    \begin{subfigure}[]{\linewidth}
        \centering
        \includegraphics[width=0.35\linewidth]{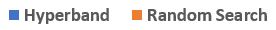}
        \vspace{-0.00cm}
    \end{subfigure}
    \hfill
    \begin{subfigure}[]{0.49\linewidth}
        \centering
        \includegraphics[width=\linewidth]{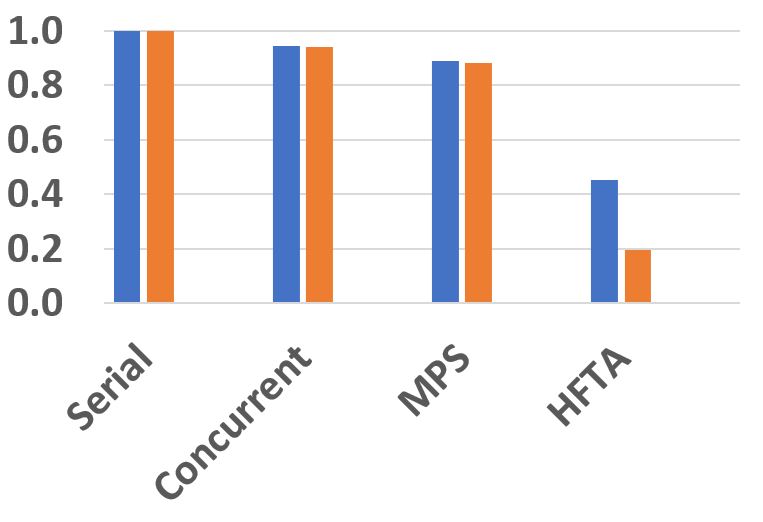}
        \vspace{-0.5cm}
        \caption{PointNet}
        \label{fig:hfht_pn_v100}
    \end{subfigure}
    %\hfill
    \begin{subfigure}[]{0.42\linewidth}
        \centering
        \includegraphics[width=\linewidth]{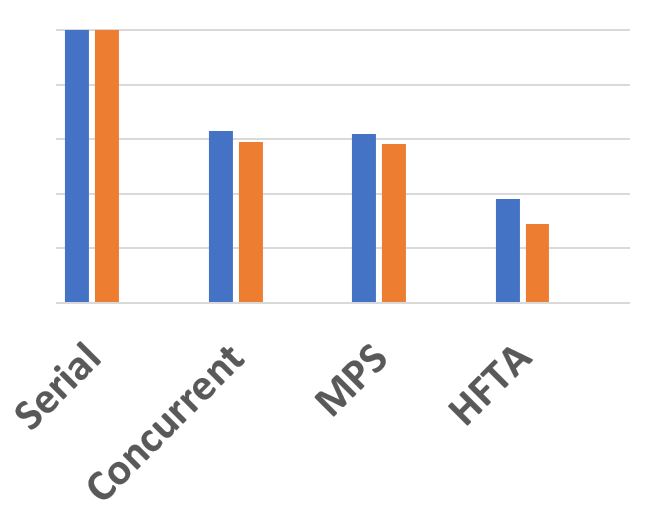}
        \vspace{-0.5cm}
        \caption{MobileNet}
        \label{fig:hfht_mn_v100}
    \end{subfigure}
    \vspace{-0.3cm}
    \caption{Total GPU hours of four hyper-parameter tuning workloads on V100. For each algorithm and scheduler, we show the lower total GPU hours between the FP32 and AMP training.}
    \label{fig:hfht_v100}
    \vspace{-0.5cm}
\end{figure}

To demonstrate that HFTA can be efficient in improving the hardware utilization for existing hyper-parameter tuning algorithms,
Figure~\ref{fig:hfht_v100} lists the total GPU hour cost of tuning eight hyper-parameters via HFHT for the PointNet and MobileNet classification tasks on the V100 GPU using two tuning algorithms (random search~\cite{random_search_for_hyper_parameter_optimization} and Hyperband~\cite{hyperband}) and four job schedulers (\emph{serial}, \emph{concurrent}, \emph{MPS}, and \emph{HFTA}). We make two major observations from this figure.

First, \emph{HFTA can reduce the total GPU hour cost by up to $5.10 \times$} and lead to significantly better hardware utilization than all other baselines.

Second,
as we theoretically discuss in Appendix~\ref{append:hfht},
\emph{random search benefits more from HFTA than Hyperband}. This is because, during certain iterations, Hyperband proposes to run \emph{many} epochs on just a \emph{few} sets of hyper-parameters. Thus, within such iterations, Hyperband generates not enough parallel jobs to either share the GPU with \emph{MPS} or to be fused by \emph{HFTA}. Therefore, such iterations become the bottleneck of the total cost when tuning is conducted with a job scheduler based on hardware sharing.
%\footnote{we will investigate how to adjust existing algorithms to better leverage HFTA as part of our future work.}

%As part of our future work, we will investigate how we can adjust existing tuning algorithms to be more HFTA-``friendly" (Appendix~\ref{append:hfht}).

%improve the HFTA-``friendliness" of existing tuning algorithms (Appendix~\ref{append:hfht}). 

%\begin{table}[]
%\centering
%\vspace{-0.5cm}
%\caption{Normalized total GPU hours of tuning 8 hyper-parameters for the PointNet classification task on V100. For each combination between algorithms and schedulers, the lower total GPU hours between FP32 and AMP training are shown below.} \label{tab:hfht_pointnet_classification_v100}
%\scriptsize
%\begin{tabular}{c|cc}
%\toprule
%Algorithm    & Hyperband & Random Search \\ 
%\midrule
%\emph{Serial}     & \hfhtGpuvPnClsHbSerialFp      & \textbf{\hfhtGpuvPnClsRsSerialFp}          \\   
%\emph{Concurrent} & \hfhtGpuvPnClsHbConcurrentFp      & \hfhtGpuvPnClsRsConcurrentFp          \\    
%\emph{MPS}        & \hfhtGpuvPnClsHbMpsFp      & \hfhtGpuvPnClsRsMpsFp          \\ 
%\emph{HFTA}       & \hfhtGpuvPnClsHbHftaAmp         & \hfhtGpuvPnClsRsHftaAmp             \\ 
%\bottomrule
%\end{tabular}
%\end{table}

\vspace{-0.3cm}
%\subsection{Prior Works} \label{sec:prior_works}
\section{Related Works} \label{sec:prior_works}
\vspace{-0.2cm}

Major prior works on DL job fusion~\cite{pack,modelbatch,hivemind} suffer from three key weaknesses: (i) avoiding directly addressing hardware under-utilization, (ii) strongly depending on the CUDA stream primitive~\cite{cuda_stream},
%that is 
a generic GPU-sharing method but inefficient for repetitive training jobs, and (iii) employing very restricted fusion schemes that are ineffective in practice. We discuss these prior works in detail below.

\texttt{pack}~\cite{pack} merges TensorFlow~\cite{tensorflow} graphs from multiple training jobs into a single graph 
%in order to 
to amortize \emph{only} the IO and data preprocessing cost, but does not address the hardware under-utilization from the model forward and backward passes.

%In addition
Furthermore, ModelBatch~\cite{modelbatch} attempts to parallelize the kernel launches from multiple training jobs via CUDA streams (the CUDA programming interface of Hyper-Q~\cite{hyperq}), which suffers from similar pitfalls of runtime overhead duplication as MPS. 

Although \emph{intra}-model vertical and horizontal fusion of DL operators have been studied extensively by many prior works \cite{nvidia_rnn_fusion, tensorrt, tensor_comprehensions, glow, tvm, metaflow}, \emph{inter}-model horizontal fusion has only been explored in extremely limited depth: 
HiveMind~\cite{hivemind} proposes fusion schemes for 1) non-stateful operators with the same shapes, 2) stateful operators that share the same weights, and 3) stateful operators that share the same shapes and inputs. Unfortunately, condition 2) is rarely applicable to training workloads since each individual model has its own weights, while condition 3) usually only applies to the first operator in a DL model since the following operators will have different inputs, leaving most of the fusion opportunities completely untapped. 
%HiveMind~\cite{hivemind} \hl{proposes fusion schemes that either 1) are rarely applicable to training workloads, or 2) leaves most of fusion opportunities completely untapped.}
%In addition
Besides, HiveMind does not demonstrate any performance improvement over MPS as it also relies on CUDA streams to extract utilization when its fusion scheme becomes ineffective. Therefore, HiveMind approach is hard to generalize to accelerators with no hardware-specific sharing features (e.g., TPUs).

In contrast, %our proposal, 
HFTA %, 
can fuse any operators of the same types 
%that share 
and 
%the same 
shapes across training jobs, which generally leads to full inter-model fusions. 
Moreover, HFTA demonstrates significant performance improvement against the existing widely-adopted generic hardware-based sharing approaches (e.g., MPS and MIG) since operator fusion does not possess the same shortcomings of those approaches, as we show in Section~\ref{sec:hardware_sharing_hardware_feature}. 
Finally, HFTA requires no hardware or DL framework stack modifications and applies to any existing hardware backends, including GPUs, TPUs, and any other accelerators that the major DL frameworks support.

\emph{In parallel with our work:}
Rammer~\cite{rammer} proposes a data flow graph compiler that enables hardware sharing via operator and accelerator abstractions which requires special operator implementations (e.g., using the Persistent Thread programming model~\cite{persistent_thread} for CUDA), and focuses mostly on intra-model horizontal  fusion and inference workloads.
%However, this approach requires unconventional re-implementations of all operators (e.g., using the Persistent Thread programming model~\cite{persistent_thread} for CUDA). We speculate Rammer's lack of sufficient training support is mainly due to (1) the lack of SOTA open-sourced backward pass operator implementations and (2) the significant engineering efforts to implement those operators from scratch. Moreover, Rammer does not demonstrate speedups against MPS. 
Retiarii~\cite{retiarii} hints that observations similar to ours are possible, but does not focus on exploring this idea in depth.

\vspace{-0.3cm}
\section{Conclusion} \label{sec:conclusion}
\vspace{-0.2cm}

%In this work, we collect ``real-world'' GPU cluster usage statistics and learn that, 
%Despite many efforts on optimizing DL training workloads, 
In this work, we learn from the 
%``real-world'' 
GPU cluster usage analysis that repetitive single-accelerator training jobs (e.g., for hyper-parameters tuning) often 
dominate cluster-wide hardware resource usage and can severely under-utilize the hardware. 
%These training jobs also tend to
%have low hardware utilization, since DL researchers and practitioners often lack the relevant expertise to 
%independently 
%optimize their 
%own 
%workloads independently. 
To address this challenge, 
we observe that these jobs possess unique characteristics which enables the inter-model horizontal fusion. 
%optimization.
%we make the following observations on the unique characteristics of these jobs: (1) the models among such jobs often have the same types of operators with the same shapes;\ and (2) the inter-model horizontal fusion of such operators is mathematically equivalent to other already well-optimized operators. Built upon these observations,
%\hl{by leveraging the the unique characteristics of these jobs,}
Therefore, we propose the HFTA (DL framework extension) library that 
%enables DL researchers and practitioners to 
horizontally fuses the models deeply down to operators with a minimal extra effort from DL researchers and practitioners, significantly improving the hardware utilization of these workloads by simultaneously training many models on the same accelerator. 
On %the PointNet classification and segmentation tasks, and DCGAN, 
six highly impactful DL models,
HFTA achieves up to $\tpuPeakHFTAvsSerialUpperRange \times$ higher training throughput than running each job on a separate accelerator, a common practice employed by hyper-parameter tuning frameworks.
%, and on GPUs, $\gpuPeakHFTAvsMPSUpperRange \times$ than hardware-based sharing via MPS and $\gpuPeakHFTAvsMIGUpperRange \times$ than MIG. 
We continue to expand the coverage of HFTA with more operators, optimizers, and learning rate schedulers,
%integrating HFTA into existing hyper-parameter tuning and model architecture search frameworks. 
and to investigate how existing DL frameworks, hyper-parameter tuning and model architecture search algorithms can be adjusted to extract more performance from hardware sharing via HFTA.
We hope our work inspires future research on assisting ML researchers and developers with limited optimization experience to better utilize the hardware for their novel DL models.

\vspace{-0.3cm}
\section*{Acknowledgements}
\vspace{-0.2cm}

We want to especially thank Suvinay Subramanian for TPU-related issues and discussions. We want to thank Xiaodan (Serina) Tan, Suvinay Subramanian, James Gleeson, Anand Jayarajan,
and Garth Gibson for their constructive feedback during the development of this work. We want to thank Google for TPU credits and early accesses to the GCP A2 Alpha version instances.
This project was supported in part by the Canada Foundation for Innovation JELF grant, NSERC Discovery grant, AWS Machine Learning Research Award, and Facebook Faculty Research Award.

\bibliography{main}
\bibliographystyle{mlsys2021}

%%%%%%%%%%%%%%%%%%%%%%%%%%%%%%%%%%%%%%%%%%%%%%%%%%%%%%%%%%%%%%%%%%%%%%%%%%%%%%%
%%%%%%%%%%%%%%%%%%%%%%%%%%%%%%%%%%%%%%%%%%%%%%%%%%%%%%%%%%%%%%%%%%%%%%%%%%%%%%%
% SUPPLEMENTAL CONTENT AS APPENDIX AFTER REFERENCES
%%%%%%%%%%%%%%%%%%%%%%%%%%%%%%%%%%%%%%%%%%%%%%%%%%%%%%%%%%%%%%%%%%%%%%%%%%%%%%%
%%%%%%%%%%%%%%%%%%%%%%%%%%%%%%%%%%%%%%%%%%%%%%%%%%%%%%%%%%%%%%%%%%%%%%%%%%%%%%%
\clearpage

\section*{Summary of Appendices}

%Due to space constraints, the appendices provide additional important details that accompany the main text of this paper. We include the following content:
These appendices cover the following content that does not fit into the paper's main text due to space constraints.
%\begin{itemize}
    %\item 
    
Appendix~\ref{append:hardware_under_utilization} describes the methodology that we use to collect 
%``real-world'' 
the GPU cluster usage statistics from the Vector Institute. It also provides 
%the 
empirical evidence to support our observation that the dominating single-GPU training jobs often have low hardware utilization.
    %\item 
    
Appendix~\ref{append:hfta_op_fusion_rule} lists the operators that HFTA currently supports as well as their corresponding horizontally-fused counterparts.
    %\item 
    
Appendix~\ref{append:loss_scale} describes how fused loss functions are handled in order to reconstruct mathematically equivalent gradients. 
    
Appendix~\ref{append:converge} provides empirical evidence that HFTA has no impact on the model's original convergence.

Appendix~\ref{append:hfht} discusses our approach to integrate HFTA with existing hyper-parameter tuning algorithms via HFHT in detail.
    
Appendix~\ref{append:dcgm} shows how we collect the GPU hardware performance counters and provides the related references.
    %\item 
    
Appendix~\ref{append:additional_eval_data} provides additional statistics and insights that can help 
%to 
clarify our observations and conclusions in Section~\ref{sec:eval}.
%, which does not fit into the paper's main text due to space constraints.
%\end{itemize}

Appendix~\ref{append:secondary_eval_method} includes additional evaluation methodology regarding our secondary benchmarks (i.e., ResNet-18, MobileNetV3-Large, Transformer and BERT-Medium), convergence validation, HFHT and partial fusion.

Appendix~\ref{append:secondary_eval} shows the additional evaluation results on our secondary benchmarks and partially fused ResNet-18.

\appendix

\section{``Real-world'' GPU Cluster Usage Statistics} \label{append:hardware_under_utilization}

We analyzed the job submissions and execution logs for a two-month period (July \nth{1} to Sept. \nth{1}, 2020) from a large GPU cluster belonging to the Vector Institute, an independent, not-for-profit corporation dedicated to research in the field of artificial intelligence and machine learning~\cite{vector}. 
The cluster services a variety of 
%deep learning 
DL training workloads from the Vector Institute's community. 
The community consists of 501 faculty, postdoc and student researchers who have published 263 conference and journal papers from April 2019 to March 2020, including 61 papers in NeurIPS, ICLR, CVPR and ICML. 

The cluster includes 4 GPU partitions, V1a (200 P100 GPUs), V1b (40 T4 GPUs), V2 (480 T4 GPUs) and V3 (240 RTX6000 GPUs), where V3 came online in the last few days of the collection period. 
V2 was recorded for the entire period, and the other three partitions were recorded for the last 11 days.  V2 is distinguished as the largest partition with the least powerful GPUs. The data contains information on 51338 jobs. The total number of GPU hours spent in these two months amounts to 471768 (equivalent to $\sim$317 GPU days per day). 

We classify the submitted jobs as ``repetitive single-GPU training jobs'' if they contain the following submission and execution patterns:

\begin{enumerate}
\item Each job only requests a single GPU despite the availability of multiple GPUs on the same node (i.e., not single-node distributed training). The job also does not require specifically which node the GPU resides (i.e., not multi-node distributed training). Therefore, it can only be a single-GPU training job.
\item Within a short time period (60 seconds), a batch of such single-GPU jobs are submitted from the same user, which means that the submission of these jobs is automated, and possibly contains the same code/program with varying parameters.\footnote{The exact code for each job was not available to us due to security/IP concerns.}
\item The job names are very similar within the batch for such a short time period. We determine the similarity by calculating the normalized Levenshtein distance~\cite{levenshtein_distance} among job names with a threshold of 0.9. As a reference, the distance score between two job names ranges from 0 to 1, where 1 represents being completely identical and 0 represents being different. This filter further verifies that these jobs are repetitive single-GPU jobs since the job names are very similar. Afterwards, a manual inspection of the job names within the batches indicates that those names usually contain small variations such as learning rate value or optimizer choices and settings.
\end{enumerate}

\begin{figure}[t]
    \centering
    \includegraphics[width=0.9\linewidth]{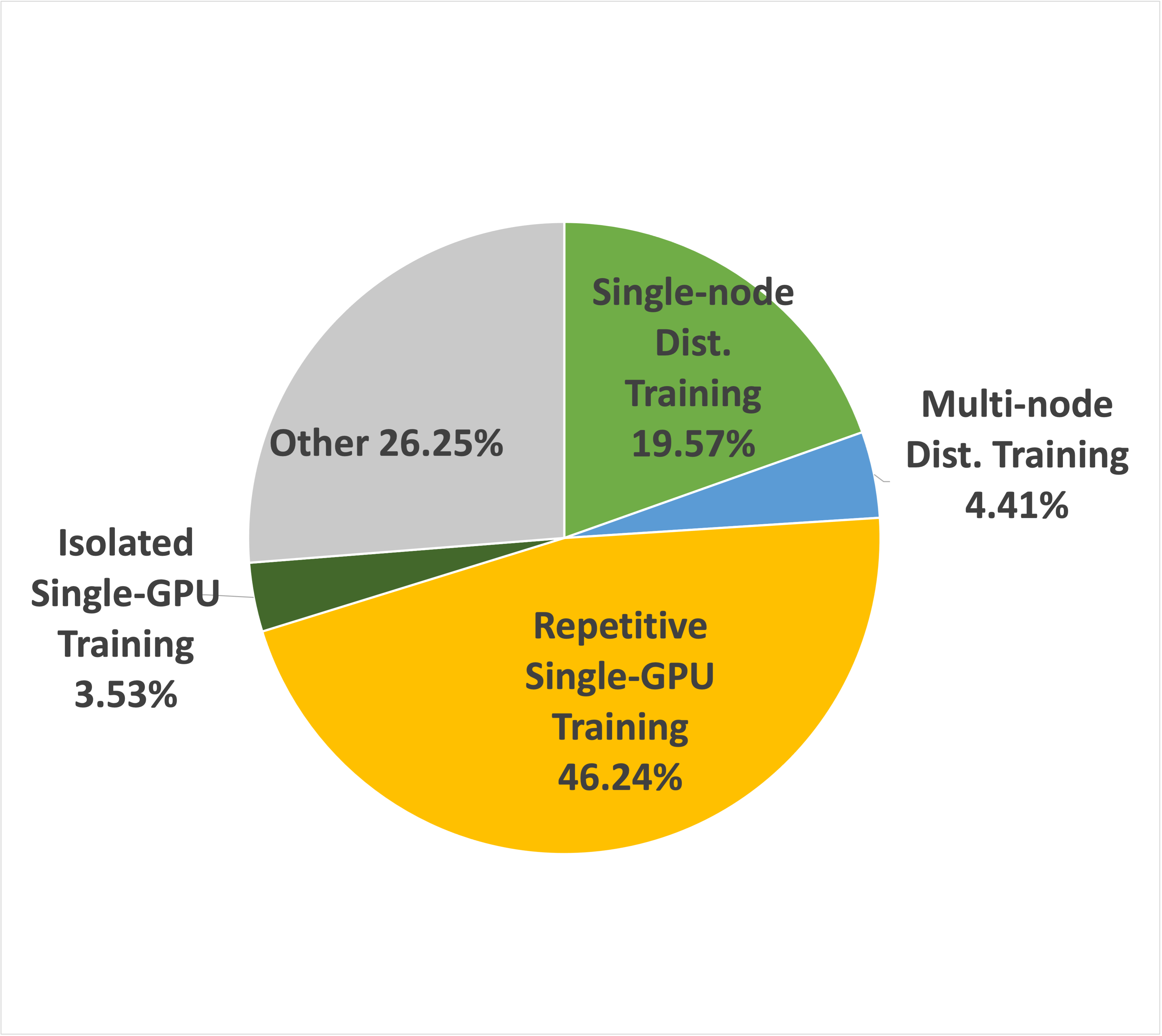}
    \caption{GPU hour usage breakdown for two consecutive months of a large GPU cluster from the Vector Institute.}
    \label{fig:vector_gpu_hour}
\end{figure}

We further reached out to individual users to confirm our conclusion. We interviewed 11 active (i.e., most frequent) users of the GPU cluster: (1) 7 users responded that more than 50\% of their jobs are repetitive single-GPU training for purposes including hyper-parameter tuning; and (2) 4 of those 7 users submitted over 95\% of their jobs for repetitive single-GPU training. The GPU hour usage distribution is plotted in Figure~\ref{fig:vector_gpu_hour}.

\begin{figure}[t]
\centering
\begin{subfigure}[]{\linewidth}
    \centering
    \includegraphics[width=0.9\linewidth]{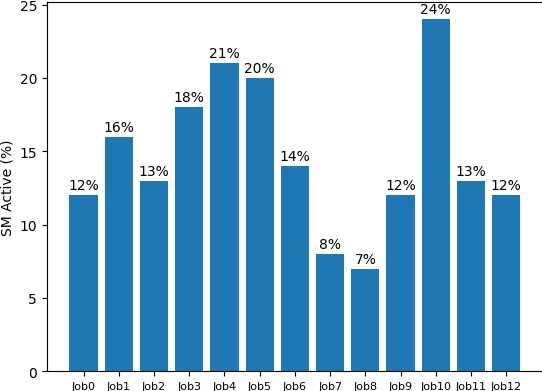}
    \caption{\texttt{sm\_active}} \label{fig:vector_gpu_sm_active}
\end{subfigure}
\hfill
\begin{subfigure}[]{\linewidth}
    \centering
    \includegraphics[width=0.9\linewidth]{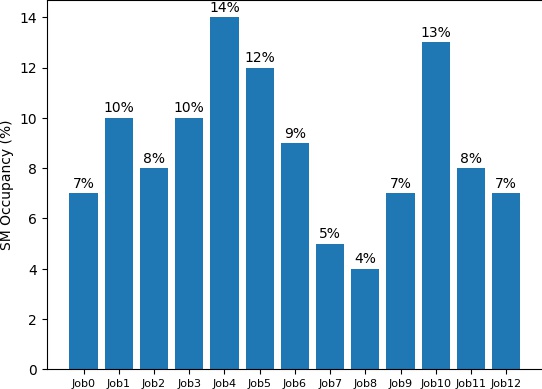}
    \caption{\texttt{sm\_occupancy}} \label{fig:vector_gpu_sm_occupancy}
\end{subfigure}
\caption{GPU hardware performance counters measured via DCGM~\cite{dcgm} for 13 jobs sampled from the clump of repetitive single-GPU training jobs.} \label{}
\end{figure}

Since the cluster does not actively monitor GPU hardware performance counters, we randomly sampled several jobs tagged as repetitive single-GPU training jobs and manually gathered the performance counters. 
Based on the \texttt{sm\_active} and \texttt{sm\_occupancy} (explained in Section~\ref{sec:eval_method} and elaborated in Appendix~\ref{append:dcgm}) metrics from our samples, we observe that many of the repetitive single-GPU training jobs can severely under-utilize the GPUs both temporally and spatially (as shown in Figure~\ref{fig:vector_gpu_sm_active} and Figure~\ref{fig:vector_gpu_sm_occupancy}, respectively). 
The maximum \texttt{sm\_active} among the sampled jobs is 24\%, and maximum \texttt{sm\_occupancy} 
%among them 
is 14\%.

\section{HFTA Operator Fusion Rules} \label{append:hfta_op_fusion_rule}

\begin{table*}[t]
\centering
\caption{The horizontal fusion rules for the operators that HFTA currently supports. ``ConvT'' stands for ``ConvTranspose'' (a.k.a., deconvolution). $\vec{x}$, $\vec{y}$, $\vec{w}$ and $\vec{b}$ represent the input, output, weight and bias tensors, respectively. $N$, $C$, $H$, $W$ and $L$ represent the batch sizes, channel sizes, heights, widths, and signal lengths of the tensors used in convolutions, deconvolution, batch-norms, MaxPool2d, and Dropout2d. $G$ represents the numbers of groups used in the convolutions and deconvolution. $F$ represents the feature sizes of the tensors used in linear layers. $D$ represents arbitrary unmodified dimensions in LayerNorm and embedding layers. $E$ represents the dimensions for normalization in LayerNorm. $\varepsilon$ and $\xi$ represents the number of embedding vectors and the size of each embedding vector for the embedding layer. $\ast$ represents zero or more arguments whose values are kept the same. $B$ represents the number of operators horizontally fused together via HFTA.} \label{tab:hfta_operators}
\scriptsize
\begin{tabular}{l|l}
\toprule
PyTorch Operator (Tensors: Shapes, Other Parameters = Arguments) & HFTA Horizontally Fused Operator (Tensors: Shapes, Other Parameters = Arguments) \\\midrule
Conv2d($\vec{x}: [N, C_{\vec{x}}, H_{\vec{x}}, W_{\vec{x}}]$, $\vec{w}: [C_{\vec{y}}, \frac{C_{\vec{x}}}{\text{G}}, H_{\vec{w}}, W_{\vec{w}}]$, $\vec{b}: [C_{\vec{y}}]$, $\text{G}=g$, $\ast$) &
Conv2d($\vec{x}: [N, B \times C_{\vec{x}}, H_{\vec{x}}, W_{\vec{x}}]$, $\vec{w}: [C_{\vec{y}}, \frac{B \times C_{\vec{x}}}{\text{G}}, H_{\vec{w}}, W_{\vec{w}}]$, $\vec{b}: [B \times C_{\vec{y}}]$, $\text{G}=B \times g$, $\ast$) \\

Conv1d($\vec{x}: [N, C_{\vec{x}}, L_{\vec{x}}]$,  $\vec{w}: [C_{\vec{y}}, \frac{C_{\vec{x}}}{\text{G}}, L_{\vec{w}}]$, $\vec{b}: [C_{\vec{y}}]$, $\text{G}=g$, $\ast$)  &  
Conv1d($\vec{x}: [N, B \times C_{\vec{x}}, L_{\vec{x}}]$, $\vec{w}: [C_{\vec{y}},\frac{B \times C_{\vec{x}}}{\text{G}}, L_{\vec{w}}]$, $\vec{b}: [B \times C_{\vec{y}}]$, $\text{G}=B \times g$, $\ast$) \\

% This line is too long.
ConvT2d($\vec{x}: [N, C_{\vec{x}}, H_{\vec{x}}, W_{\vec{x}}]$, $\vec{w}: [C_{\vec{y}}, \frac{C_{\vec{x}}}{\text{G}}, H_{\vec{w}}, W_{\vec{w}}]$, $\vec{b}: [C_{\vec{y}}]$, $\text{G}=g$, $\ast$) &  
ConvT2d($\vec{x}: [N, B \times C_{\vec{x}}, H_{\vec{x}}, W_{\vec{x}}]$, $\vec{w}: [C_{\vec{y}}, \frac{B \times C_{\vec{x}}}{\text{G}}, H_{\vec{w}}, W_{\vec{w}}]$, $\vec{b}: [B \times C_{\vec{y}}]$, $\text{G}=B \times g$, $\ast$)  \\

Linear($\vec{x}: [N,  F_{\vec{x}}]$,  $\vec{w}: [F_{\vec{x}}, F_{\vec{y}}]$, $\vec{b}: [F_{\vec{y}}]$)  &  
baddbmm($\vec{b}: [B, 1, F_{\vec{y}}]$, $\vec{x}: [B, N, F_{\vec{x}}]$, $\vec{w}: [B, F_{\vec{x}}, F_{\vec{y}}]$) \\

BatchNorm1d($\vec{x}: [N, C_{\vec{x}}]$ or $[N, C_{\vec{x}}, L_{\vec{x}}]$,  $\vec{w}: [C_{\vec{x}}]$, $\vec{b}: [C_{\vec{x}}]$, $\ast$)  &  
BatchNorm1d($\vec{x}: [N, B, C_{\vec{x}}]$ or $[N, B \times C_{\vec{x}}, L_{\vec{x}}]$, $\vec{w}: [B \times C_{\vec{x}}]$, $\vec{b}: [B \times C_{\vec{x}}]$, $\ast$) \\

BatchNorm2d($\vec{x}: [N, C_{\vec{x}}, H_{\vec{x}}, W_{\vec{x}}]$, $\vec{w}: [C_{\vec{x}}]$, $\vec{b}: [C_{\vec{x}}]$, $\ast$) &
BatchNorm2d($\vec{x}: [N, B \times C_{\vec{x}}, H_{\vec{x}}, W_{\vec{x}}]$, $\vec{w}: [B \times C_{\vec{x}}]$, $\vec{b}: [B \times C_{\vec{x}}]$, $\ast$)  \\

LayerNorm($\vec{x}: [N, D_1, ..., D_m, E_{1}, ..., E_{n}]$, $\vec{w}: [E_{1}, ..., E_{n}]$, $\vec{b}: [E_{1}, ..., E_{n}]$, $\ast$) &
(LayerNorm($\vec{x}: [B, N, D_1, ..., D_m, E_{1}, ..., E_{n}]$, $\ast$) $\times \vec{w} + \vec{b}$); $\vec{w}, \vec{b}: [B, 1_1, ..., 1_m, E_{1}, ..., E_{n}]$  \\

Embedding($\vec{x}: [D_1, ..., D_m]$, $\vec{w}: [\varepsilon, \xi]$) & 
Embedding($\vec{x}: [B, D_1, ..., D_m] + (\varepsilon, 2 \varepsilon, ..., B \varepsilon): [B, 1_1, ..., 1_m]$, $\vec{w}: [B \times \varepsilon, \xi]$) \\

MaxPool2d($\vec{x}: [N, C_{\vec{x}}, H_{\vec{x}}, W_{\vec{x}}]$, $\ast$)  &  
MaxPool2d($\vec{x}: [N, B \times C_{\vec{x}}, H_{\vec{x}}, W_{\vec{x}}]$, $\ast$) \\

AdaptiveAvgPool2d($\vec{x}: [N, C_{\vec{x}}, H_{\vec{x}}, W_{\vec{x}}]$, $\ast$)  &  
AdaptiveAvgPool2d($\vec{x}: [N, B \times C_{\vec{x}}, H_{\vec{x}}, W_{\vec{x}}]$, $\ast$) \\

Dropout2d($\vec{x}: [N, C_{\vec{x}}, H_{\vec{x}}, W_{\vec{x}}]$, $\ast$)  &  
Dropout2d($\vec{x}: [N, B \times C_{\vec{x}}, H_{\vec{x}}, W_{\vec{x}}]$, $\ast$) \\

Dropout($\vec{x}: [\ast]$)  &  
Dropout($\vec{x}: [\ast, B, \ast]$, $\ast$) \\

LeakyReLU($x: [\ast]$, $\ast$)  &  
LeakyReLU($x: [\ast, B, \ast]$, $\ast$) \\

ReLU($x: [\ast]$, $\ast$)  &  
ReLU($x: [\ast, B, \ast]$, $\ast$) \\

ReLU6($x: [\ast]$, $\ast$)  &  
ReLU6($x: [\ast, B, \ast]$, $\ast$) \\

LeakyReLU($x: [\ast]$, $\ast$)  &  
LeakyReLU($x: [\ast, B, \ast]$, $\ast$) \\

Tanh($x: [\ast]$)  &  
Tanh($x: [\ast, B, \ast]$) \\

%MultiheadAttention($\vec{q}, \vec{k}, \vec{v}:[L, N, E], \vec{w_q},\vec{w_k},\vec{w_v}:[L, E], \vec{w_y}:[L, L], \vec{b_q},\vec{b_k},\vec{b_v}:[E], \vec{b_y}:[L]$) &
%MultiheadAttention($\vec{q}, \vec{k}, \vec{v}:[B, N, L, E], \vec{w_q},\vec{w_k},\vec{w_v}:[B, L, E], \vec{w_y}:[B, L, L], \vec{b_q},\vec{b_k},\vec{b_v}:[B, E], \vec{b_y}:[B, L]$) \\
\bottomrule
\end{tabular}
\end{table*}

We list the horizontal operator fusion rules for 17 PyTorch operators in Table~\ref{tab:hfta_operators}.
The left column contains the original operators, and the right column indicates using which operator we could get the mathematically equivalent horizontally-fused version of $B$ original operators.
These operators are commonly used in DL research and development, and sufficient to implement a wide range of state-of-the-art DL models. 
%(e.g., SqueezeNet~\cite{squeezenet}, VGG~\cite{vgg}, ConvLSTM~\cite{convlstm}, DenseNet~\cite{densenet} and Inception~\cite{inception}) beyond what we showcase in our benchmarks.
Building on top of these fusion rules, we further develop the fused multihead attention layer and the fused Transformer encoder layer to support models that are based on the attention mechanisms or transformers.

\section{Fused Loss} \label{append:loss_scale}

In the following, we show how loss fusion is handled in order to reconstruct mathematically equivalent gradients. The inter-model horizontally fused loss with \emph{mean reduction} is shown as:\vspace{-0.0cm}
{
\begin{equation} \label{eqn:fused_loss}
    \mathcal{L} = \frac{1}{B} \sum_{b=0}^{B} \ell_{b}
\end{equation}
}\vspace{-0.0cm}
where $\ell_{b}$ is the loss of the $b$-th model, and there are $B$ models in total contributing to the fused loss $\mathcal{L}$. Taking the gradients on both side of Equation~\ref{eqn:fused_loss} with respect to the parameters $\vec{\theta}_{\beta}$ of a specific model $\beta$ results in:\vspace{-0.0cm}
{
\begin{equation} \label{eqn:grad_avg}
%\begin{align*}
    \nabla_{\vec{\theta}_{\beta}} \mathcal{L} 
    = \frac{1}{B}  \nabla_{\vec{\theta}_{\beta}} \sum_{b=0}^{B} \ell_{b} = \frac{1}{B} \sum_{b=0}^{B} \nabla_{\vec{\theta}_{\beta}} \ell_{b}
    %&= \frac{1}{B} (\sum_{b=0}^{\beta-1} \nabla_{\vec{\theta}_{\beta}} \ell_{b} + \nabla_{\vec{\theta}_{\beta}} \ell_{\beta} + \sum_{b=\beta+1}^{B} \nabla_{\vec{\theta}_{\beta}} \ell_{b})\\
    %&= \frac{1}{B} (0 + \nabla_{\vec{\theta}_{\beta}} \ell_{\beta} + 0) 
    = \frac{1}{B} \nabla_{\vec{\theta}_{\beta}} \ell_{\beta}
%\end{align*}
\end{equation}
}\vspace{-0.0cm}
because $\nabla_{\vec{\theta}_{\beta}} \ell_{b} = 0$ if $b \neq \beta$. We can rearrange Equation~\ref{eqn:grad_avg} into:\vspace{-0.0cm}
{
\begin{equation} \label{eqn:loss_scale}
    \nabla_{\vec{\theta}_{\beta}} \ell_{\beta} = B \nabla_{\vec{\theta}_{\beta}} \mathcal{L} = \nabla_{\vec{\theta}_{\beta}} B \mathcal{L}
\end{equation}
}\vspace{-0.0cm}

We can recognize that the expression on the left-hand side of Equation~\ref{eqn:loss_scale} is precisely the gradients for model $\beta$ if each model were trained independently. 
Therefore, in order to reconstruct exactly the same gradients when training via HFTA, the final fused loss $\mathcal{L}$ needs to be scaled by $B$. Similarly for the fused loss with \emph{sum reduction}:
{
\begin{equation} \label{eqn:fused_loss_sum}
    \mathcal{L} = \sum_{b=0}^{B} \ell_{b}
\end{equation}
}
we can derive that such scaling is no longer needed:
{
\begin{equation} \label{eqn:grad_sum}
%\begin{align*}
    \nabla_{\vec{\theta}_{\beta}} \mathcal{L} 
    = \nabla_{\vec{\theta}_{\beta}} \sum_{b=0}^{B} \ell_{b} = \sum_{b=0}^{B} \nabla_{\vec{\theta}_{\beta}} \ell_{b}
    %&= \frac{1}{B} (\sum_{b=0}^{\beta-1} \nabla_{\vec{\theta}_{\beta}} \ell_{b} + \nabla_{\vec{\theta}_{\beta}} \ell_{\beta} + \sum_{b=\beta+1}^{B} \nabla_{\vec{\theta}_{\beta}} \ell_{b})\\
    %&= \frac{1}{B} (0 + \nabla_{\vec{\theta}_{\beta}} \ell_{\beta} + 0) 
    = \nabla_{\vec{\theta}_{\beta}} \ell_{\beta}
%\end{align*}
\end{equation}
}

In these derivations, no assumption is made on the exact formula of $\ell_{\beta}$, which means such scaling rules are universal to any types of loss functions including regularization.

\section{Convergence} \label{append:converge}

\begin{figure}[t]
    \centering
    \vspace{-0.0cm}
    \includegraphics[width=\linewidth]{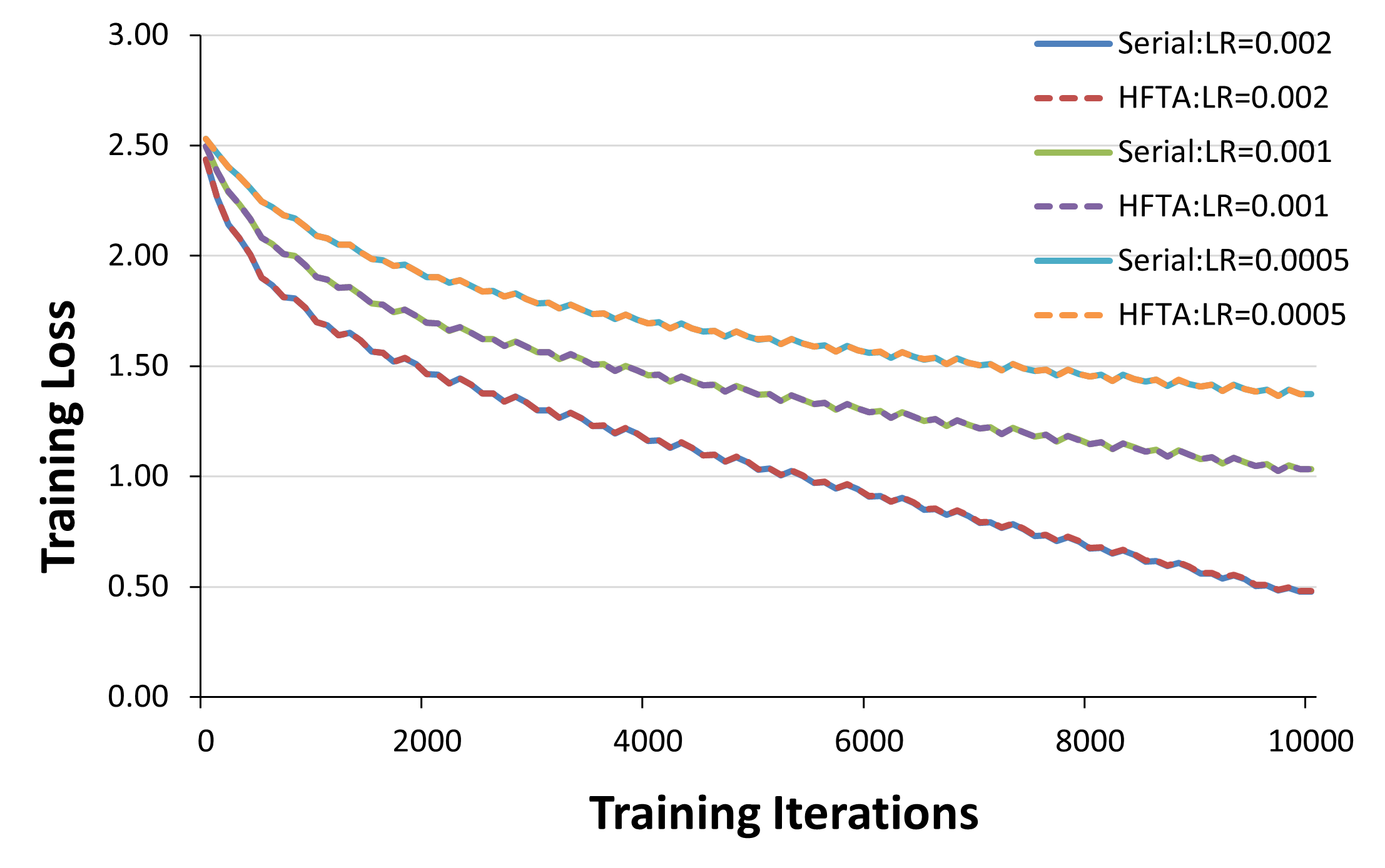} 
    \vspace{-0.0cm}
    \caption{Training loss per iteration when training ResNet-18 on CIFAR-10. \emph{LR} represents the learning rate. \emph{Serial} represents training each model separately, and \emph{HFTA} represents our method.}
    \label{fig:res18_convergence}\vspace{-0.0cm}
    \vspace{-0.0cm}
\end{figure}

Even though HFTA reconstructs the mathematically equivalent gradients for each independently trained model, minor numerical differences can still exist
% since the fused operators and their implementations can be different from the original ones. 
since the order of computations in fused operators can be different from the original ones.
To demonstrate that such numerical differences do not affect the models' original convergence empirically, we train a well-known ResNet-18~\cite{resnet} model on the CIFAR-10~\cite{cifar10} dataset with three different learning rates.\footnote{We provide
the detailed methodology behind this and other experiments on our secondary benchmarks in Appendix~\ref{append:secondary_eval_method}} Figure~\ref{fig:res18_convergence} shows the training-loss-per-iteration curves for 
%both
%the original serial model execution 
training each model independently (solid curves) and collectively as a horizontally fused job via HFTA (dotted curves). 
Since the dotted curves overlap 
%completely 
entirely with the solid ones, we conclude that HFTA-based training maintains exactly the same
% training curves 
convergence as independent model training.

\section{HFHT Design} \label{append:hfht}

Many hyper-parameter tuning algorithms and frameworks (e.g., Ray Tune~\cite{ray_tune}) are built based on the paradigm shown in Algorithm~\ref{algo:hp_tuning}: (i) an \emph{algorithm} (e.g., Hyperband~\cite{hyperband}) proposes a batch of sets of hyper-parameters (e.g., $H$ in Algorithm~\ref{algo:hp_tuning}) to a job \emph{scheduler} that schedules and runs the evaluations (i.e., the training jobs) of these sets of hyper-parameters; (ii) the algorithm then updates (and hopefully improves) itself using the evaluation results (e.g., $R$ in Algorithm~\ref{algo:hp_tuning}, such as validation losses or accuracy) gathered from the output artifacts after those jobs finish. This routine repeats until a certain terminating condition is met while tracking a globally optimal set of hyper-parameters (e.g., $\hat{h}$ in Algorithm~\ref{algo:hp_tuning}).

Therefore, to integrate HFTA with existing hyper-parameter tuning algorithms, we need to modify the interface between the algorithm and the job scheduler (e.g., line~12 in Algorithm~\ref{algo:hp_tuning}). 
When the algorithm proposes a batch of hyper-parameter sets, we should partition these sets before passing the partitions to the scheduler, and ask the scheduler to run each partition as a fused job (e.g., \emph{partition\_and\_fuse()} on line~8 in Algorithm~\ref{algo:hp_tuning}). When the fused jobs finish, we then scatter/``unfuse" the results into their original orders before sending them back to the algorithm (e.g., \emph{unfuse\_and\_reorder()} on line~8 in Algorithm~\ref{algo:hp_tuning}).

\begin{figure}[]
    \centering
    \vspace{-0.0cm}
    \includegraphics[width=\linewidth]{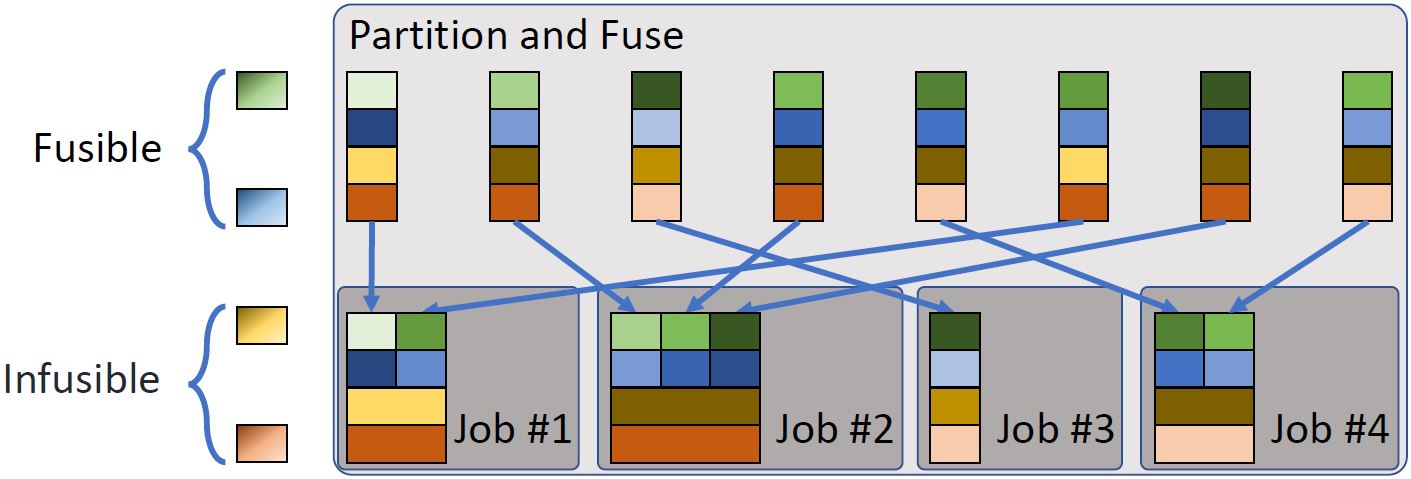} 
    \vspace{-0.0cm}
    \caption{An example showing that the proposed sets of hyper-parameters are partitioned and fused based on their \emph{infusible} hyper-parameters' values.}
    \label{fig:partition_and_fuse}
    \vspace{-0.0cm}
\end{figure}

The aforementioned hyper-parameter set partition can be conducted in different ways. 
In HFHT, we leverage the most straightforward approach. 
Given a bag of different hyper-parameters, we divide them into two categories: \emph{fusible} and \emph{infusible}. 
Being fusible means that different values of such hyper-parameters can be co-evaluated in a fused job (e.g., different learning rates or weight initializers), whereas being infusible means that different values of such hyper-parameters can lead to significant difference in the operator types and shapes (e.g., batch size) or change the model architecture completely (e.g., MobileNet V2 vs. V3-Large). 
Once the users specify which hyper-parameters are fusible and infusible, as we show in Figure~\ref{fig:partition_and_fuse}, HFHT can utilize this information to partition arbitrary sets of hyper-parameter. 
Afterwards, each partition only has a single value for each infusible hyper-parameter.

To integrate our hardware-based sharing baselines (i.e., \emph{concurrent}, \emph{MPS}, and \emph{MIG})\footnote{These baselines are elaborated in Section~\ref{sec:eval_method}.} with existing hyper-parameter tuning algorithms, since these hardware features treat each job as a separate process, there is no scheduling constraint on the values of the hyper-parameters. 
Therefore, we simply ask the scheduler to run these jobs as concurrently as the GPU allows.

\begin{algorithm}[]
  %\scriptsize
  \caption{Integrating \emph{HFTA}, \emph{concurrent}, \emph{MPS} and \emph{MIG} with a General Hyper-parameter Tuning Routine}\label{algo:hp_tuning}
  \begin{algorithmic}[1]
    \STATE $\hat{h} \gets \textsc{nul}$ \COMMENT{The best set of hyper-parameters.}
    \STATE $\hat{r} \gets \textsc{nul}$ \COMMENT{The best training result.}
    \STATE $H \gets \emptyset$ \COMMENT{A batch of arbitrary sets of hyper-parameters.}
    \STATE $R \gets \emptyset$ \COMMENT{Training results corresponding to $H$.}
    \WHILE{terminating condition}
      \STATE $H \gets$ algorithm.propose()
      \IF{\emph{HFTA} enabled}
        \STATE $\begin{aligned}R \gets &\text{unfuse\_and\_reorder}(\\ 
        &\text{\quad scheduler.schedule\_and\_run}(\\
        &\text{\quad\quad partition\_and\_fuse}(H)))\end{aligned}$
      \ELSIF{\emph{concurrent} or \emph{MPS} or \emph{MIG} enabled}
        \STATE $R \gets$ scheduler.schedule\_and\_run\_in\_parallel($H$)
      \ELSE
        \STATE $R \gets$ scheduler.schedule\_and\_run($H$)
      \ENDIF
      \STATE $\hat{r}$, $\hat{h} \gets$ select\_best(($R$, $H$), ($\hat{r}$, $\hat{h}$))
      \STATE algorithm.update($H$, $R$)
    \ENDWHILE
  \end{algorithmic}
\end{algorithm}

By default, the tuning algorithms only concern about the impact of each hyper-parameter on the training results' quality (e.g., accuracy). However, with the introduction of HFTA, each hyper-parameter now is additionally correlated with the hardware utilization of the underlying tuning jobs in terms of how well the fusion opportunities are exposed. Thus, for a specific iteration during tuning, if the algorithm either (1) only proposes a single set of hyper-parameters, or (2) proposes sets that don't share any common value for the infusible hyper-parameters, there would be few opportunities for HFTA to exploit. In other words, \emph{the exact design of the tuning algorithm itself can be either HFTA-``friendly" or ``unfriendly"}. For example, as we show in Section~\ref{sec:hfht_cost_saving}, random search is a more HFTA-``friendly" algorithm than Hyperband. Therefore, as one direction for our future work, we will investigate how to adjust existing tuning algorithms to extract the most benefits offered by HFTA.

\section{DCGM Metrics} \label{append:dcgm}

\begin{table}[]
\centering
\caption{DCGM Metric Names, Field Identifier Macros and IDs} \label{tab:dcgm_metrics}
\scriptsize
\begin{tabular}{l|ll}
\toprule
Name                   & Field Identifier Macro                &  ID          \\\midrule
\texttt{sm\_active}    & DCGM\_FI\_PROF\_SM\_ACTIVE           & 1002                          \\
\texttt{sm\_occupancy}  & DCGM\_FI\_PROF\_SM\_OCCUPANCY        & 1003                          \\
\texttt{tensor\_active} & DCGM\_FI\_PROF\_PIPE\_TENSOR\_ACTIVE & 1004                          \\
GPU Utilization        & DCGM\_FI\_DEV\_GPU\_UTIL              & 203                           \\
\bottomrule
\end{tabular}
\end{table}

The \texttt{sm\_active}, \texttt{sm\_occupancy} and \texttt{tensor\_active} performance counters are measured through DCGM \cite{dcgm}. Their field identifier macros and IDs are listed in Table~\ref{tab:dcgm_metrics}.

The \texttt{active} performance counters measure the fraction of cycles that a hardware unit is active averaged across all hardware units of the same type on the entire GPU, which we refer to as ``temporal" utilization. 
We highlight that an SM being active only implies that there are resident warps assigned to the SM, and does not imply that all of its functional units are active. 
For example, an SM is considered active even during the cycles when all of its resident warps are stalled due to outstanding memory requests. 
Therefore, a high \texttt{sm\_active} is a necessary but insufficient condition for good GPU utilization. 

The \texttt{sm\_occupancy} performance counter is the ratio of the achieved number of resident warps on an SM over the maximum number of warps that the SM can hold (e.g., 64 warps/SM on V100~\cite{volta_whitepaper}). 
A higher \texttt{sm\_occupancy} for the same kernel means that the kernel occupies more resources related to warp placement (e.g., register file and shared memory) on each SM. 
Hence, we refer to \texttt{sm\_occupancy} as ``spatial" utilization. 
A high \texttt{sm\_occupancy} combined with a high \texttt{sm\_active} can (but not necessarily) indicate good GPU utilization.
%when the kernel is limited by memory bandwidth. 
This is because, the more resident warps an SM has, the more likely the SM can hide the cost of memory requests or other instructions from one warp with the computation of another warp. For kernels limited by memory bandwidth, although a higher \texttt{sm\_occupancy} generally leads to better performance, an overly high \texttt{sm\_occupancy} might hurt performance in rare exceptions (e.g., due to more frequent cache thrashing).

Please refer to the DCGM Library API Reference Manual \cite{dcgm_field_identifiers} for their precise definitions.

\section{Additional Evaluation Statistics} \label{append:additional_eval_data}

% Peak throughput table
\begin{table}[t]
    \centering
    \caption{The peak training throughput speedups of HFTA over the baselines.} \label{tab:gpus_peak_speedups_fp32_amp}
    \scriptsize
    \begin{tabular}{lll|lll}
        \toprule
        \multicolumn{3}{l|}{Benchmark} & \begin{tabular}[c]{@{}l@{}}PointNet\\ Cls.\end{tabular} & \begin{tabular}[c]{@{}l@{}}PointNet \\ Seg.\end{tabular} & DCGAN                                                                                                 \\ \midrule
        \multirow{6}{*}{V100}          & \multirow{3}{*}{FP32}     & serial                    & \gpuvFpPeakHFTAvsSerialCls      & \gpuvFpPeakHFTAvsSerialSeg      & \gpuvFpPeakHFTAvsSerialDCGAN      \\
                                       &                           & concurrent                & \gpuvFpPeakHFTAvsConcurrentCls  & \gpuvFpPeakHFTAvsConcurrentSeg  & \gpuvFpPeakHFTAvsConcurrentDCGAN  \\
                                       &                           & MPS                       & \gpuvFpPeakHFTAvsMPSCls         & \gpuvFpPeakHFTAvsMPSSeg         & \gpuvFpPeakHFTAvsMPSDCGAN         \\ \cline{2-6}
                                       & \multirow{3}{*}{AMP}      & serial                    & \gpuvAmpPeakHFTAvsSerialCls     & \gpuvAmpPeakHFTAvsSerialSeg     & \gpuvAmpPeakHFTAvsSerialDCGAN     \\
                                       &                           & concurrent                & \gpuvAmpPeakHFTAvsConcurrentCls & \gpuvAmpPeakHFTAvsConcurrentSeg & \gpuvAmpPeakHFTAvsConcurrentDCGAN \\
                                       &                           & MPS                       & \gpuvAmpPeakHFTAvsMPSCls        & \gpuvAmpPeakHFTAvsMPSSeg        & \gpuvAmpPeakHFTAvsMPSDCGAN        \\ \hline

        \multirow{6}{*}{\begin{tabular}[c]{@{}l@{}}RTX\\ 6000\end{tabular}}
                                       & \multirow{3}{*}{FP32}     & serial                    & \gpurFpPeakHFTAvsSerialCls      & \gpurFpPeakHFTAvsSerialSeg      & \gpurFpPeakHFTAvsSerialDCGAN      \\
                                       &                           & concurrent                & \gpurFpPeakHFTAvsConcurrentCls  & \gpurFpPeakHFTAvsConcurrentSeg  & \gpurFpPeakHFTAvsConcurrentDCGAN  \\
                                       &                           & MPS                       & \gpurFpPeakHFTAvsMPSCls         & \gpurFpPeakHFTAvsMPSSeg         & \gpurFpPeakHFTAvsMPSDCGAN         \\ \cline{2-6}
                                       & \multirow{3}{*}{AMP}      & serial                    & \gpurAmpPeakHFTAvsSerialCls     & \gpurAmpPeakHFTAvsSerialSeg     & \gpurAmpPeakHFTAvsSerialDCGAN     \\
                                       &                           & concurrent                & \gpurAmpPeakHFTAvsConcurrentCls & \gpurAmpPeakHFTAvsConcurrentSeg & \gpurAmpPeakHFTAvsConcurrentDCGAN \\
                                       &                           & MPS                       & \gpurAmpPeakHFTAvsMPSCls        & \gpurAmpPeakHFTAvsMPSSeg        & \gpurAmpPeakHFTAvsMPSDCGAN        \\ \hline

        \multirow{8}{*}{A100}          & \multirow{4}{*}{FP32}     & serial                    & \gpuaFpPeakHFTAvsSerialCls      & \gpuaFpPeakHFTAvsSerialSeg      & \gpuaFpPeakHFTAvsSerialDCGAN      \\
                                       &                           & concurrent                & \gpuaFpPeakHFTAvsConcurrentCls  & \gpuaFpPeakHFTAvsConcurrentSeg  & \gpuaFpPeakHFTAvsConcurrentDCGAN  \\
                                       &                           & MPS                       & \gpuaFpPeakHFTAvsMPSCls         & \gpuaFpPeakHFTAvsMPSSeg         & \gpuaFpPeakHFTAvsMPSDCGAN         \\
                                       &                           & MIG                       & \gpuaFpPeakHFTAvsMIGCls         & \gpuaFpPeakHFTAvsMIGSeg         & \gpuaFpPeakHFTAvsMIGDCGAN         \\ \cline{2-6}
                                       & \multirow{4}{*}{AMP}      & serial                    & \gpuaAmpPeakHFTAvsSerialCls     & \gpuaAmpPeakHFTAvsSerialSeg     & \gpuaAmpPeakHFTAvsSerialDCGAN     \\
                                       &                           & concurrent                & \gpuaAmpPeakHFTAvsConcurrentCls & \gpuaAmpPeakHFTAvsConcurrentSeg & \gpuaAmpPeakHFTAvsConcurrentDCGAN \\
                                       &                           & MPS                       & \gpuaAmpPeakHFTAvsMPSCls        & \gpuaAmpPeakHFTAvsMPSSeg        & \gpuaAmpPeakHFTAvsMPSDCGAN        \\
                                       &                           & MIG                       & \gpuaAmpPeakHFTAvsMIGCls        & \gpuaAmpPeakHFTAvsMIGSeg        & \gpuaAmpPeakHFTAvsMIGDCGAN        \\
        \bottomrule
    \end{tabular}
\end{table}

% Throughput table with the same B
\begin{table}[t]
    \centering
    \caption{The maximum training throughput speedups of HFTA over the baselines given the same number of models sharing one GPU.} \label{tab:gpus_max_speedups_fixing_B}
    \scriptsize
    \begin{tabular}{lll|lll}
        \toprule
        \multicolumn{3}{l|}{Benchmark} & \begin{tabular}[c]{@{}l@{}}PointNet\\ Cls.\end{tabular} & \begin{tabular}[c]{@{}l@{}}PointNet \\ Seg.\end{tabular} & DCGAN                                                                                                    \\ \midrule
        \multirow{6}{*}{V100}          & \multirow{2}{*}{FP32}     & concurrent                & \gpuvFpSameBHFTAvsConcurrentCls  & \gpuvFpSameBHFTAvsConcurrentSeg  & \gpuvFpSameBHFTAvsConcurrentDCGAN  \\
                                       &                           & MPS                       & \gpuvFpSameBHFTAvsMPSCls         & \gpuvFpSameBHFTAvsMPSSeg         & \gpuvFpSameBHFTAvsMPSDCGAN         \\ \cline{2-6}
                                       & \multirow{2}{*}{AMP}      & concurrent                & \gpuvAmpSameBHFTAvsConcurrentCls & \gpuvAmpSameBHFTAvsConcurrentSeg & \gpuvAmpSameBHFTAvsConcurrentDCGAN \\
                                       &                           & MPS                       & \gpuvAmpSameBHFTAvsMPSCls        & \gpuvAmpSameBHFTAvsMPSSeg        & \gpuvAmpSameBHFTAvsMPSDCGAN        \\ \hline

        \multirow{6}{*}{\begin{tabular}[c]{@{}l@{}}RTX\\ 6000\end{tabular}}
                                       & \multirow{2}{*}{FP32}     & concurrent                & \gpurFpSameBHFTAvsConcurrentCls  & \gpurFpSameBHFTAvsConcurrentSeg  & \gpurFpSameBHFTAvsConcurrentDCGAN  \\
                                       &                           & MPS                       & \gpurFpSameBHFTAvsMPSCls         & \gpurFpSameBHFTAvsMPSSeg         & \gpurFpSameBHFTAvsMPSDCGAN         \\ \cline{2-6}
                                       & \multirow{2}{*}{AMP}      & concurrent                & \gpurAmpSameBHFTAvsConcurrentCls & \gpurAmpSameBHFTAvsConcurrentSeg & \gpurAmpSameBHFTAvsConcurrentDCGAN \\
                                       &                           & MPS                       & \gpurAmpSameBHFTAvsMPSCls        & \gpurAmpSameBHFTAvsMPSSeg        & \gpurAmpSameBHFTAvsMPSDCGAN        \\ \hline

        \multirow{8}{*}{A100}          & \multirow{3}{*}{FP32}     & concurrent                & \gpuaFpSameBHFTAvsConcurrentCls  & \gpuaFpSameBHFTAvsConcurrentSeg  & \gpuaFpSameBHFTAvsConcurrentDCGAN  \\
                                       &                           & MPS                       & \gpuaFpSameBHFTAvsMPSCls         & \gpuaFpSameBHFTAvsMPSSeg         & \gpuaFpSameBHFTAvsMPSDCGAN         \\
                                       &                           & MIG                       & \gpuaFpSameBHFTAvsMIGCls         & \gpuaFpSameBHFTAvsMIGSeg         & \gpuaFpSameBHFTAvsMIGDCGAN         \\ \cline{2-6}
                                       & \multirow{3}{*}{AMP}      & concurrent                & \gpuaAmpSameBHFTAvsConcurrentCls & \gpuaAmpSameBHFTAvsConcurrentSeg & \gpuaAmpSameBHFTAvsConcurrentDCGAN \\
                                       &                           & MPS                       & \gpuaAmpSameBHFTAvsMPSCls        & \gpuaAmpSameBHFTAvsMPSSeg        & \gpuaAmpSameBHFTAvsMPSDCGAN        \\
                                       &                           & MIG                       & \gpuaAmpSameBHFTAvsMIGCls        & \gpuaAmpSameBHFTAvsMIGSeg        & \gpuaAmpSameBHFTAvsMIGDCGAN        \\
        \bottomrule
    \end{tabular}
\end{table}

\begin{table}[t]
    \centering
    \caption{The maximum speedups of AMP training over FP32.} \label{tab:max_speedups_amp_over_fp32}
    \scriptsize
    \begin{tabular}{ll|lll}
        \toprule
        \multicolumn{2}{c|}{Benchmark} & \begin{tabular}[c]{@{}l@{}}PointNet\\ Classification\end{tabular} & \multicolumn{1}{c}{\begin{tabular}[c]{@{}c@{}}PointNet\\ Segmentation\end{tabular}} & DCGAN                                                   \\ \midrule
        \multirow{4}{*}{V100}          & \emph{serial}             & \gpuvAmpvsFpSerialCls                         & \gpuvAmpvsFpSerialSeg     & \gpuvAmpvsFpSerialDCGAN     \\
                                       & \emph{concurrent}         & \gpuvAmpvsFpConcurrentCls                     & \gpuvAmpvsFpConcurrentSeg & \gpuvAmpvsFpConcurrentDCGAN \\
                                       & \emph{MPS}                & \gpuvAmpvsFpMPSCls                            & \gpuvAmpvsFpMPSSeg        & \gpuvAmpvsFpMPSDCGAN        \\
                                       & HFTA                      & \gpuvAmpvsFpHFTACls                           & \gpuvAmpvsFpHFTASeg       & \gpuvAmpvsFpHFTADCGAN       \\ \midrule
        \multirow{4}{*}{RTX6000}       & \emph{serial}             & \gpurAmpvsFpSerialCls                         & \gpurAmpvsFpSerialSeg     & \gpurAmpvsFpSerialDCGAN     \\
                                       & \emph{concurrent}         & \gpurAmpvsFpConcurrentCls                     & \gpurAmpvsFpConcurrentSeg & \gpurAmpvsFpConcurrentDCGAN \\
                                       & \emph{MPS}                & \gpurAmpvsFpMPSCls                            & \gpurAmpvsFpMPSSeg        & \gpurAmpvsFpMPSDCGAN        \\
                                       & HFTA                      & \gpurAmpvsFpHFTACls                           & \gpurAmpvsFpHFTASeg       & \gpurAmpvsFpHFTADCGAN       \\ \midrule
        \multirow{5}{*}{A100}          & \emph{serial}             & \gpuaAmpvsFpSerialCls                         & \gpuaAmpvsFpSerialSeg     & \gpuaAmpvsFpSerialDCGAN     \\
                                       & \emph{concurrent}         & \gpuaAmpvsFpConcurrentCls                     & \gpuaAmpvsFpConcurrentSeg & \gpuaAmpvsFpConcurrentDCGAN \\
                                       & \emph{MPS}                & \gpuaAmpvsFpMPSCls                            & \gpuaAmpvsFpMPSSeg        & \gpuaAmpvsFpMPSDCGAN        \\
                                       & \emph{MIG}                & \gpuaAmpvsFpMIGCls                            & \gpuaAmpvsFpMIGSeg        & \gpuaAmpvsFpMIGDCGAN        \\
                                       & HFTA                      & \gpuaAmpvsFpHFTACls                           & \gpuaAmpvsFpHFTASeg       & \gpuaAmpvsFpHFTADCGAN       \\
        \bottomrule
    \end{tabular}
\end{table}

%In order to 
To facilitate the reading of the results from our GPU experiments in Figure~\ref{fig:speedups_gpus}, we summarize the comparison from different angles between HFTA and the baselines into three tables.

Table~\ref{tab:gpus_peak_speedups_fp32_amp} shows the peak training throughput comparison between HFTA and the baselines. It is important to highlight that, for both \emph{MPS} and \emph{concurrent}, the training throughput could decrease as we increase the number of models sharing the same GPU (due to host resource contention). Therefore, the ``peak'' is determined by the highest possible throughput instead of the largest number of models that the GPU can fit (which might or might not lead to the highest throughput). Unlike Table~\ref{tab:gpus_peak_speedups}, the results here are split between FP32 and AMP to demonstrate how well HFTA performs for each %type of training.
training type.

Table~\ref{tab:gpus_max_speedups_fixing_B} shows the maximum training throughput speedups of HFTA over the baselines, given the same number of models sharing the same GPU. The maximum is picked by varying the number of models sharing the same GPU and finding the largest performance gap between HFTA and the baselines. This helps to isolate the benefits of better SMs and TCs utilization from the benefits of better memory utilization when training via HFTA.

Table~\ref{tab:max_speedups_amp_over_fp32} shows the maximum training throughput speedups of AMP over FP32 for both HFTA and the baselines. The maximum here is also picked by varying the number of models (except for \emph{serial}, which always only runs one model per GPU) and finding the largest performance gap between FP32 and AMP. This helps to demonstrate that HFTA is more efficient in utilizing advanced hardware compute units such as TCs.

\begin{figure}[]
    \centering
    \includegraphics[width=0.9\linewidth]{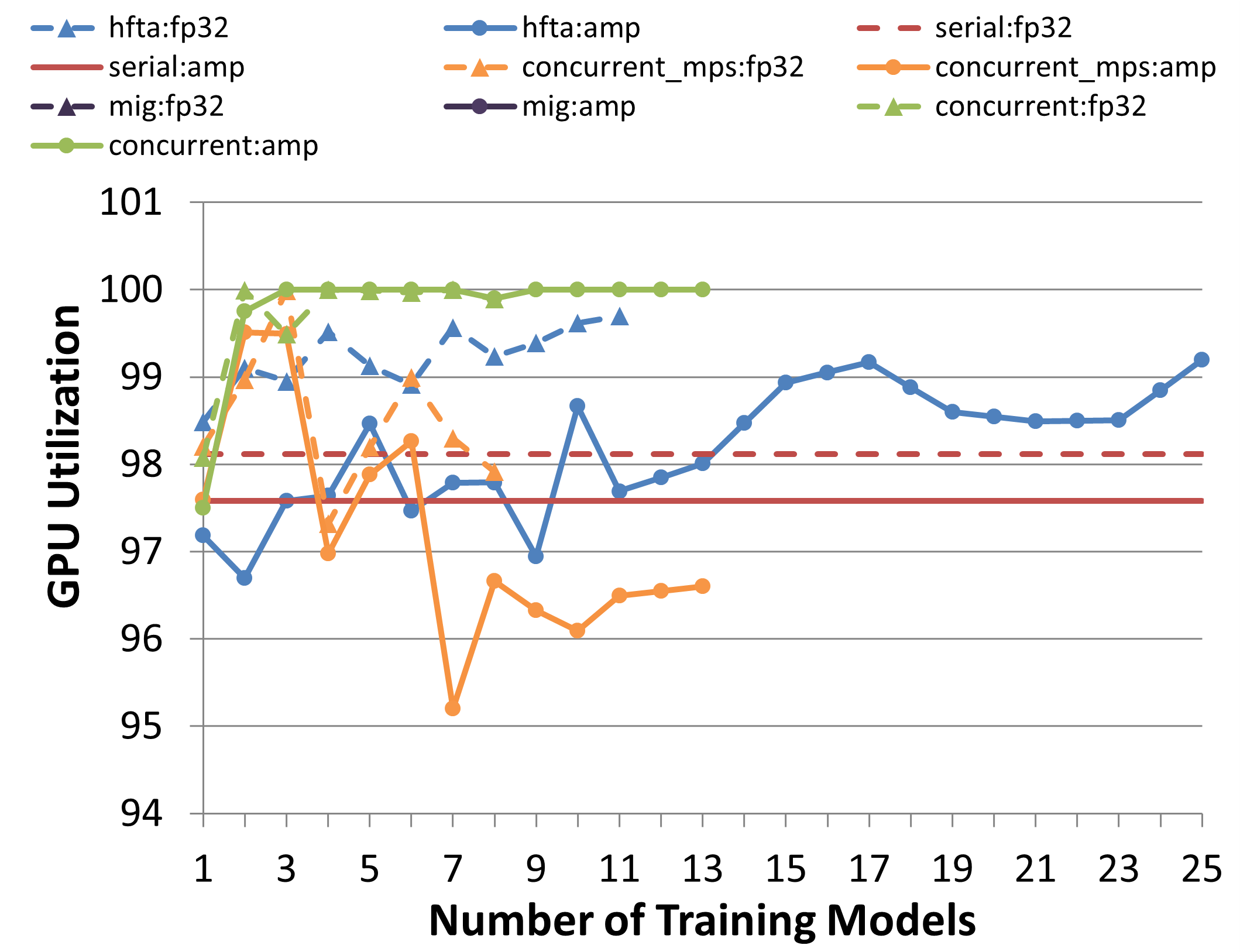}
    \caption{\texttt{nvidia-smi}-defined ``GPU utilization'' for PointNet classification task on A100.}
    \label{fig:gpu_util_pointnet_cls_a100}
\end{figure}

Similar to Figure~\ref{fig:sm_active_pointnet_cls_a100}, \ref{fig:sm_occupancy_pointnet_cls_a100} and \ref{fig:tensor_active_pointnet_cls_a100}, Figure~\ref{fig:gpu_util_pointnet_cls_a100} plots the \texttt{nvidia-smi}-defined ``GPU utilization''~\cite{nvidia_smi} for PointNet classification task training on the A100 GPU. Contrary to a popular belief~\cite{aws_gpu_utilization, fastai_gpu_utilization}, we observe that the \texttt{nvidia-smi}-defined ``GPU utilization'' can be sometimes a weak utilization indicator, since the curves in Figure~\ref{fig:gpu_util_pointnet_cls_a100} appear rather noisy and do not follow the trends of throughput improvements in Figures~\ref{fig:speedups_pointnet_cls_a100} or any hardware counters' trend in Figure~\ref{fig:sm_active_pointnet_cls_a100}, ~\ref{fig:sm_occupancy_pointnet_cls_a100} or ~\ref{fig:tensor_active_pointnet_cls_a100}.

\begin{figure*}[t]
    \centering
    \vspace{-0.3cm}
    \begin{subfigure}[]{\linewidth}
        \centering
        \includegraphics[width=0.9\linewidth]{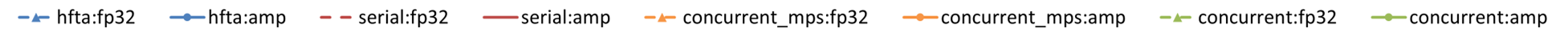}
    \end{subfigure}
    \hfill
    \begin{subfigure}[]{0.32\linewidth}
        \centering
        \includegraphics[width=\linewidth]{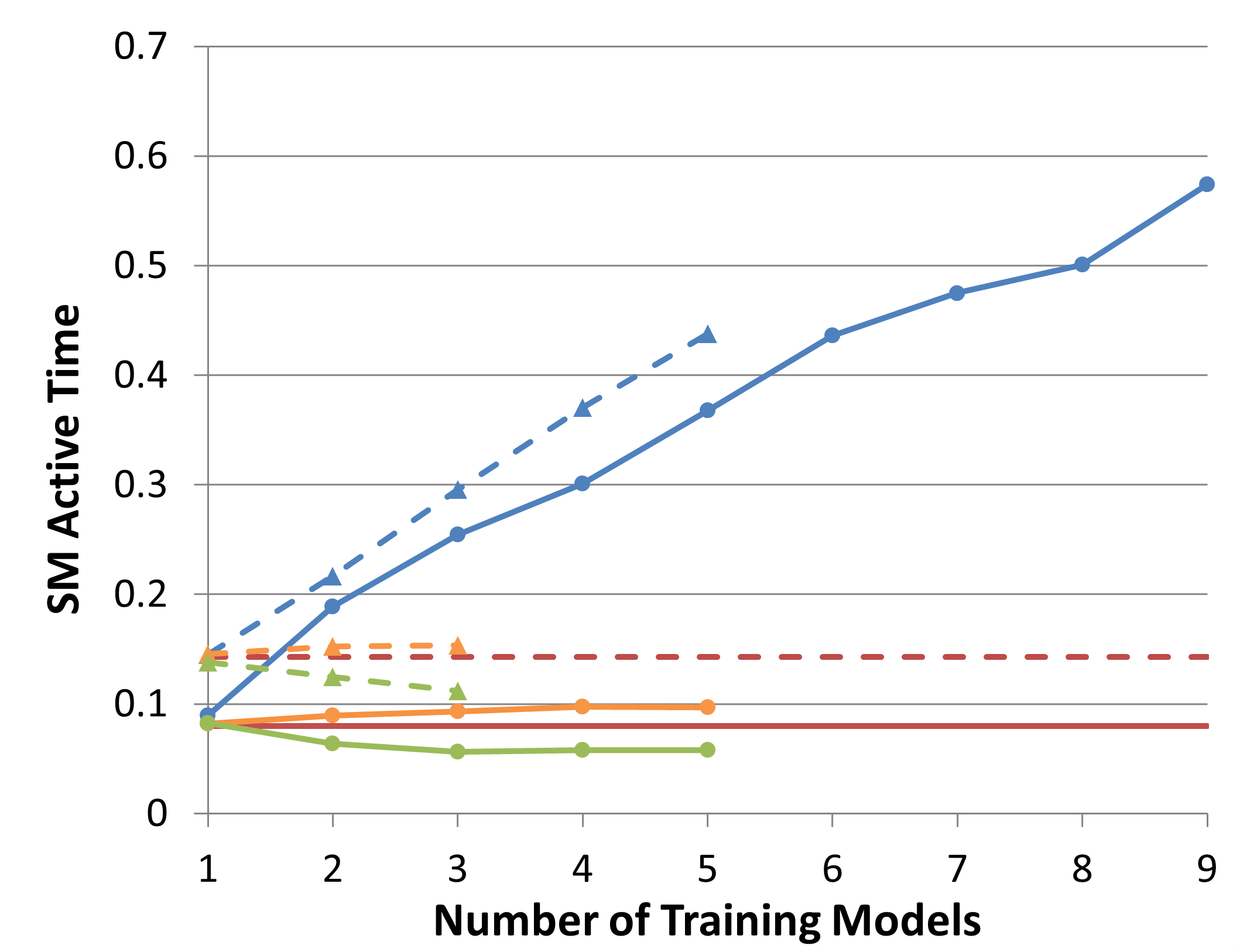}
        \caption{\texttt{sm\_active}}
        \label{fig:sm_active_pointnet_cls_v100}
    \end{subfigure}
    \hfill
    \begin{subfigure}[]{0.32\linewidth}
        \centering
        \includegraphics[width=\linewidth]{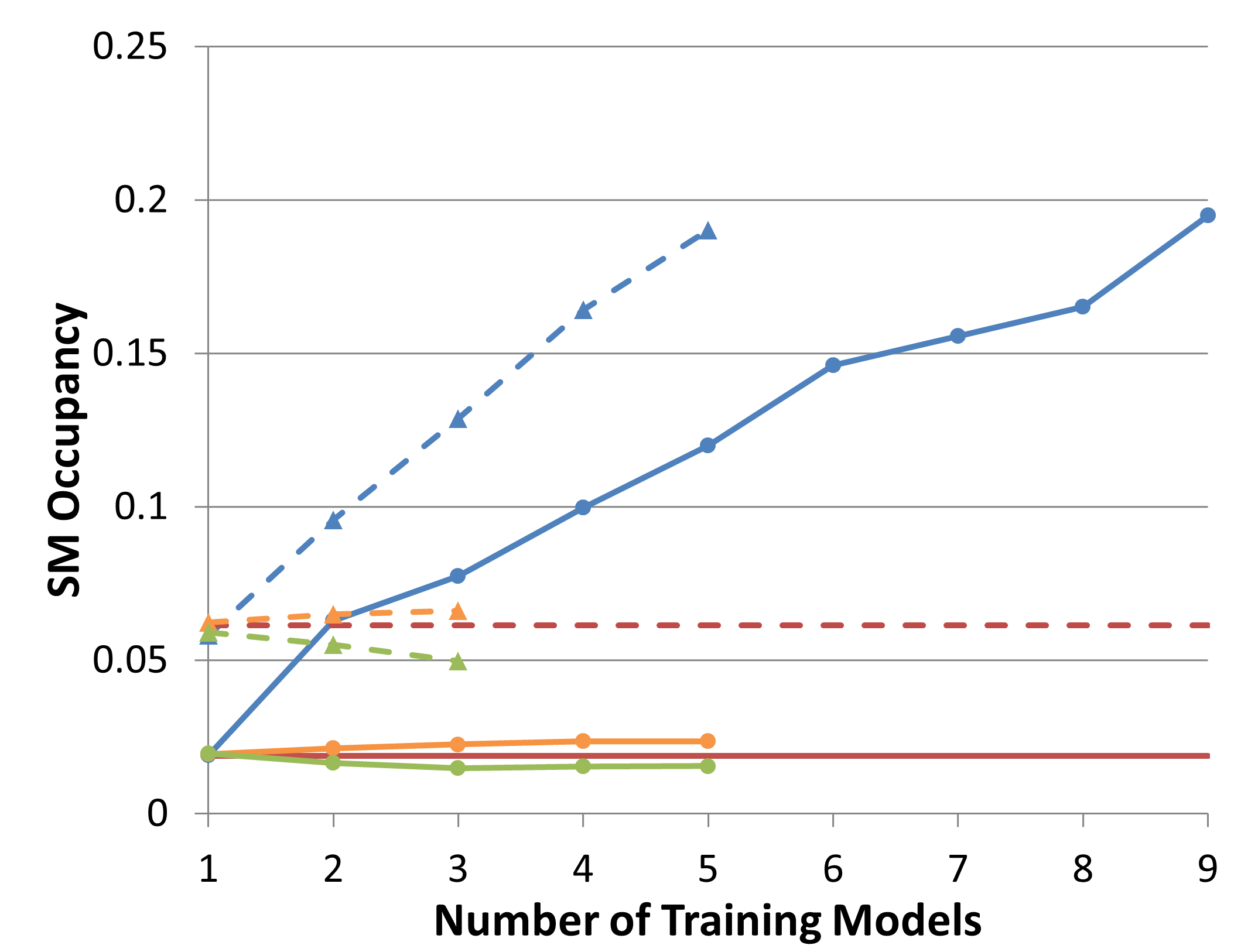}
        \caption{\texttt{sm\_occupancy}}
        \label{fig:sm_occupancy_pointnet_cls_v100}
    \end{subfigure}
    \hfill
    \begin{subfigure}[]{0.32\linewidth}
        \centering
        \includegraphics[width=\linewidth]{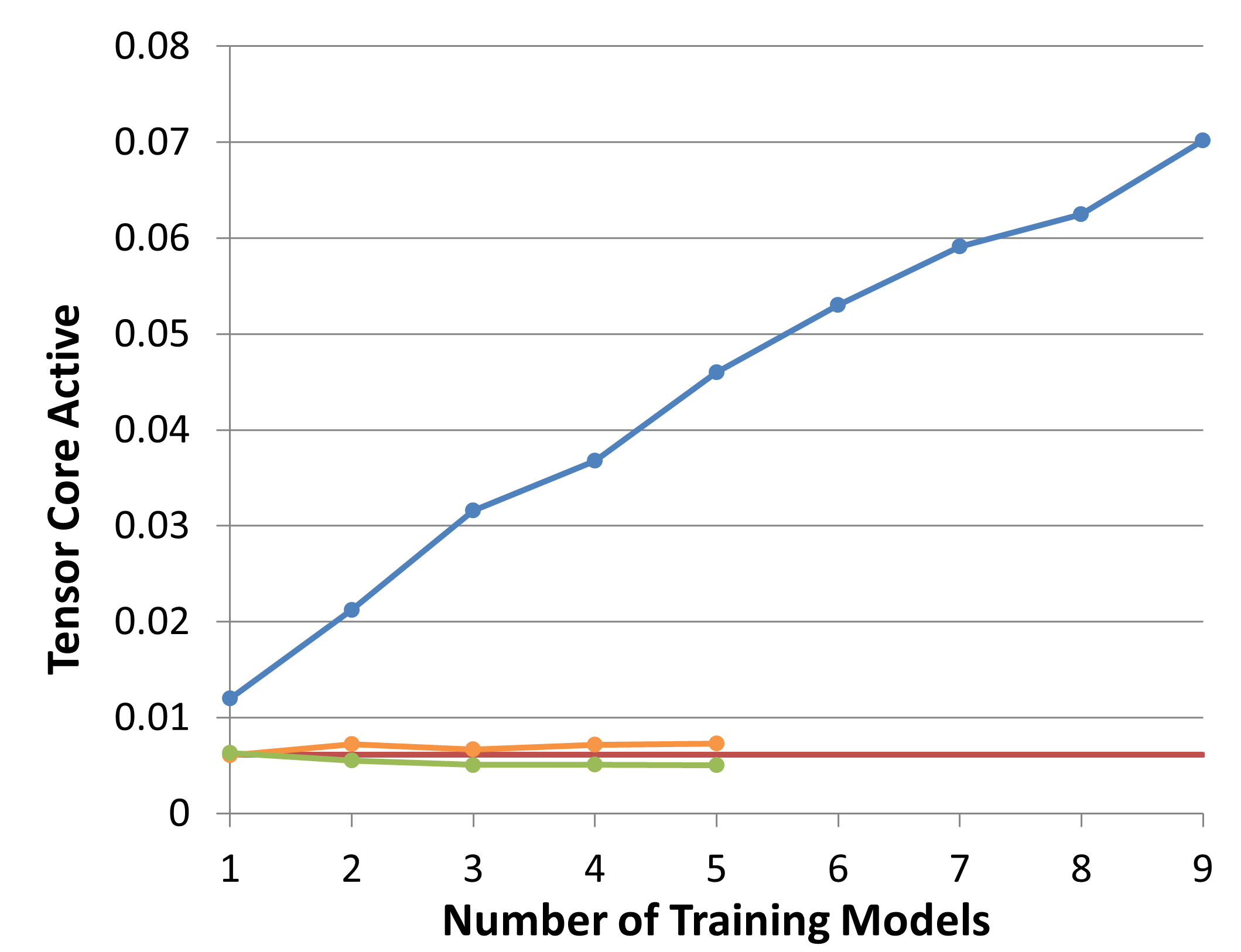}
        \caption{\texttt{tensor\_active}}
        \label{fig:tensor_active_pointnet_cls_v100}
    \end{subfigure}
    \vspace{-0.3cm}
    \caption{The hardware performance counters for PointNet classification task as we increase the number of models sharing the same V100.}
    \label{fig:perf_counters_pointnet_cls_v100}
    \vspace{-0.3cm}
\end{figure*}

Similar to Figure~\ref{fig:perf_counters_pointnet_cls_a100}, Figure~\ref{fig:perf_counters_pointnet_cls_v100} plots the \texttt{sm\_active}, \texttt{sm\_occupancy}, and \texttt{tensor\_active} of HFTA and the baselines as we increase the number of models sharing the same V100 GPU. 
In addition to the observations we already present in Section~\ref{sec:gpu_perf_analysis}, we also observe that the \emph{serial} baselines' hardware utilization is lower on A100 than on V100. 
Therefore, Figure~\ref{fig:perf_counters_pointnet_cls_v100} provides empirical evidence to support our argument in Section~\ref{sec:inefficiency_repetitive_training_workloads} and Section~\ref{sec:eval_gpus} that newer GPU generations suffer more significantly from the hardware under-utilization of repetitive single-accelerator training workloads.

\section{Additional Methodology Details} \label{append:secondary_eval_method}

\subsection{Secondary Benchmarks}

To evaluate the HFTA's general effectiveness in extracting hardware utilization for conventional models, we further include the following models and tasks as our secondary benchmarks.

\emph{ResNet}~\cite{resnet} and \emph{MobileNet}~\cite{mobilenets, mobilenets_v2, mobilenetv3l} are two classes of convolutional neural networks that are generally used (or as backbones) in computer vision (CV) tasks. Both classes contain many variants of the models. As our benchmarks, we train ResNet-18 and MobileNetV3-Large~\cite{mobilenetv3l} to perform image classification tasks, though HFTA can be applied on the other variants and tasks as well. We leverage an implementation of ResNet-18 from PyTorch's official examples~\cite{pytorch_examples}, and an open-sourced PyTorch re-implementation~\cite{mobilenet_v3_implementation} of MobileNetV3-Large. Both models are trained on the CIFAR-10~\cite{cifar10} dataset. To evaluate HFTA with different training configurations, ResNet-18 is trained using the Adadelta~\cite{adadelta} optimizer with a batch size of 128, whereas MobileNetV3-Large is trained using the Adam~\cite{adam} optimizer with a batch size of 1024.

\emph{Transformer}~\cite{transformer} and \emph{BERT}~\cite{bert, bert_medium} are two classes of attention-based language models that are generally used in natural language processing (NLP) tasks. For Transformer, we leverage an implementation from PyTorch's official examples~\cite{pytorch_examples} and configure our variant to have 2 encoder layers with 2 attention heads and the hidden size of 128 (similar to BERT-Tiny~\cite{bert_medium} in the parameter size). We train our Transformer variant for the language modeling (LM) task. For BERT, we leverage an open-sourced PyTorch re-implementation~\cite{bert_implementation} and select the BERT-Medium~\cite{bert_medium} variant to perform the masked LM task. Both models are trained on the WikiText-2~\cite{wikitext} dataset using the Adadelta optimizer with the batch size and sequence length of 32.

We perform the experiments of all four secondary benchmarks on two ML accelerators: the V100 GPU and TPU v3. 
We list the detailed specifications of our experimental setup in Table~\ref{tab:platform_spec}. 
We provide the corresponding evaluation results in Appendix~\ref{append:secondary_eval}.

\subsection{Convergence}

To empirically prove that HFTA does not affect convergence, using three different learning rates ($0.0005$, $0.001$, $0.002$), we train ResNet-18 on V100 with the CIFAR-10 dataset using the Adadelta optimizer with a batch size of 1000. The corresponding validation results are provided in Appendix~\ref{append:converge}. 

\subsection{HFHT}

To demonstrate that HFTA can significantly improve the hardware utilization for existing hyper-parameter tuning algorithms via HFHT, we construct the following four end-to-end hyper-parameter tuning workloads: for each of the PointNet and MobileNet classification tasks, we tune eight different hyper-parameters using each of the Hyperband~\cite{hyperband} and random search~\cite{random_search_for_hyper_parameter_optimization} algorithms. We leverage an open-sourced implementation of the Hyperband algorithm~\cite{hyperband_implementation} and adopt it into HFHT, and we implement the random search algorithm from scratch. We evaluate HFHT across different algorithm settings listed in Table~\ref{tab:hfht_algorithm_settings}, and we list the hyper-parameters that we tune for each task in Table~\ref{tab:hfht_hyperparameter_list}. We provide the corresponding evaluation results in Section~\ref{sec:hfht_cost_saving}.

% \usepackage{multirow}
\begin{table}[t]
\caption{The algorithm settings for all four end-to-end hyper-parameter tuning workloads. A \emph{set} refers to a set of hyper-parameters. An \emph{epoch} refers to an epoch of training (i.e., the training loop iterates through the entire dataset once). A \emph{round} refers to a single round of successive halving~\cite{successive_halving} within Hyperband.} \label{tab:hfht_algorithm_settings}
\scriptsize
\begin{tabular}{l|l|l|l}
\toprule
Algorithm                                                                & \begin{tabular}[c]{@{}l@{}}Algorithm\\ Settings\end{tabular}                                        & PointNet & MobileNet \\ \midrule
\multirow{3}{*}{Hyperband}                                               & \begin{tabular}[c]{@{}l@{}}Maximum Epochs Allowed \\ for a Single Set ($R$)\end{tabular}            & 250      & 81        \\ \cline{2-4} 
                                                                         & \begin{tabular}[c]{@{}l@{}}Inverse of  the Fraction of Sets \\ Discarded in Each Round ($\eta$)\end{tabular} & 5        & 3         \\ \cline{2-4} 
                                                                         & Last Rounds to Skip                                                                                 & 1        & 2         \\ \midrule
\multirow{2}{*}{\begin{tabular}[c]{@{}l@{}}Random\\ Search\end{tabular}} & Total Sets                                                                                          & 60       & 50        \\ \cline{2-4} 
                                                                         & Epochs per Set                                                                                      & 25       & 20        \\ 
\bottomrule
\end{tabular}
\end{table}

\begin{table}[]
\caption{The list of hyper-parameters that are tuned for the PointNet and MobileNet classification tasks. ``[]" represents a continuous closed interval, whereas ``\{\}" represents a discrete set.} \label{tab:hfht_hyperparameter_list}
\scriptsize
\begin{tabular}{l|l|l|l}
\toprule
Hyper-parameter                                                             & Fusible & Range                                                                                       & Task(s)  \\ \midrule
Learning Rate                                                               & Yes     & [0.0001, 0.01]                                                                              & Both      \\ \hline
Adam's $\beta_1$                                                            & Yes     & [0.001, 0.999]                                                                              & Both      \\ \hline
Adam's $\beta_2$                                                            & Yes     & [0.001, 0.999]                                                                              & Both      \\ \hline
Weight Decay                                                                & Yes     & [0.0, 0.5]                                                                                  & Both      \\ \hline
\begin{tabular}[c]{@{}l@{}}Factor of \\ Learning \\ Rate Decay\end{tabular} & Yes     & [0.1, 0.9]                                                                                  & Both      \\ \hline
\begin{tabular}[c]{@{}l@{}}Period of \\ Learning \\ Rate Decay\end{tabular} & Yes     & \{5, 10, 20, 40\}                                                                           & Both      \\ \hline
Batch Size                                                                  & No      & \begin{tabular}[c]{@{}l@{}}PointNet: \{8, 16, 32\}\\ MobileNet: \{1024, 2048\}\end{tabular} & Both      \\ \hline
\begin{tabular}[c]{@{}l@{}}Feature \\ Transformation\end{tabular}           & No      & \{True, False\}                                                                             & PointNet  \\ \hline
Version                                                                     & No      & \{V2, V3-Large\}                                                                            & MobileNet \\ 
\bottomrule
\end{tabular}
\end{table}

\subsection{Partial Fusion}

Even when there are slight differences in the architectures or the operator types/shapes among the models in repetitive single-accelerator training jobs, HFTA can still improve the overall hardware utilization as long as some parts of the models are horizontally fused. \emph{Model architecture search}~\cite{neural_architecture_search} and \emph{ensemble learning}~\cite{mothernets} can be realistic use cases for such partial fusions. In this section, we want to study the performance sensitivity of the fused model with respect to the extent of the performed fusion.

To conduct such a study, we add \emph{partially fused} optimizers and learning rate schedulers to HFTA, and we extend the ResNet-18 benchmark such that we can individually configure whether each of the 8 basic blocks, the first convolutional block and the last linear block is horizontally fused. We then fix the number of jobs that share the same accelerator (V100) as a constant (30), and measure the per-device training throughput as we incrementally turn \emph{off} the horizontal fusion of each block. We provide the corresponding evaluation results in Appendix~\ref{append:secondary_eval}.

\section{Evaluation: Secondary Benchmarks} \label{append:secondary_eval}

\begin{figure*}[]
    \centering
    \vspace{-0.3cm}
    \begin{subfigure}[]{\linewidth}
        \centering
        \includegraphics[width=0.8\linewidth]{imgs/speedups_legend.png}
        \vspace{-0.2cm}
    \end{subfigure}
    \hfill
    \begin{subfigure}[]{0.49\linewidth}
        \centering
        \includegraphics[width=0.65\linewidth]{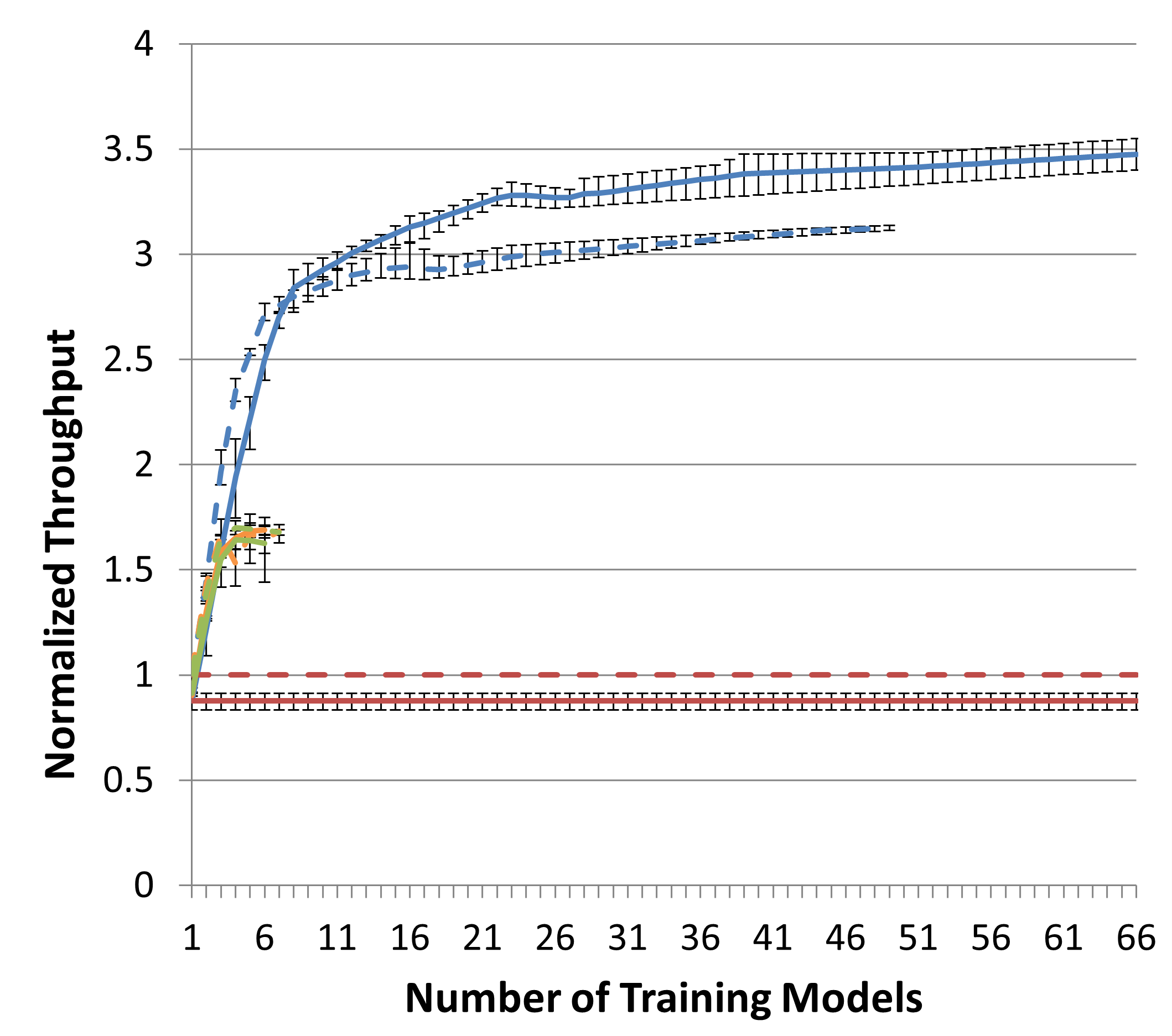}
        \vspace{-0.2cm}
        \caption{ResNet-18}
        \label{fig:speedups_resnet_v100}
    \end{subfigure}
    \hfill
    \begin{subfigure}[]{0.49\linewidth}
        \centering
        \includegraphics[width=0.65\linewidth]{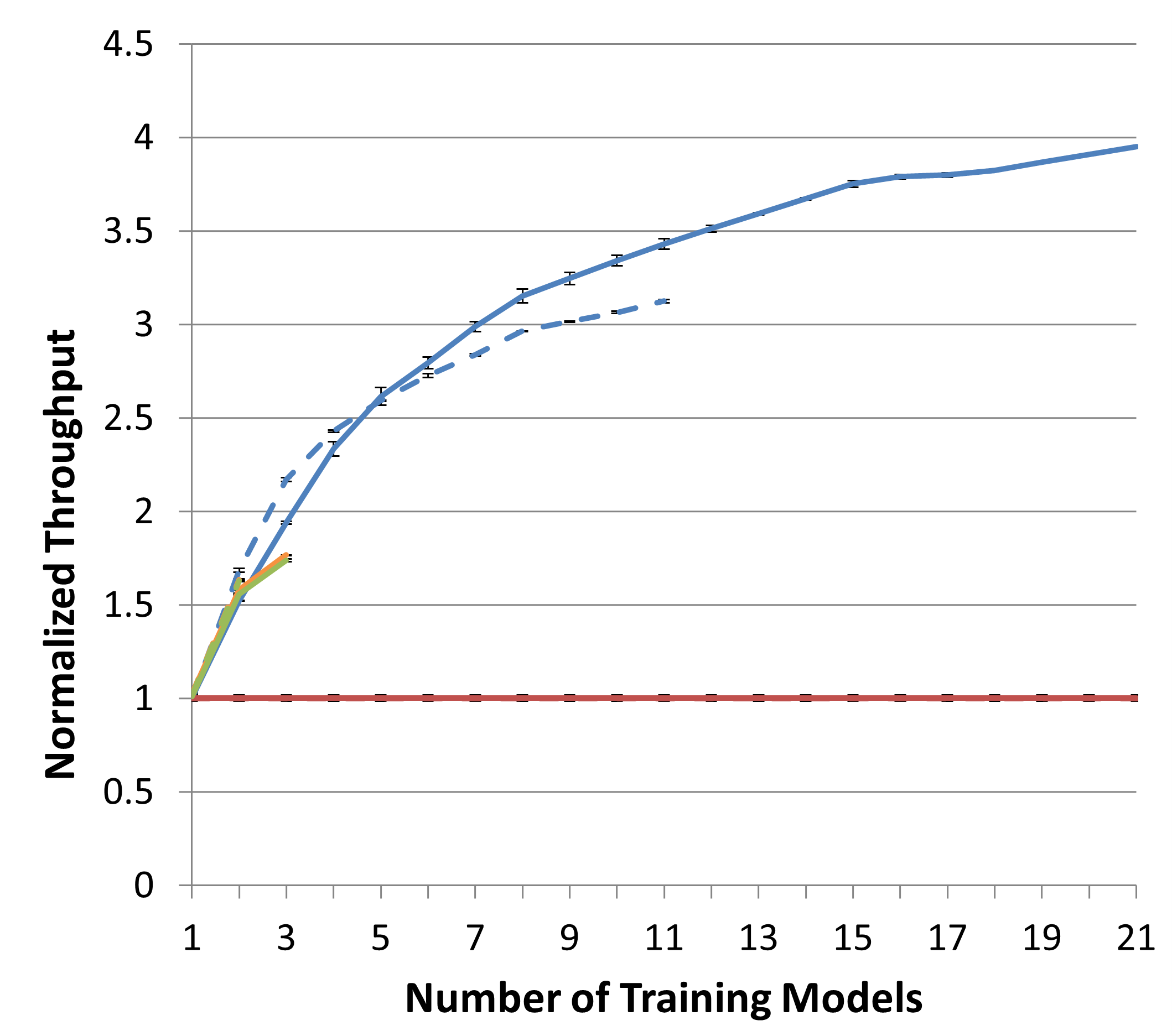}
        \vspace{-0.2cm}
        \caption{MobileNet-V3Large}
        \label{fig:speedups_mobilenetv3l_v100}
    \end{subfigure}
    \hfill
    \begin{subfigure}[]{0.49\linewidth}
        \centering
        \includegraphics[width=0.65\linewidth]{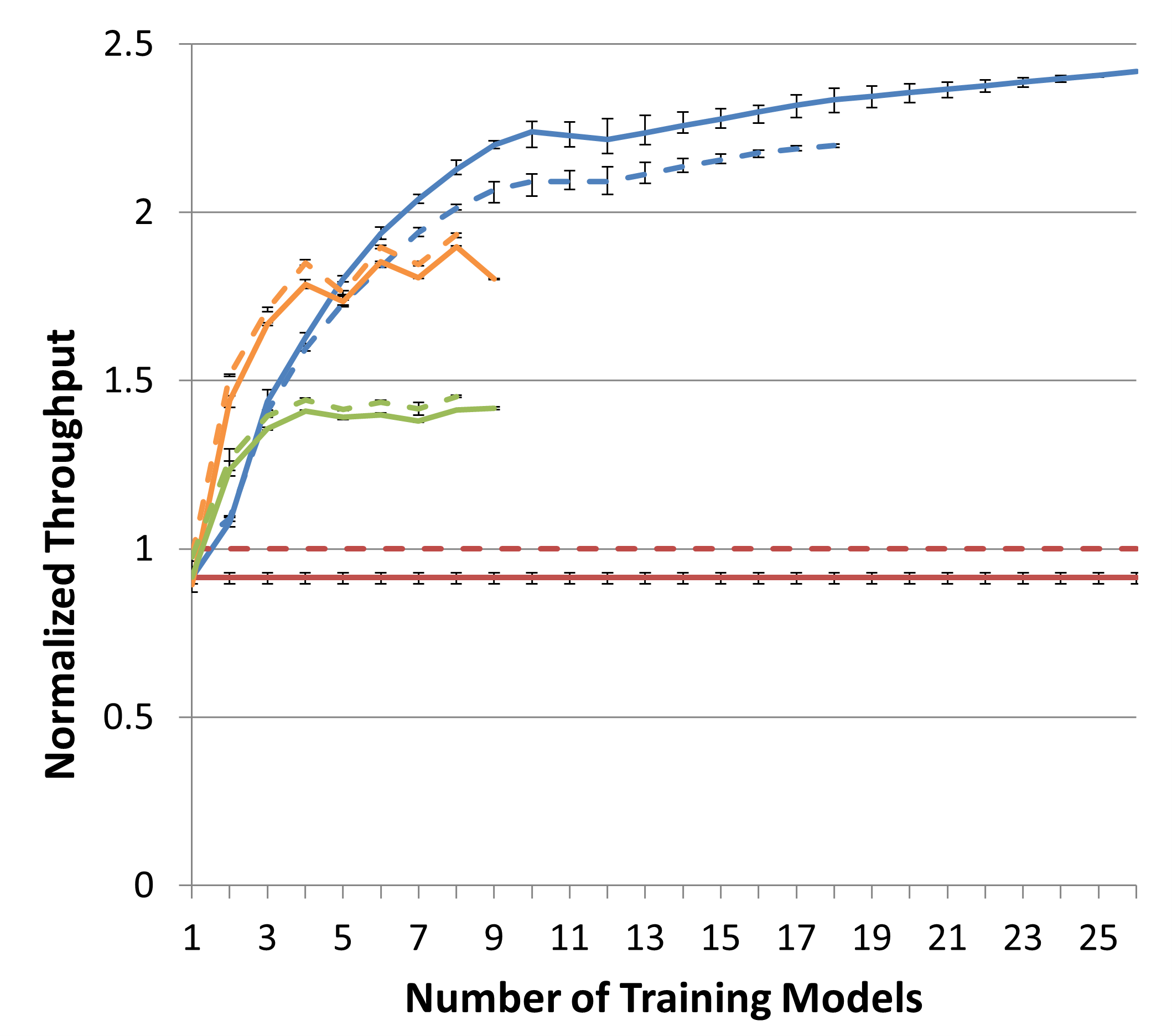}
        \vspace{-0.2cm}
        \caption{Transformer}
        \label{fig:speedups_transformer_v100}
    \end{subfigure}
    \hfill
    \begin{subfigure}[]{0.49\linewidth}
        \centering
        \includegraphics[width=0.65\linewidth]{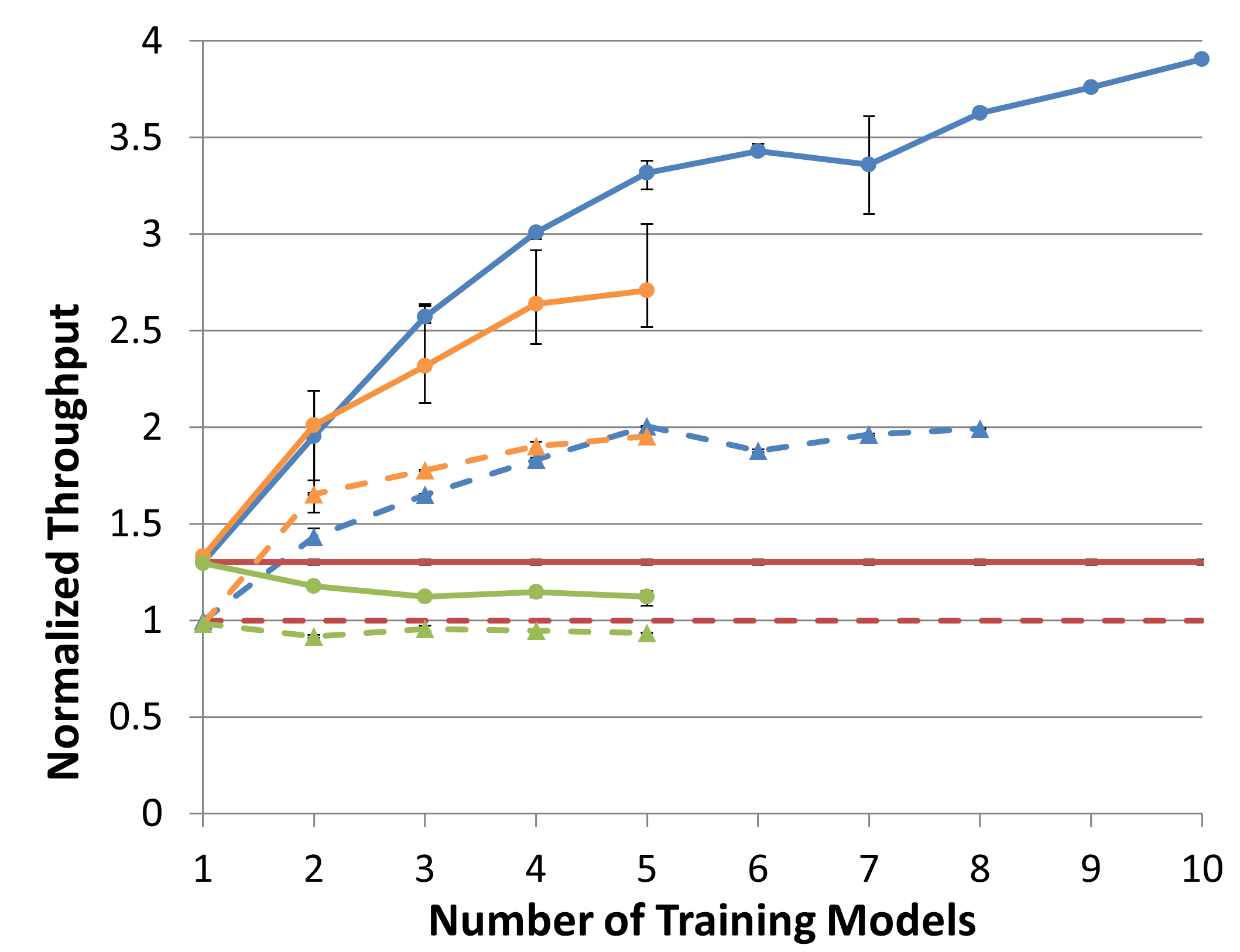}
        \vspace{-0.2cm}
        \caption{BERT-Medium}
        \label{fig:speedups_bert_v100}
    \end{subfigure}
    \vspace{-0.2cm}
    \caption{The normalized training throughput as we increase the number of models sharing the same V100 GPU.}
    \label{fig:speedups_gpus_secondary}
    \vspace{-0.0cm}
\end{figure*}

\begin{figure*}[]
    \centering
    \vspace{-0.3cm}
    \begin{subfigure}[]{\linewidth}
        \centering
        \includegraphics[width=0.3\linewidth]{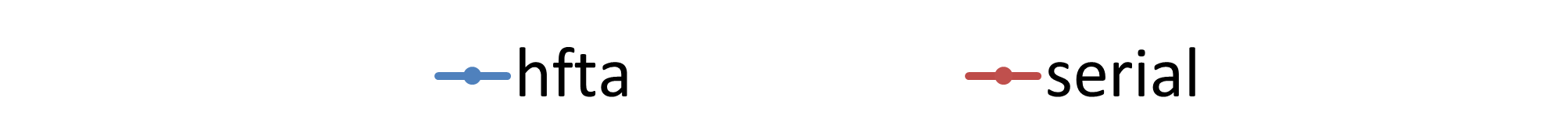}
        \vspace{-0.5cm}
    \end{subfigure}
    \hfill
    \begin{subfigure}[]{0.49\linewidth}
        \centering
        \includegraphics[width=0.65\linewidth]{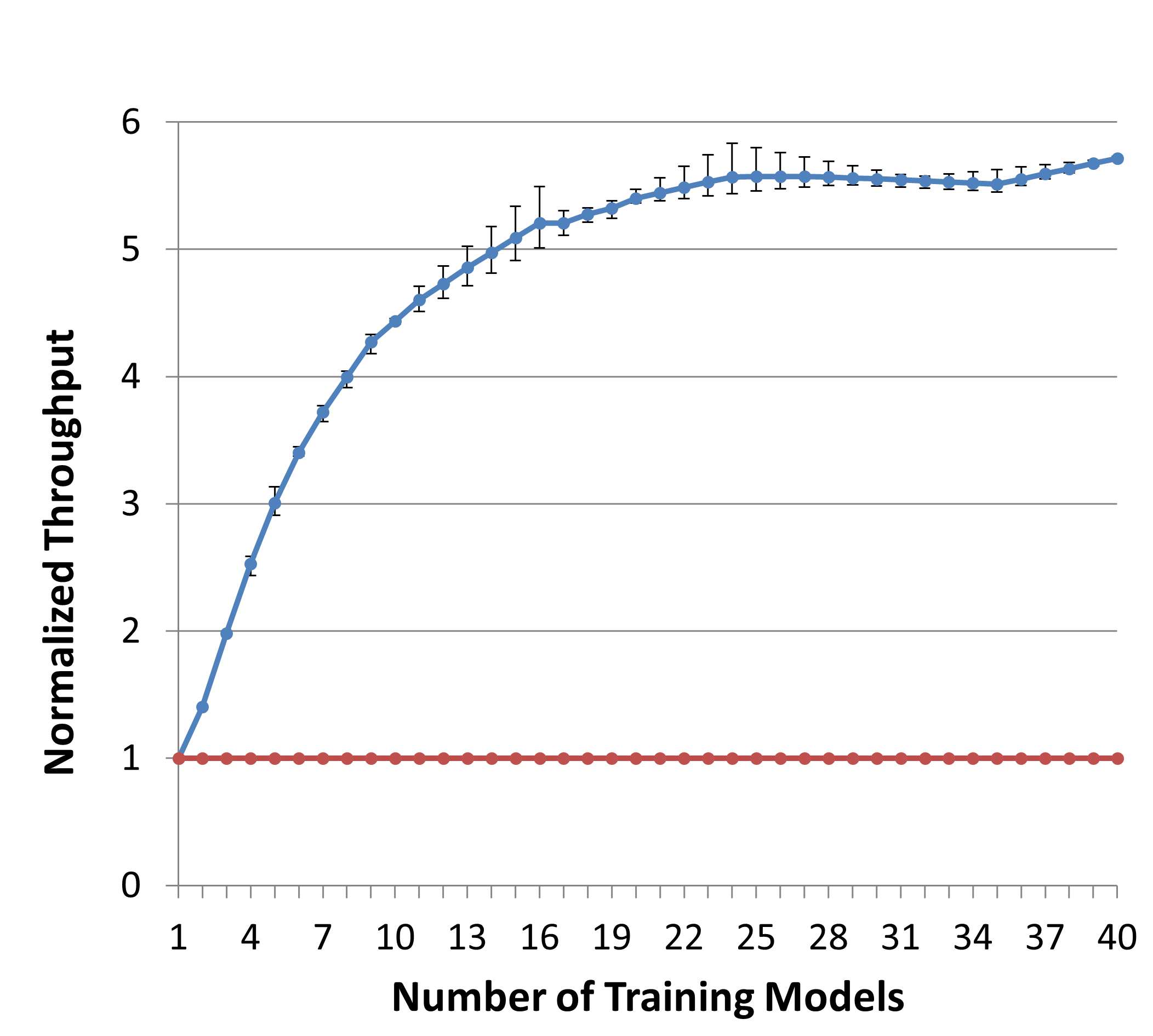}
        \vspace{-0.2cm}
        \caption{ResNet-18}
        \label{fig:speedups_resnet_tpu_v3}
    \end{subfigure}
    \hfill
    \begin{subfigure}[]{0.49\linewidth}
        \centering
        \includegraphics[width=0.65\linewidth]{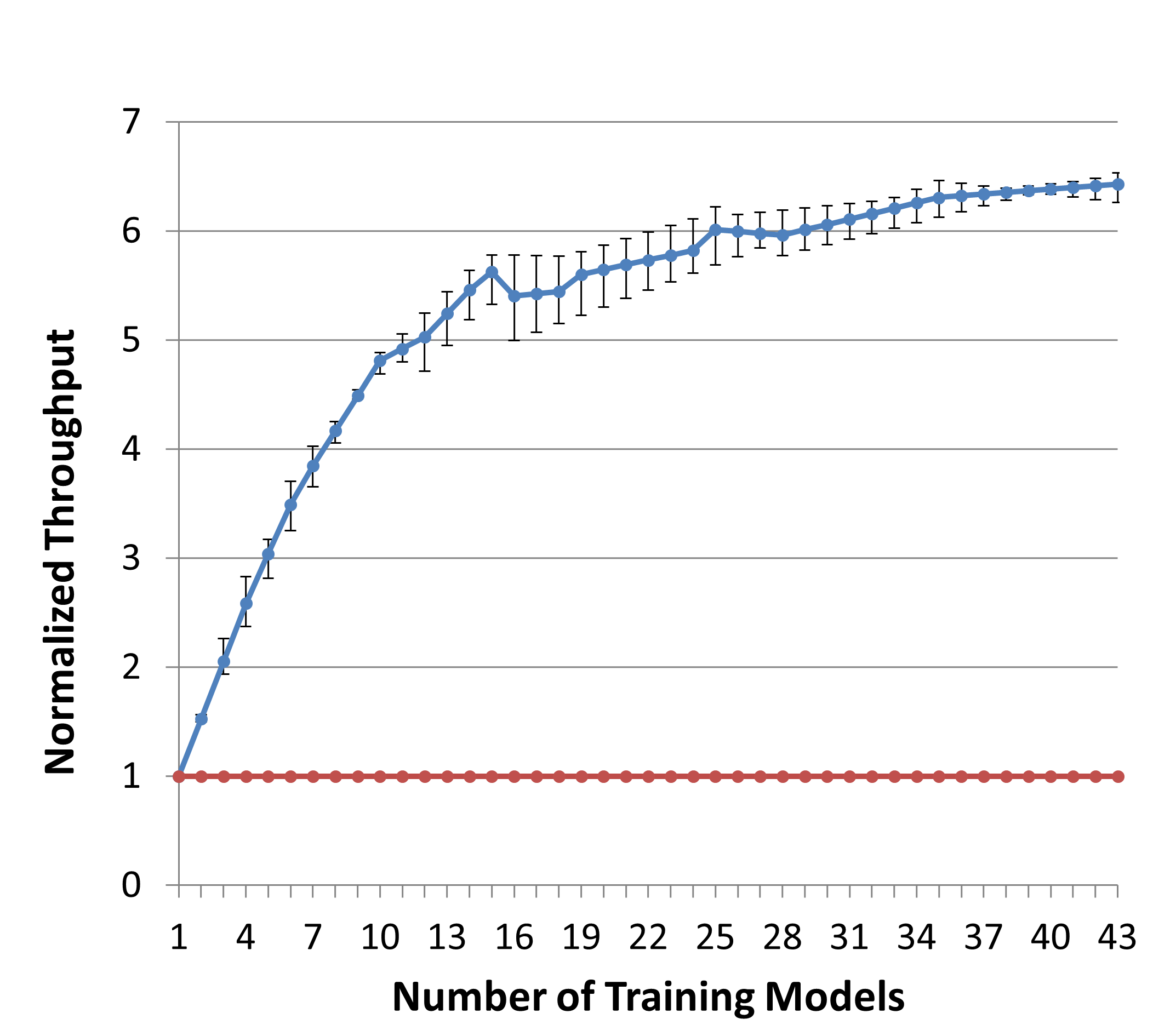}
        \vspace{-0.2cm}
        \caption{MobileNet-V3Large}
        \label{fig:speedups_mobilenetv3l_tpu_v3}
    \end{subfigure}
    \hfill
    \begin{subfigure}[]{0.49\linewidth}
        \centering
        \includegraphics[width=0.65\linewidth]{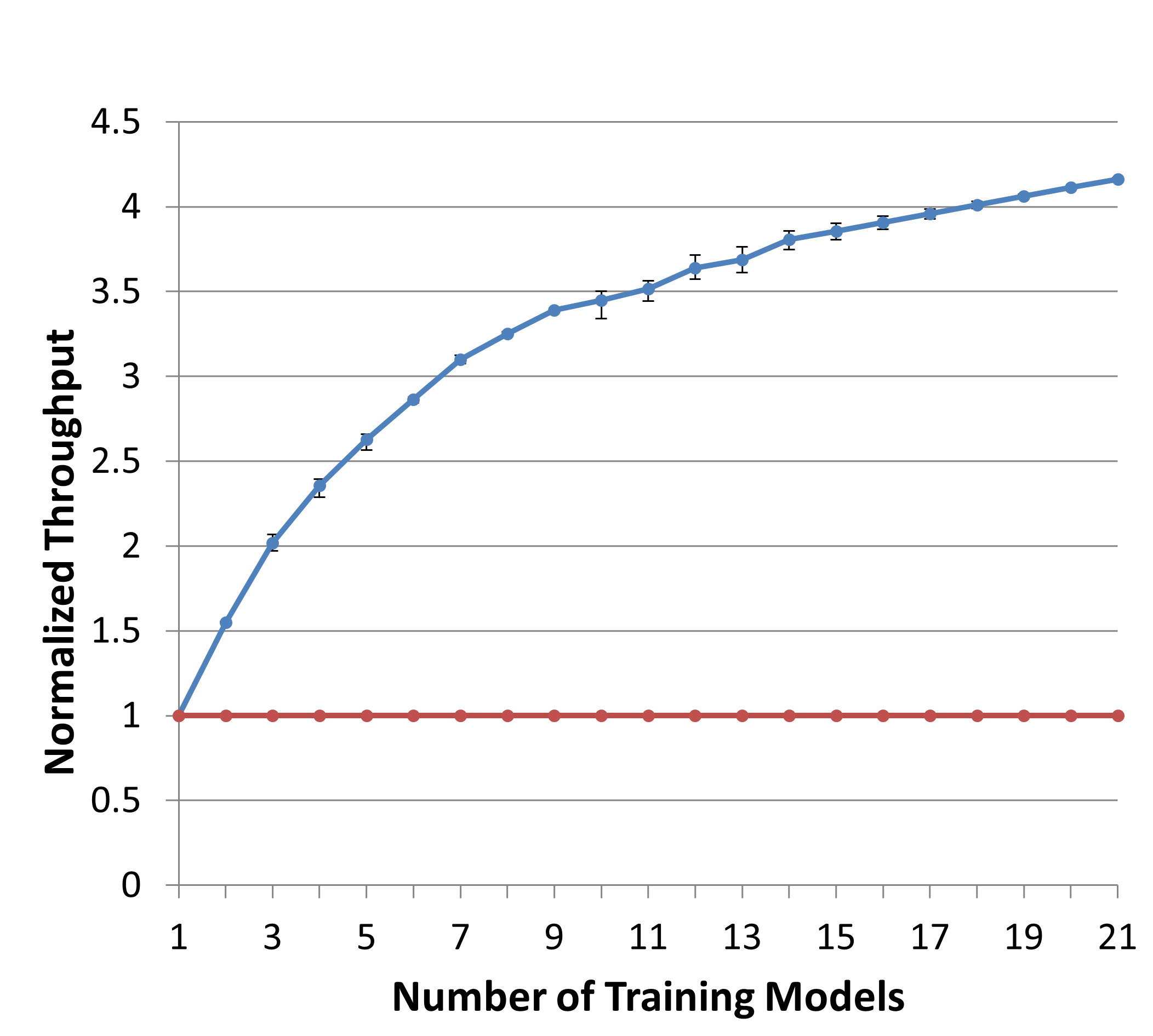}
        \vspace{-0.2cm}
        \caption{Transformer}
        \label{fig:speedups_transformer_tpu_v3}
    \end{subfigure}
    \hfill
    \begin{subfigure}[]{0.49\linewidth}
        \centering
        \includegraphics[width=0.65\linewidth]{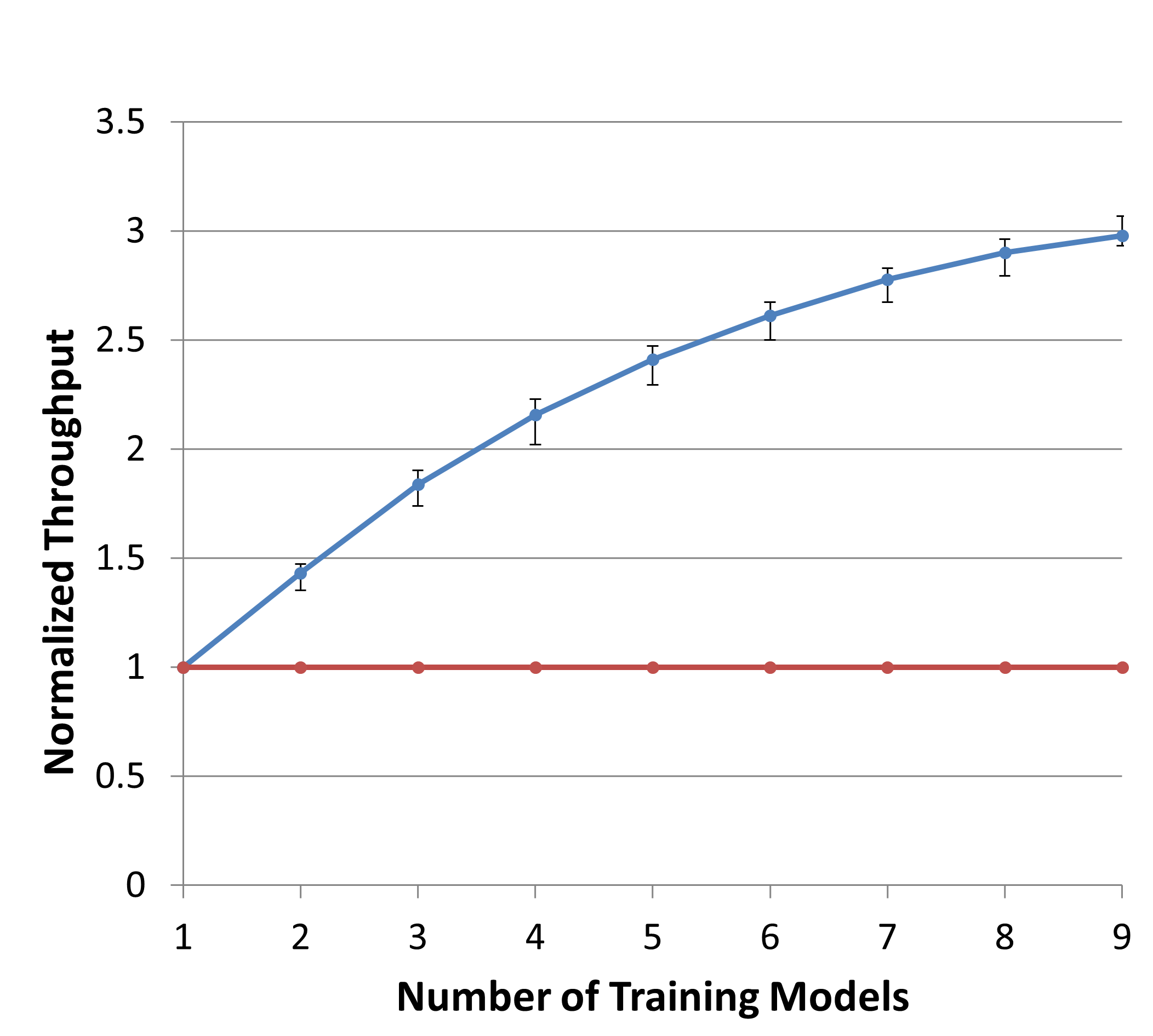}
        \vspace{-0.2cm}
        \caption{BERT-Medium}
        \label{fig:speedups_bert_tpu_v3}
    \end{subfigure}
    \vspace{-0.2cm}
    \caption{The normalized training throughput as we increase the number of models sharing the same TPU v3 core.}
    \label{fig:speedups_tpus_secondary}
    \vspace{-0.0cm}
\end{figure*}

\begin{figure}[]
    \centering
    \includegraphics[width=0.8\linewidth]{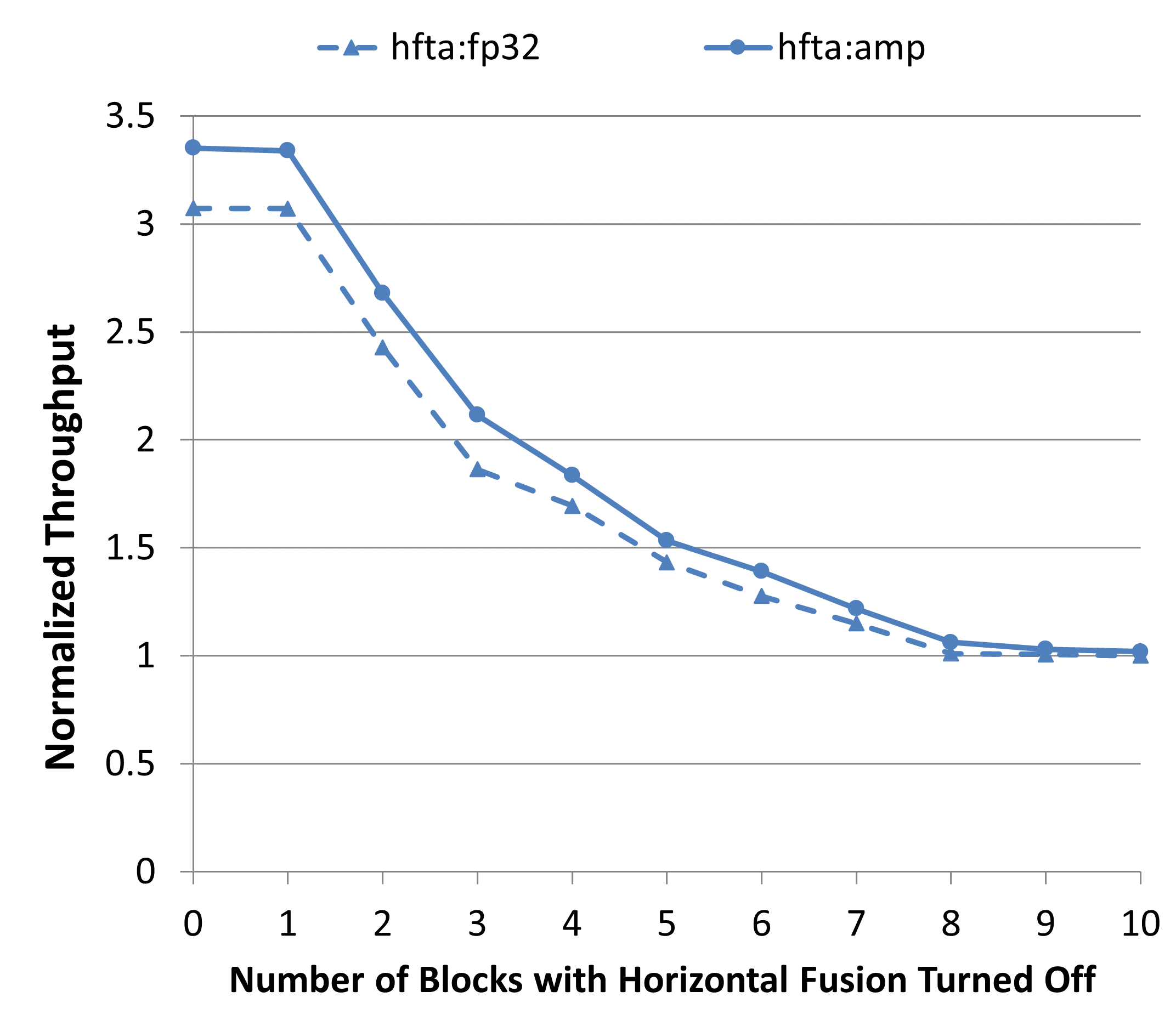}
    \caption{ResNet-18's normalized training throughput as we incrementally turn off the horizontal fusion for each block.}
    \label{fig:speedups_resnet_partially_fused_v100}
\end{figure}

Similar to Section~\ref{sec:eval_gpus} and Section~\ref{sec:eval_tpus}, to compare HFTA's end-to-end training performance with our baselines (i.e., \emph{serial}, \emph{concurrent}, and \emph{MPS}) for our secondary benchmarks (i.e., ResNet-18, MobileNet-V3Large, Transformer, and BERT-Medium), Figure~\ref{fig:speedups_gpus_secondary} and Figure~\ref{fig:speedups_tpus_secondary} plot the per-device normalized training throughput on the V100 GPUs and the TPU v3 respectively. 
We normalize the throughput for each experiment by the \emph{serial} (FP32 for V100) baseline. 
We show both FP32 and AMP training results for the experiments on V100. 
Each curve grows as we increase the number of models that share the same device. 
Each curve ``stops" when it reaches the maximum number of models before a out-of-memory error occurs, except that the curve for ResNet-18 on TPU stops when the throughput starts to degrade\footnote{When we check the TPU usage monitoring dashboard, we observe that the TPU memory usage surpasses its per-core memory capacity when we fuse too many models. Therefore, we hypothesize that the performance degradation could be due to certain memory optimizations (e.g., swapping or rematerialization) employed by the TPU's memory management system or the XLA compiler.}.

These figures coincide with the trends in our main observations from Section~\ref{sec:eval_gpus} and Section~\ref{sec:eval_tpus}.
%Additionally, we can observe from Figure~\ref{fig:speedups_resnet_tpu_v3} that the performance could degrade (compared to the HFTA's peak performance) if we fuse too many models together and run on a shared TPU. When we check the TPU usage monitoring dashboard, we observe that the TPU memory usage surpasses its per-core memory capacity when we fuse too many models. Therefore, we hypothesize that the performance degradation could be due to certain memory optimizations (e.g., swapping or rematerialization) employed by the TPU's memory management system or the XLA compiler.
Across these secondary benchmarks, on V100, HFTA achieves $\gpuPeakHFTAvsSerialUnderRangeNewModels \times$ to $\gpuPeakHFTAvsSerialUpperRangeNewModels \times$ higher peak training throughput over \emph{serial}, $\gpuPeakHFTAvsConcurrentUnderRangeNewModels \times$ to $\gpuPeakHFTAvsConcurrentUpperRangeNewModels \times$ over \emph{concurrent} and $\gpuPeakHFTAvsMPSUnderRangeNewModels \times$ to $\gpuPeakHFTAvsMPSUpperRangeNewModels \times$ over \emph{MPS}; on TPU v3, HFTA achieves $\tpuPeakHFTAvsSerialUnderRangeNewModels \times$ to $\tpuPeakHFTAvsSerialUpperRangeNewModels \times$ peak training throughput than \emph{serial}. 
Therefore, we conclude that HFTA can be also effective in improving the hardware utilization for conventional models during repetitive single-accelerator training.

To evaluate the partial fusion's performance sensitivity, Figure~\ref{fig:speedups_resnet_partially_fused_v100} plots the normalized training throughput when 30 ResNet-18 models share the same V100 GPU as we incrementally turn off the horizontal fusion for each block. 
We make two observations from this figure. 
First, a higher degree of fusion does lead to better performance.
Thus, even if the models among repetitive training jobs can not be fully fused, \emph{every little bit of fusion can still be helpful.} 
Second, the fusion of different blocks (thus, different operators) contributes differently to the overall performance improvement. 
This is because some operators within a model suffer more from hardware under-utilization than others.
%%%%%%%%%%%%%%%%%%%%%%%%%%%%%%%%%%%%%%%%%%%%%%%%%%%%%%%%%%%%%%%%%%%%%%%%%%%%%%%
%%%%%%%%%%%%%%%%%%%%%%%%%%%%%%%%%%%%%%%%%%%%%%%%%%%%%%%%%%%%%%%%%%%%%%%%%%%%%%%

\end{document}